\documentclass[lettersize,journal]{IEEEtran}
\usepackage{amsmath,amsfonts}
\usepackage{algorithmic}
\usepackage{algorithm}
\usepackage{array}
\usepackage{textcomp}
\usepackage{stfloats}
\usepackage{url}
\usepackage{verbatim}
\usepackage{graphicx}
\usepackage{cite}
\usepackage{caption}
\usepackage{subcaption}
\usepackage{booktabs} 
\usepackage{multirow}
\usepackage{graphicx}
\usepackage{hyperref}
\usepackage{siunitx}
\usepackage{pifont}
\usepackage[inkscapelatex=false]{svg}
\usepackage{xcolor}
\definecolor{amber}{rgb}{1.0, 0.95, 0.8}
\definecolor{amber2}{rgb}{1.0, 0.90, 0.8}

\begin{document}

\title{STHN: Deep Homography Estimation for UAV Thermal Geo-localization with Satellite Imagery}

\author{Jiuhong Xiao$^1$, 
Ning Zhang${^2}{^\dag}$,   
Daniel Tortei${^2}{^\dag}$, 
and Giuseppe Loianno$^1$
\thanks{Manuscript received: March 28, 2024; Revised: June 24, 2024; Accepted: August 12, 2024. This paper was recommended for publication by Editor Cesar Cadena Lerma upon evaluation of the Associate Editor and Reviewers’ comments. This work was supported by the Technology Innovation Institute, the NSF CAREER Award 2145277, the NSF CPS Grant CNS-2121391, and the NYU IT High Performance Computing resources, services, and staff expertise. Giuseppe Loianno serves as consultant for the Technology Innovation Institute. This arrangement has been reviewed and approved by the New York University in accordance with its policy on objectivity in research.}
\thanks{$^\dag$denotes equal contribution.}
\thanks{$^1$The authors are with the New York University, Brooklyn, NY 11201, USA. {\tt\footnotesize email: \{jx1190, loiannog\}@nyu.edu}.}
\thanks{$^2$The authors are with the Autonomous Robotics Research Center-Technology Innovation Institute, Abu Dhabi, UAE. {\tt\footnotesize email:  \{ning.zhang, daniel.tortei\}@tii.ae}.}
\thanks{Digital Object Identifier (DOI): see top of this page.}
}



\maketitle

\markboth{IEEE Robotics and Automation Letters. Preprint Version. Accepted August, 2024}
{FirstAuthorSurname \MakeLowercase{\textit{et al.}}: ShortTitle} 

\begin{abstract}
Accurate geo-localization of Unmanned Aerial Vehicles (UAVs) is crucial for outdoor applications including search and rescue operations, power line inspections, and environmental monitoring. The vulnerability of Global Navigation Satellite Systems (GNSS) signals to interference and spoofing necessitates the development of additional robust localization methods for autonomous navigation. Visual Geo-localization (VG), leveraging onboard cameras and reference satellite maps, offers a promising solution for absolute localization. Specifically, Thermal Geo-localization (TG), which relies on image-based matching between thermal imagery with satellite databases, stands out by utilizing infrared cameras for effective nighttime localization. However, the efficiency and effectiveness of current TG approaches, are hindered by dense sampling on satellite maps and geometric noises in thermal query images. To overcome these challenges, we introduce STHN, a novel UAV thermal geo-localization approach that employs a coarse-to-fine deep homography estimation method. This method attains reliable thermal geo-localization within a 512-meter radius of the UAV's last known location even with a challenging 11\% size ratio between thermal and satellite images, despite the presence of indistinct textures and self-similar patterns. We further show how our research significantly enhances UAV thermal geo-localization performance and robustness against geometric noises under low-visibility conditions in the wild. The code is made publicly available.
\end{abstract}

\begin{IEEEkeywords}
Deep Learning for Visual Perception, Aerial Systems: Applications, Localization
\end{IEEEkeywords}
\vspace{-10pt}

\section*{Supplementary Material}
\noindent \textbf{Project page:}  \url{https://xjh19971.github.io/STHN/}
\section{Introduction}~\label{sec:introduction}
The increasing deployment of Unmanned Aerial Vehicles (UAVs) across a diverse range of applications, including agriculture~\cite{agriculture}, search and rescue operations~\cite{searchrescue}, tracking~\cite{saviolo2023unifying}, power line inspections~\cite{powerline}, and solar power plant inspections~\cite{drones6110347}, underscores the growing importance of robust UAV localization for autonomous navigation to guarantee the effective execution of these tasks. In outdoor environments, absolute localization technology~\cite{review_avl} is crucial as reliance on relative localization methods can cause error accumulation over time, particularly during long-time missions or in scenarios lacking loop closure detection. While Global Navigation Satellite Systems (GNSS) have become the preferred solutions, their reliability can be compromised by vulnerabilities to signal interference, jamming, and spoofing. Visual geo-localization~\cite{foundloc, vgscience, directalign3, imgregistration} emerges as a significant alternative solution, utilizing onboard cameras to facilitate absolute localization and navigation. This approach aligns captured RGB imagery, taken from nadir (top-down) or oblique views, with an existing reference map (such as a satellite map), enabling accurate positioning in GNSS-denied environments. However, this approach poses significant challenges in low-visibility or nighttime environments.

\begin{figure}
    \centering
    \includegraphics[width=\linewidth]{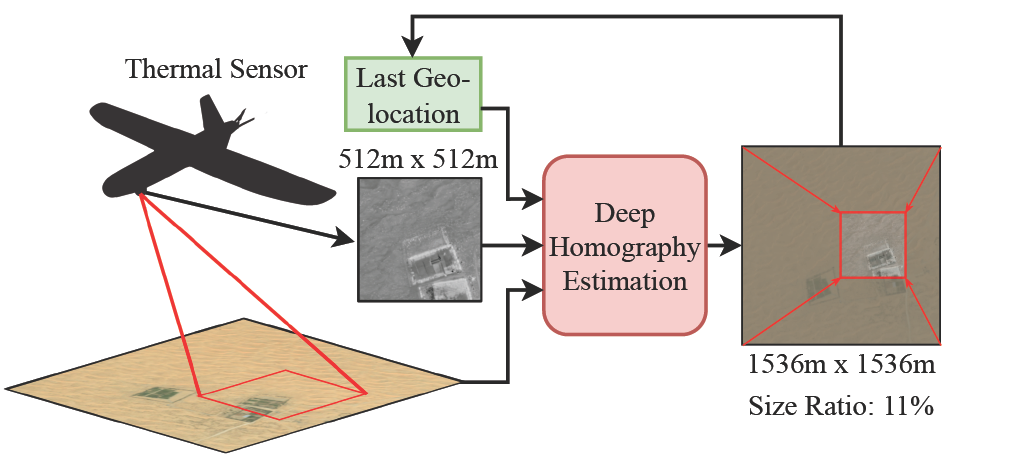}
    \caption{STHN framework for UAV thermal geo-localization with satellite maps. This framework achieves robust UAV thermal localization with a challenging size ratio of 11\% between thermal and satellite images. }
    \label{teaser}
    \vspace{-15pt}
\end{figure}

In response to these challenges, recent advancements in UAV thermal geo-localization~\cite{stl} explore an image-based matching approach with an onboard thermal camera to match nadir-view images to satellite image crops from a database. However, this method encounters several drawbacks. Firstly, the localization accuracy is majorly influenced by the density of satellite image samples in the database. Reducing the sampling interval improves continuity between image crops and localization accuracy but increases computation time and memory usage for extensive sampling. Additionally, the approach has limited tolerance for thermal images that are not correctly north-aligned, with geometric distortions negatively impacting localization accuracy.

Addressing these limitations, this study introduces Satellite-Thermal Homography Network (STHN) framework  (see Fig.~\ref{teaser}) that leverages deep homography estimation techniques~\cite{cao2022iterative, detone2016deep, 10.1007/978-3-030-58452-8_38, zhao2021deep, shao2021localtrans, Cao_2023_CVPR, le2020deep} to directly align thermal images with satellite maps of the local region, optimizing localization in GNSS-denied scenarios. This approach adopts a two-stage coarse-to-fine strategy: 1) \textit{Coarse alignment}, which matches small thermal images to large satellite maps within a search radius of $512~\si{m}$ with a challenging constant size ratio of $11\%$; and 2) \textit{Refinement}, which crops and resizes the selected region and applies a second-stage estimation for enhanced accuracy. 

The main contributions of this research are outlined as follows. First, we introduce for the first time a novel satellite-thermal Deep Homography Estimation (DHE) method based on an efficient coarse-to-fine approach tailored for UAV nighttime Thermal Geo-localization (TG), eliminating the dense satellite map sampling requirement of~\cite{stl}. Second, we introduce the Thermal Generative Module (TGM)~\cite{stl} into our DHE framework, improving the alignment between thermal and satellite images with significant scale change using limited satellite-thermal paired data. Third, we validate our approach by considering extensive and comprehensive experiments in challenging scenarios where thermal images have indistinct self-similar features on the deserts and a low overlap rate ($11\%$) with satellite images. We demonstrate the superior performance of our method over state-of-the-art real-time DHE methods and better efficiency and accuracy over image-based matching methods. Our results also demonstrate that STHN can effectively tolerate and estimate certain geometric noises including rotation, resizing, and perspective transformation noises for thermal geo-localization. To our knowledge, this is the first deep homography estimation solution for UAV thermal geo-localization, facilitating reliable nighttime localization over long-distance outdoor flights.

\section{Related Works}~\label{sec:relatedworks}
\textbf{UAV Visual and Thermal Geo-localization.} UAV visual geo-localization technology has been explored by multiple works based on: 1) Template matching methods~\cite{directalign, directalign2} perform dense image alignment to optimize the image similarity measures; 2) Traditional keypoint matching methods~\cite{imgregistration, imgregistration2} extract and match the keypoints using hand-crafted detector and descriptors; and 3) Deep-learning-based matching methods~\cite{vgscience, foundloc, voplusvg, VGUAV5, VGUAV4} utilize deep neural network~\cite{lecun2015deep} to generate robust matching features against environmental noises. For UAV thermal localization with nadir views, \cite{thermaluav3, thermaluav4} adopt Thermal Inertial Odometry (TIO) for navigating short-distance outdoor flights. For long-distance geo-localization, \cite{pmlr-v155-achermann21a} uses keypoint-based visible-thermal image registration, whereas \cite{stl} employs image-based matching with generative models and domain adaptation for enhanced cross-spectral geo-localization with limited training data. Despite the efficiency of keypoint-based methods, their reliance on repeatable cross-spectral local features limits their applicability. In contrast, image-based matching methods~\cite{stl, keetha2023anyloc}, free from this requirement, face challenges with exhaustive searches and high memory demands, with performances that are heavily dependent on satellite database density. Our research diverges by introducing deep homography estimation for precise satellite and thermal image alignment, presenting a novel geo-localization framework that surpasses prior limitations by eliminating the necessity for repeatable local features or exhaustive searches, improving accuracy and efficiency.

\textbf{Deep Homography Estimation.} Deep homography estimation is first proposed by~\cite{detone2016deep}, which uses four-corner displacement as the parametrization of homography estimation and four-corner perturbed images to train the model. \cite{10.1007/978-3-030-58452-8_38} develops a content-aware deep homography estimation approach against the noise from the dynamic dominant foreground. \cite{zhao2021deep} employs inverse compositional Lucas-Kanade algorithms for multi-modal image alignment. In \cite{shao2021localtrans}, the authors propose LocalTrans to conduct cross-resolution homography estimation. \cite{cao2022iterative} shows an iterative process to iteratively refine the homography estimation results in real-time, whereas \cite{Cao_2023_CVPR} uses a focus transformer for global and local correlation to enhance estimation performance. Considering UAV localization, \cite{nguyen2018unsupervised} proposes to use an unsupervised approach with photometric consistency loss for warped aerial RGB images while requiring about $65\%$ overlap between two source images. For thermal imagery, \cite{electronics12040788} employs a multi-scale conditional GAN architecture~\cite{cgan} to conduct thermal-visible homography estimation. The subsequent work~\cite{electronics12214441} shifts to a coarse-to-fine paradigm to further improve the estimation performance. However, the previous works commonly require a minimum overlap of $25\%$ and, in rare instances, exactly $25\%$. Compared to these works, our approach adopts a coarse-to-fine paradigm but considers coarse estimation across images with major scale change for large search regions. This results in a challenging constant $11\%$ size ratio. For refinement, our approach differs from~\cite{cao2022iterative, zhao2021deep, shao2021localtrans, electronics12214441}, which typically upsample aligned images. Given the small size ratio of the thermal image, a large portion of the satellite image becomes redundant and can even hinder the refinement process. Instead, we crop the selected satellite region and perform estimation without increasing image resolution to enhance efficiency.
\vspace{-5pt}

\begin{figure*}
    \centering
    \includegraphics[trim={0 0 1.8cm 0},clip,width=0.85\linewidth]{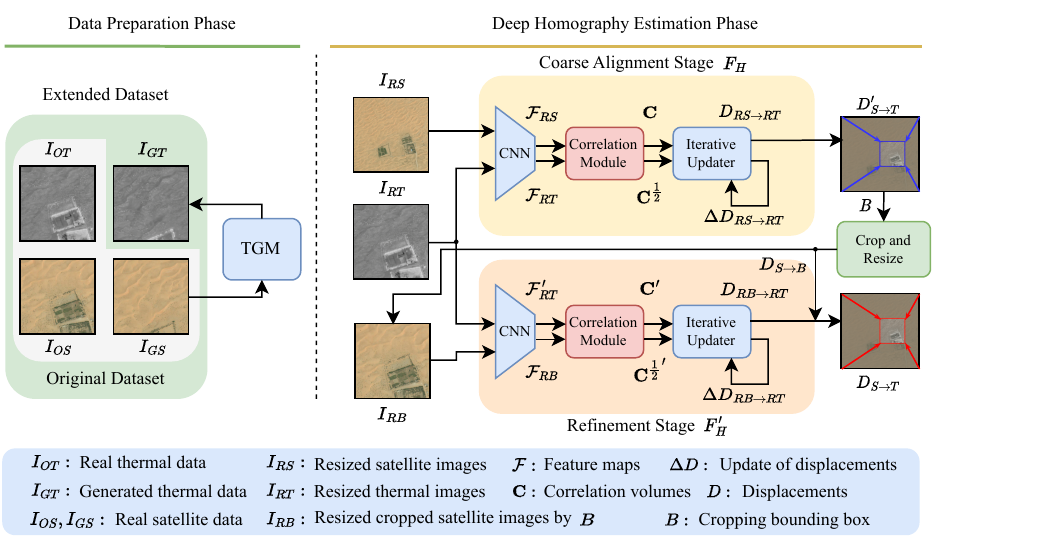}
    \vspace{-10pt}
    \caption{STHN Framework Overview: For the data preparation phase, TGM produces synthetic thermal images from unpaired satellite images, augmenting the dataset. The deep homography estimation phase employs $F_H$ for the \colorbox{amber}{Coarse Alignment Stage} by predicting the displacement $D_{RS\rightarrow RT}$ between thermal images and satellite maps. For the\colorbox{amber2}{Refinement Stage}, the framework crops and resizes the selected region $B$, utilizing $F^\prime_H$ to fine-tune the four-corner displacement prediction for enhanced accuracy.}
    \label{fig:framework}
    \vspace{-15pt}
\end{figure*}


\section{Methodology}~\label{sec:methodology}
Our STHN framework, shown in Fig.~\ref{fig:framework}, has three main components: Thermal Generative Module (TGM), coarse alignment module, and refinement module.
\vspace{-5pt}

\subsection{Thermal Generative Module (TGM)}\label{sec:TGM}
We employ TGM~\cite{stl} to enhance our training dataset with synthetic thermal images derived from satellite images. In the data preparation phase, we denote $I_{OS}$ and $I_{OT}$ as the pair of satellite and 8-bit thermal images from the original dataset, and $I_{GS}$ as the satellite images without paired thermal images. We train TGM with the input $I_{OS}$ and target output $I_{OT}$ following pix2pix~\cite{pix2pix} approach. After training TGM, we generate synthetic thermal images $I_{GT}$ using TGM and $I_{GS}$, and combine $I_{OS}$ and $I_{OT}$ to build an extended satellite-thermal dataset. We denote the quantity of actual thermal images as $N_T$ and those generated as $N_G$. We restrict our sampling from the generated dataset per epoch to $N_T$ instances to mitigate bias towards $I_{GT}$, given that $N_T \ll N_G$.

\subsection{Coarse-to-fine Iterative Homography Estimation}
Our coarse-to-fine strategy is divided into two stages: Coarse alignment and refinement.
\subsubsection{Coarse Alignment Stage} We denote $W_S$ as the width of input square satellite images $I_S$ and $W_T$ as that of input square 8-bit thermal images $I_T$. For pre-processing, we resize $I_S$ and $I_T$ to $I_{RS}$ and $I_{RT}$ at the side length of $W_R$, and the resize ratios of $I_{RS}$ is $\alpha = W_S / W_R$. Then, we run the model
\begin{equation}
    D_{RS \rightarrow RT} = F_H(I_{RS}, I_{RT}),
\end{equation}
where $D_{RS\rightarrow RT}\in \mathbb{R}^{2\times4}$ is the displacement from the four corners of $I_{RS}$ to those of $I_{RT}$ and $F_H$ is the homography estimation model. In other words, $D_{RS\rightarrow RT}$ aligns $I_{RT}$ into $I_{RS}$. $F_H$ follows an iterative estimation paradigm~\cite{cao2022iterative}, which consists of three modules: A Convolutional Neural Network (CNN)~\cite{lecun2015deep} feature extractor (multiple residual blocks with multi-layer CNNs and instance normalization) outputs the feature map $\mathcal{F}_{RS},\mathcal{F}_{RT}\in \mathbb{R}^{C\times\frac{W_R}{4}\times\frac{W_R}{4}}$ ($C=256$), a correlation module outputs correlation volumes~\cite{raft} $\mathbf{C}$ ($\frac{W_R}{4}\times\frac{W_R}{4}\times\frac{W_R}{4}\times\frac{W_R}{4}$) and $\mathbf{C}^\frac{1}{2}$ ($\frac{W_R}{4}\times\frac{W_R}{4}\times\frac{W_R}{8}\times\frac{W_R}{8}$), and an iterative homography estimator (multi-layer CNNs with group normalization) provides updates of displacement $\Delta D_{RS \rightarrow RT}$. At iteration $k$, $D_{RS \rightarrow RT}$ is updated as
\begin{equation}\label{iterative}
    D_{k+1, RS \rightarrow RT} = D_{k,RS \rightarrow RT} + \Delta D_{k, RS \rightarrow RT}.
\end{equation} Since the images are resized during pre-processing, the displacement of the coarse alignment stage on the scale of $I_S$ is $D^\prime_{S\rightarrow T} = \alpha D_{RS\rightarrow RT}$. For the loss function, we minimize the L1 distance between the predicted displacements $D_{k,RS \rightarrow RT}$ and ground truth ones $D^{gt}_{k,RS \rightarrow RT}$ with exponential decay as
\begin{equation}
    \mathcal{L_\textrm{coarse}} = \sum_{k=0}^{K_1-1}\gamma^{K_1 -k-1}\Vert  D_{k,RS \rightarrow RT} -  D^{gt}_{k,RS \rightarrow RT}\Vert_1,
\end{equation}
where $K_1$ is the number of updates in the coarse alignment. $D_{RS \rightarrow RT} = D_{K_1, RS \rightarrow RT}$. The decay factor $\gamma$ is $0.85$.

\subsubsection{Refinement Stage} We create a bounding box $B$ that bounds the corners of thermal images warped by $D^\prime_{S\rightarrow T}$. We set $B$ orthogonal to the image frame to ensure complete coverage of the target region, even if the coarse alignment result has rotation or perspective transformation errors. We denote $D_{S\rightarrow B} \in \mathbb{R}^{2 \times 4}$ as the four-corner displacement from $I_S$ to $B$. We crop out the region of $B$ to get $I_B$ at the side length of $W_B$ and resize it to $I_{RB}$ at the side length of $W_R$. The resize ratio is $\eta = W_B / W_R$. The refinement process is
\begin{equation}
    D_{RB\rightarrow RT} = F^\prime_H(I_{RB}, I_{RT}),
\end{equation}
where $F^\prime_H$ has the same structure as $F_H$ with iterative updates (see Eq.~\ref{iterative}) but does not share weights and $D_{RB\rightarrow RT}\in \mathbb{R}^{2\times4}$ are four-corner displacement from $I_{RB}$ to $I_{RT}$. We set $\kappa=\eta/\alpha$ and the loss function is 
\begin{equation}
    \mathcal{L_\textrm{fine}} = \sum_{k=0}^{K_2-1}\gamma^{K_2-k-1}\kappa\Vert  D_{k, RB\rightarrow RT} -  D^{gt}_{k,RB\rightarrow RT}\Vert_1,
\end{equation}
where $D_{k, RB\rightarrow RT}$ and $ D^{gt}_{k,RB\rightarrow RT}$ are predicted and ground truth displacements, and $K_2$ is the number of updates in the refinement. $\kappa$ maps the displacement from the scale of $I_{RB}$ to the scale of $I_{RS}$, aligning with $\mathcal{L_\textrm{coarse}}$. 
The total loss function is
\begin{equation}
    \mathcal{L} = \mathcal{L_\textrm{coarse}} + \mathcal{L_\textrm{fine}}.
\end{equation}
The displacement of the refinement stage on the scale of $I_S$ is $D_{B\rightarrow T} = \eta D_{RB\rightarrow RT}$. Combining the two stages' results, we get final displacements
\begin{equation}
    D_{S\rightarrow T} = D_{S\rightarrow B} + D_{B\rightarrow T}.
\end{equation}
With $D_{S\rightarrow T}$, we use Direct Linear Transformation (DLT)~\cite{abdel2015direct} to solve the homography matrix $H\in \mathbb{R}^{3\times3}$. The geo-localization center coordinate $(x_c, y_c)$ is calculated as
\begin{equation}
    \left(x_c, y_c, 1\right)^\top = H \times \left(\frac{W_S}{2}, \frac{W_S}{2}, 1\right)^\top.
\end{equation}

\vspace{-15pt}
\subsection{Two-stage Training Strategy}
For training the two-stage model, we first train the coarse alignment module from scratch, and then we attach the refinement module to the end of the coarse alignment module and jointly fine-tune the two modules. We discovered that augmenting the bounding box $B$ is crucial for effectively fine-tuning the refinement module. This requirement arises because the refinement module always tends to make no or only minor adjustments if the coarse alignment already performs well on training and validation sets. Furthermore, we observe that merely fixedly expanding the cropped boxes without random shifting and enlargement does not enhance performance. To boost the refinement module's effectiveness, we augment $B$ by shifting the center coordinates $(x_B, y_B)$ by $(\Delta p_1, \Delta p_2)$ and expanding the width $W_B$ by $2\Delta p_3$ during training. During the evaluation phase, we consistently expand $W_B$ by $\Delta p_4$ to mitigate the potential offset error of the coarse alignment.
\vspace{-10pt}

\section{Experimental Setup}~\label{sec:experiment_setting}
\vspace{-30pt}
\subsection{Dataset}\label{dataset}
\begin{figure}
\smallskip
\smallskip
    \centering
\rotatebox{90}{\scriptsize\hspace{-0.5em}Satellite ($W_S=512$)}
\begin{subfigure}[b]{0.11\textwidth}
    \includegraphics[width=\textwidth]{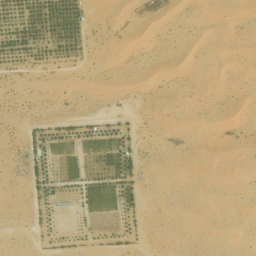}
    \vspace{-0.8\baselineskip}
\end{subfigure}
\begin{subfigure}[b]{0.11\textwidth}
    \includegraphics[width=\textwidth]{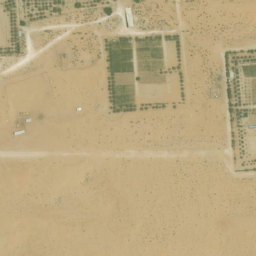}
    \vspace{-0.8\baselineskip}
\end{subfigure}
\begin{subfigure}[b]{0.11\textwidth}
    \includegraphics[width=\textwidth]{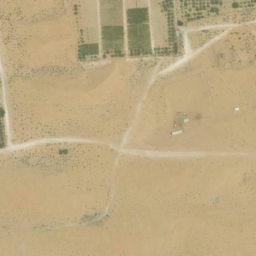}
    \vspace{-0.8\baselineskip}
\end{subfigure}
\begin{subfigure}[b]{0.11\textwidth}
    \includegraphics[width=\textwidth]{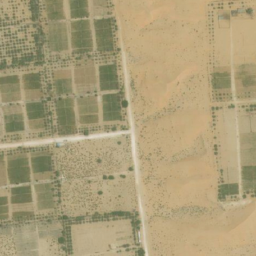}  
    \vspace{-0.8\baselineskip}
\end{subfigure}
\rotatebox{90}{\scriptsize\hspace{2.75em}Thermal}
\begin{subfigure}[b]{0.11\textwidth}
    \includegraphics[width=\textwidth]{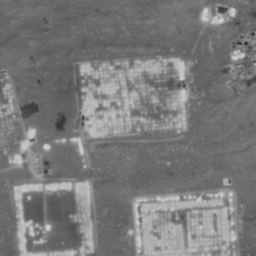}
    \vspace{-0.8\baselineskip}
\end{subfigure}
\begin{subfigure}[b]{0.11\textwidth}
    \includegraphics[width=\textwidth]{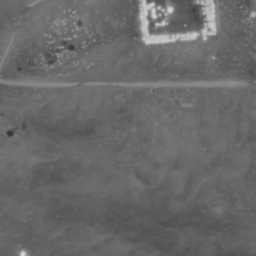}
    \vspace{-0.8\baselineskip}
\end{subfigure}
\begin{subfigure}[b]{0.11\textwidth}
    \includegraphics[width=\textwidth]{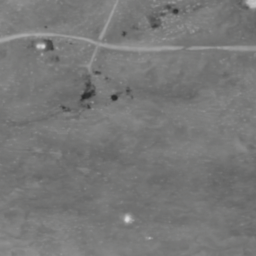}
    \vspace{-0.8\baselineskip}
\end{subfigure}
\begin{subfigure}[b]{0.11\textwidth}
    \includegraphics[width=\textwidth]{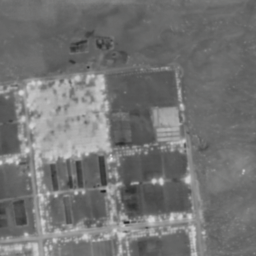}
    \vspace{-0.8\baselineskip}
\end{subfigure}
\rotatebox{90}{\scriptsize\enspace\hspace{1.5em}$W_S=512$}
\begin{subfigure}[b]{0.11\textwidth}
    \includegraphics[width=\textwidth]{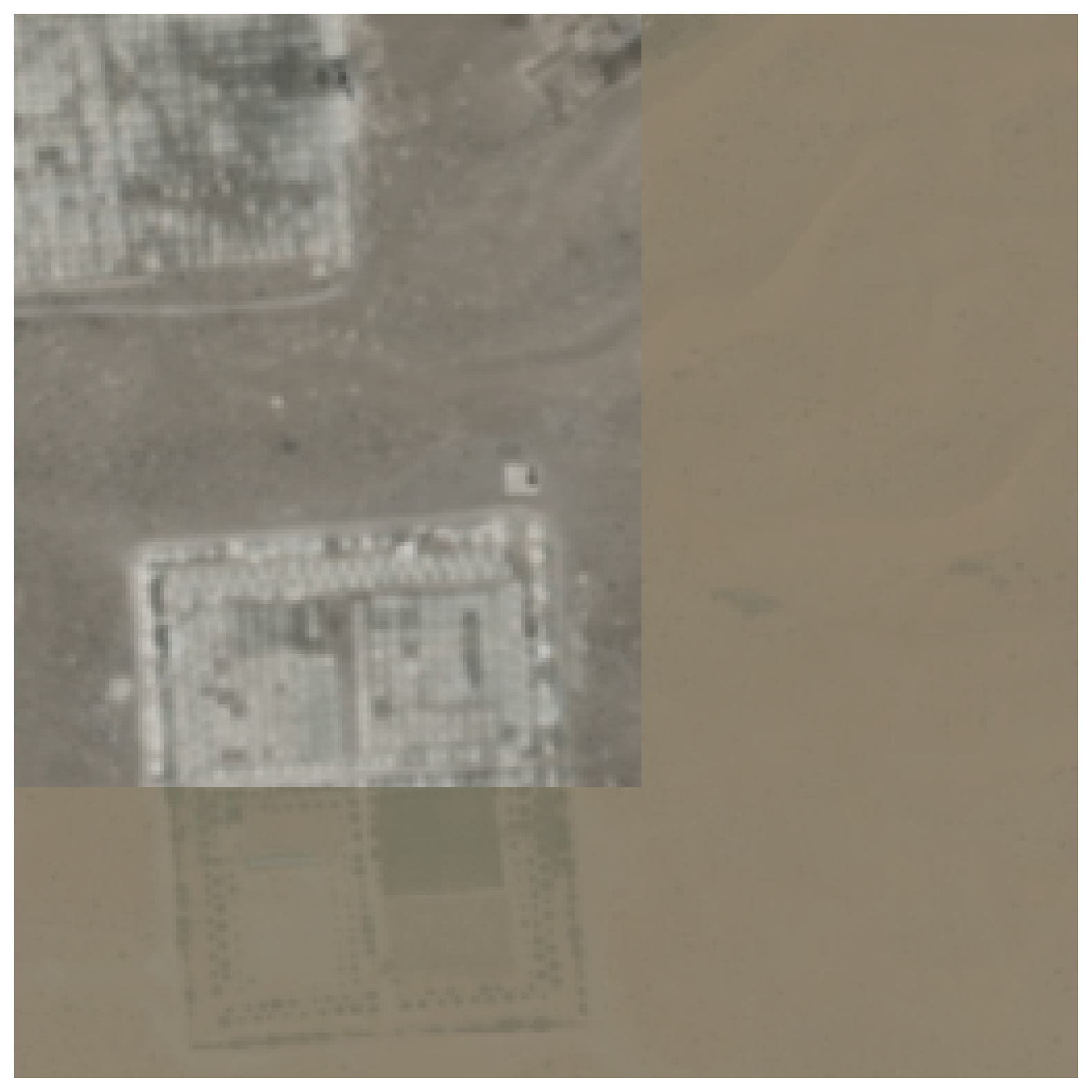}
    \vspace{-0.8\baselineskip}
\end{subfigure}
\begin{subfigure}[b]{0.11\textwidth}
    \includegraphics[width=\textwidth]{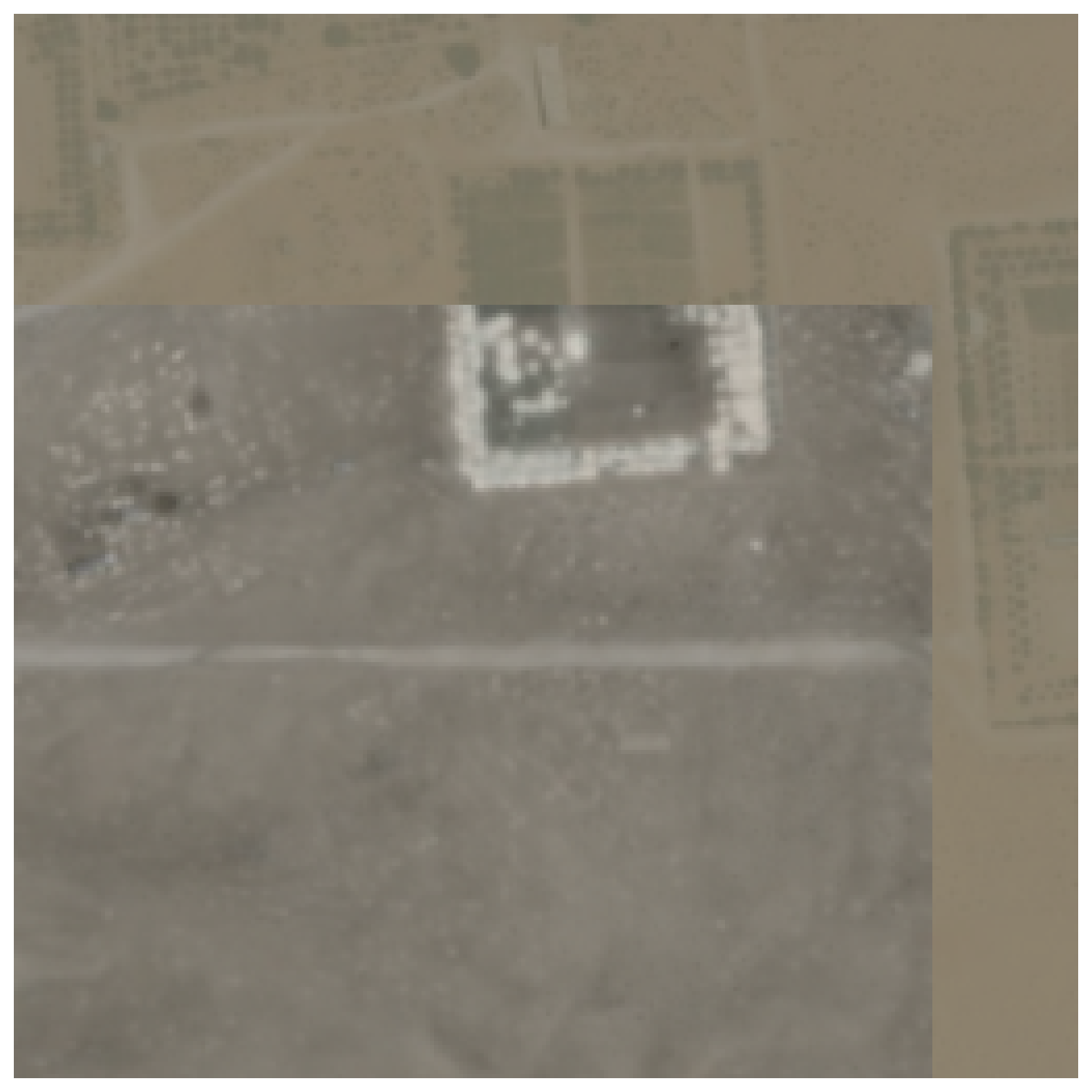}
    \vspace{-0.8\baselineskip}
\end{subfigure}
\begin{subfigure}[b]{0.11\textwidth}
    \includegraphics[width=\textwidth]{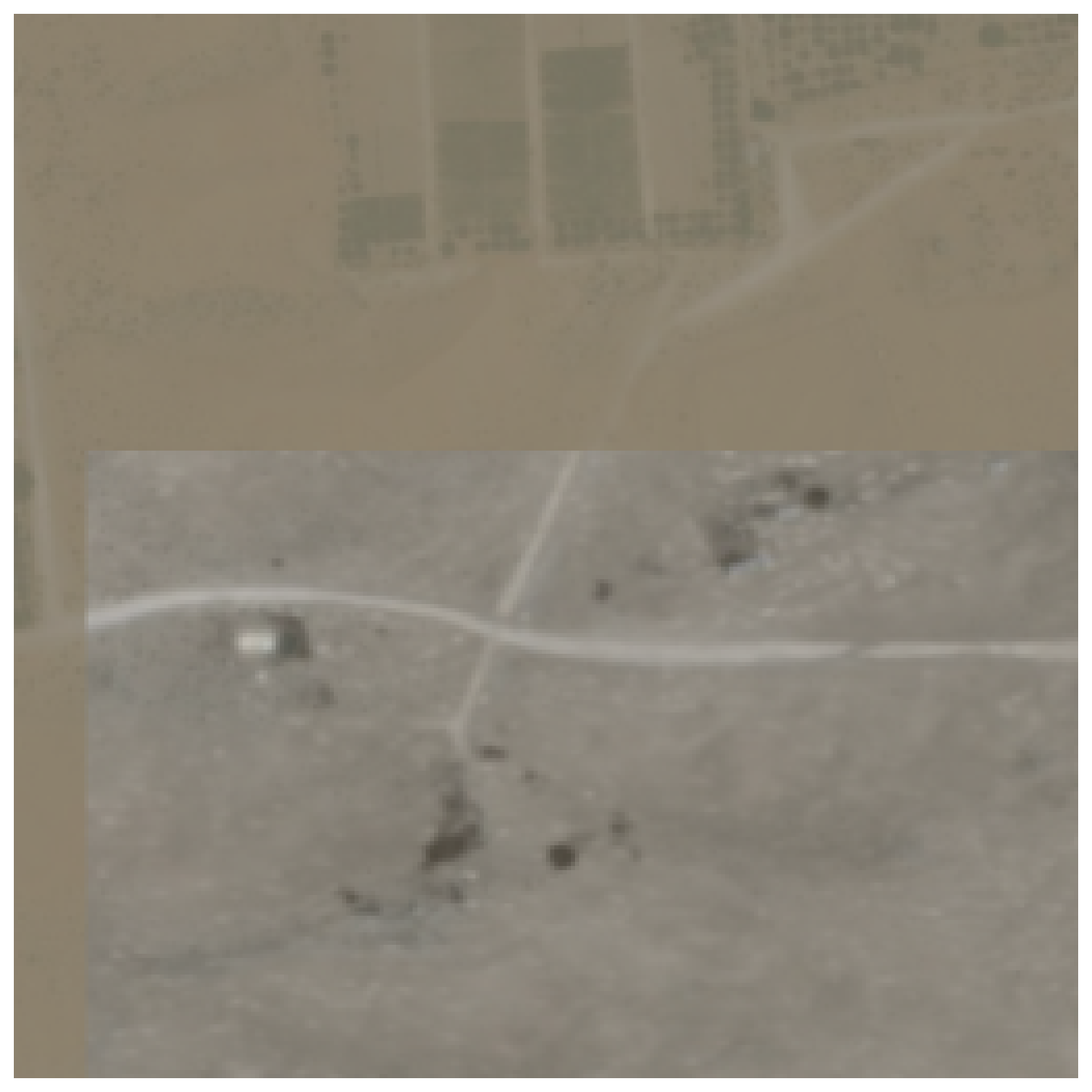}
    \vspace{-0.8\baselineskip}
\end{subfigure}
\begin{subfigure}[b]{0.11\textwidth}
    \includegraphics[width=\textwidth]{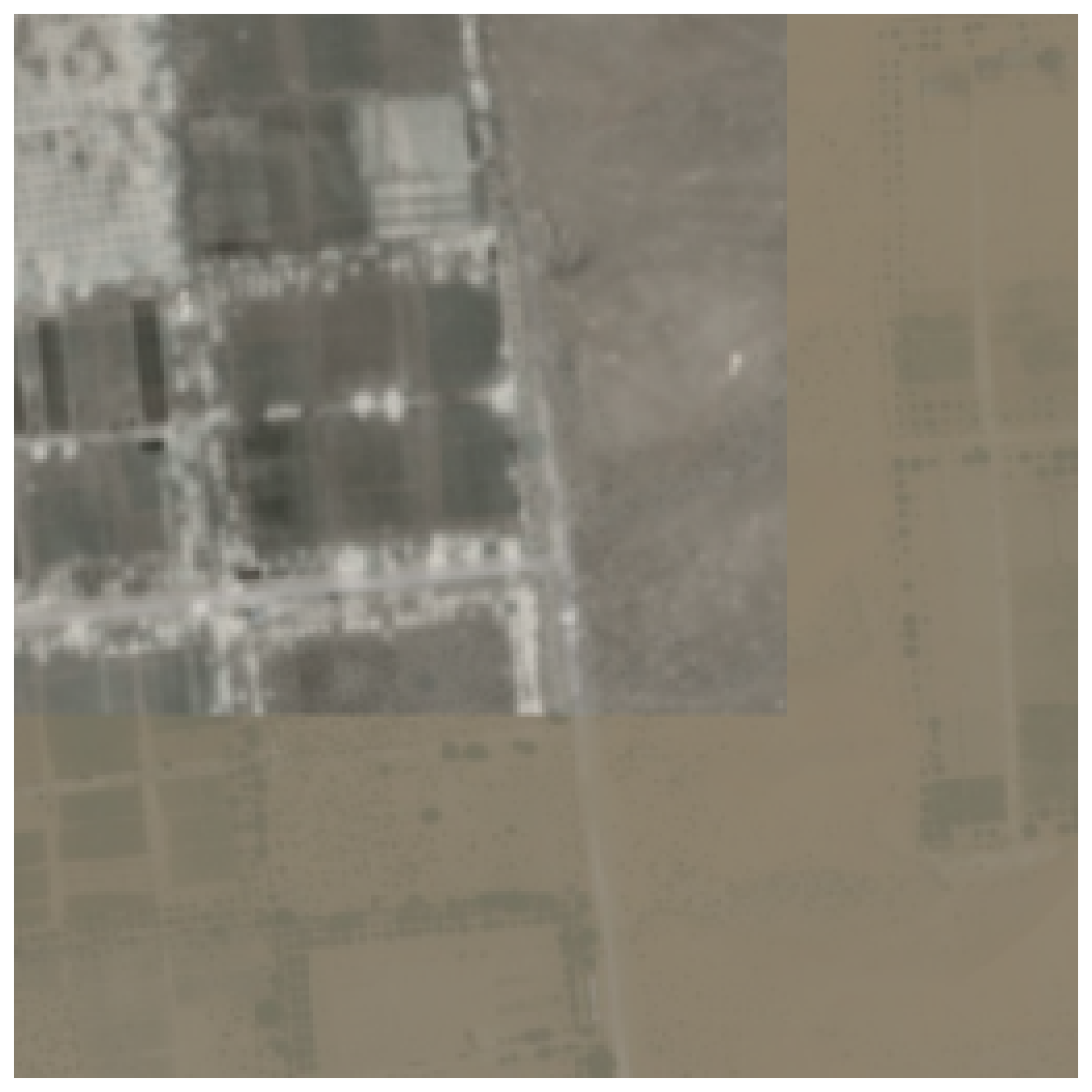}
    \vspace{-0.8\baselineskip}
\end{subfigure}
\rotatebox{90}{\scriptsize\enspace\hspace{1em}$W_S=1024$}
\begin{subfigure}[b]{0.11\textwidth}
    \includegraphics[width=\textwidth]{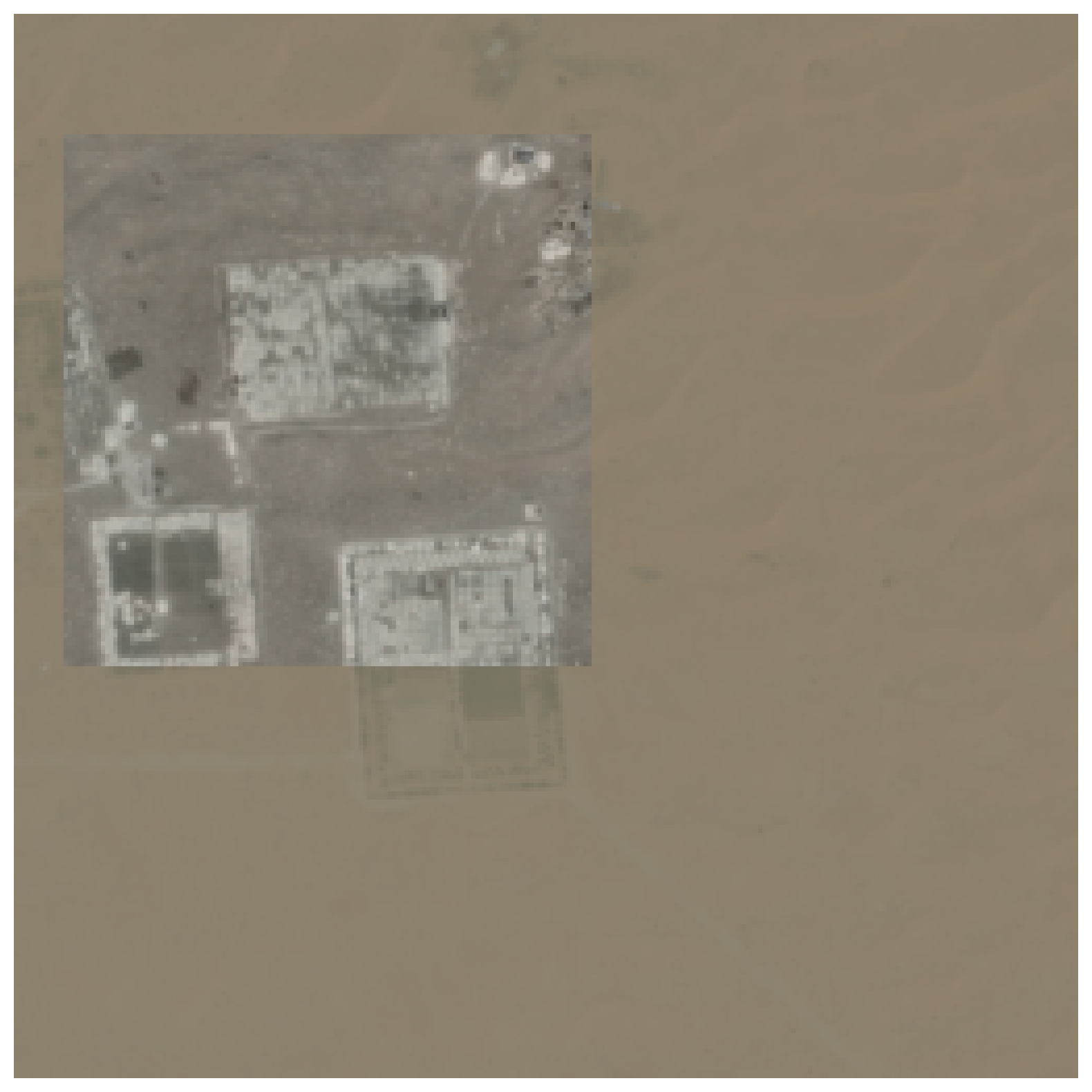}
    \vspace{-0.8\baselineskip}
\end{subfigure}
\begin{subfigure}[b]{0.11\textwidth}
    \includegraphics[width=\textwidth]{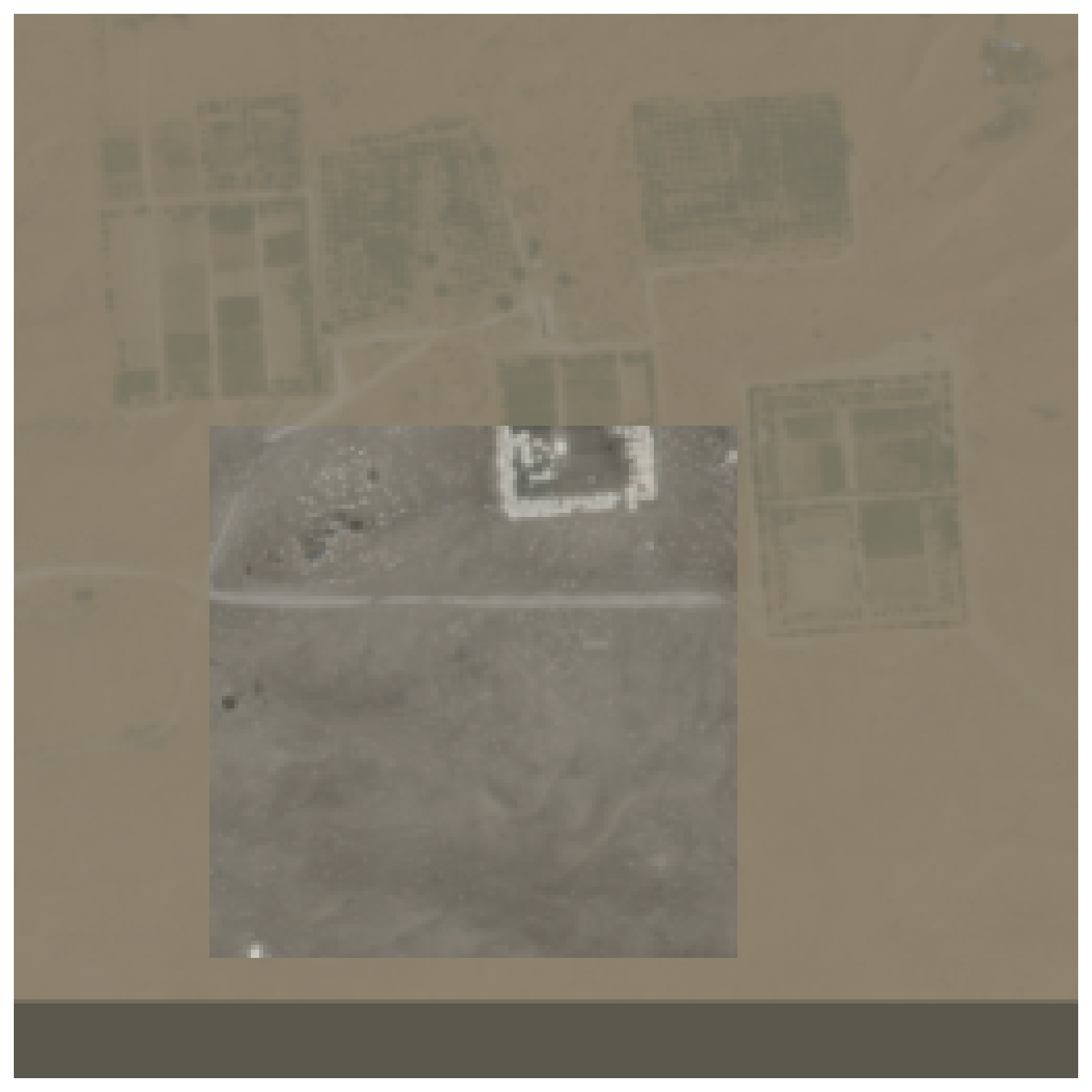}
    \vspace{-0.8\baselineskip}
\end{subfigure}
\begin{subfigure}[b]{0.11\textwidth}
    \includegraphics[width=\textwidth]{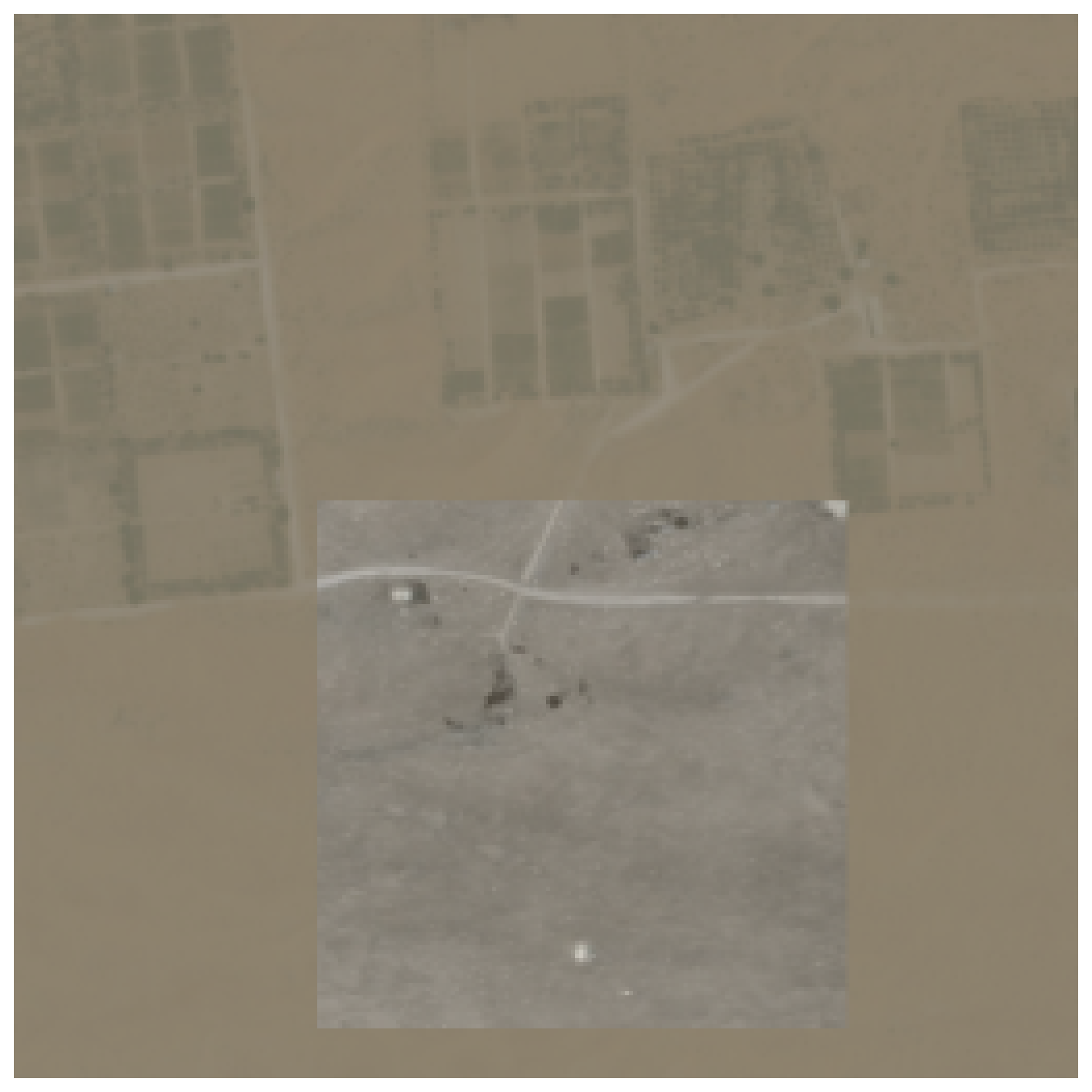}
    \vspace{-0.8\baselineskip}
\end{subfigure}
\begin{subfigure}[b]{0.11\textwidth}
    \includegraphics[width=\textwidth]{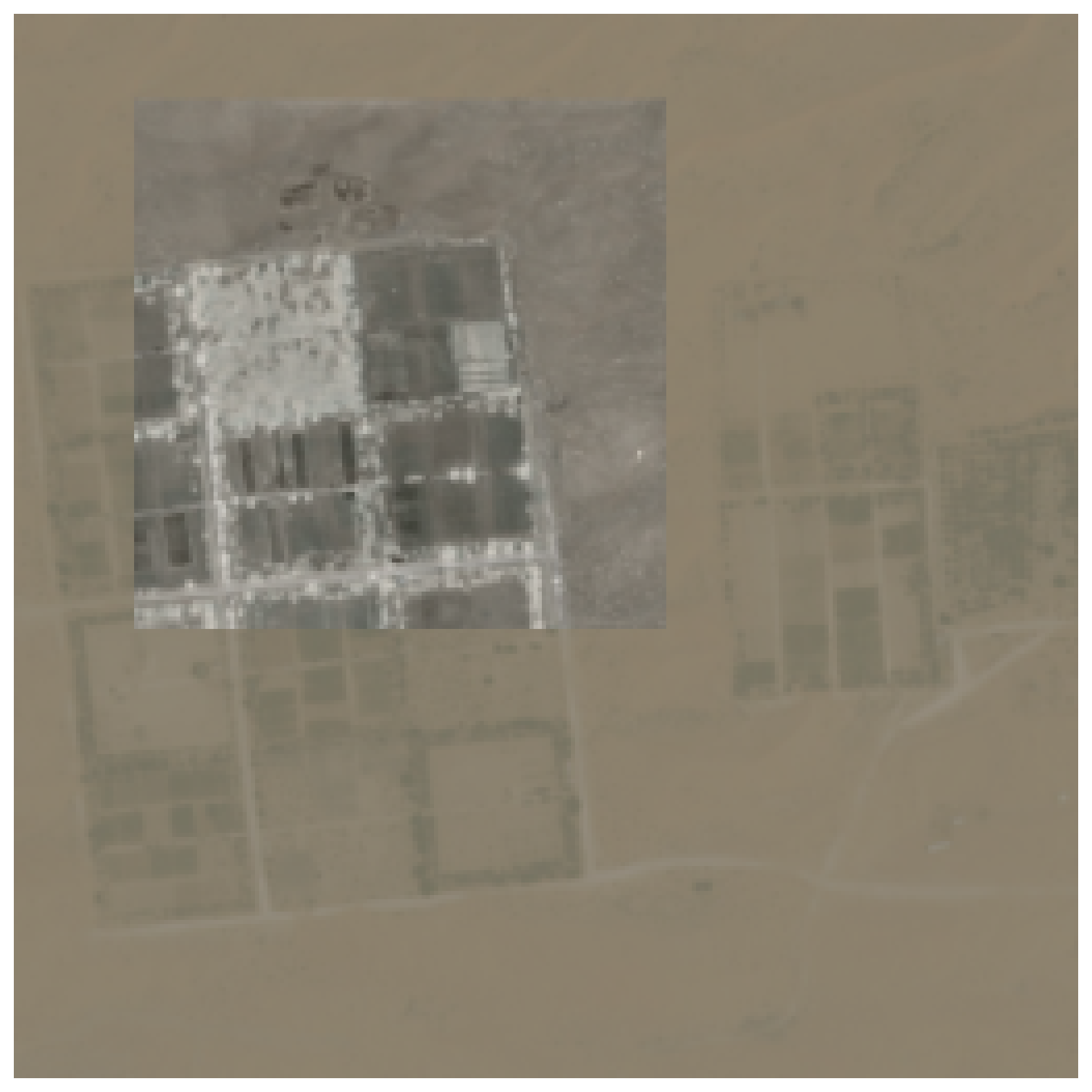}
    \vspace{-0.8\baselineskip}
\end{subfigure}
\rotatebox{90}{\scriptsize \enspace \hspace{1em}$W_S=1536$}
\begin{subfigure}[b]{0.11\textwidth}
    \includegraphics[width=\textwidth]{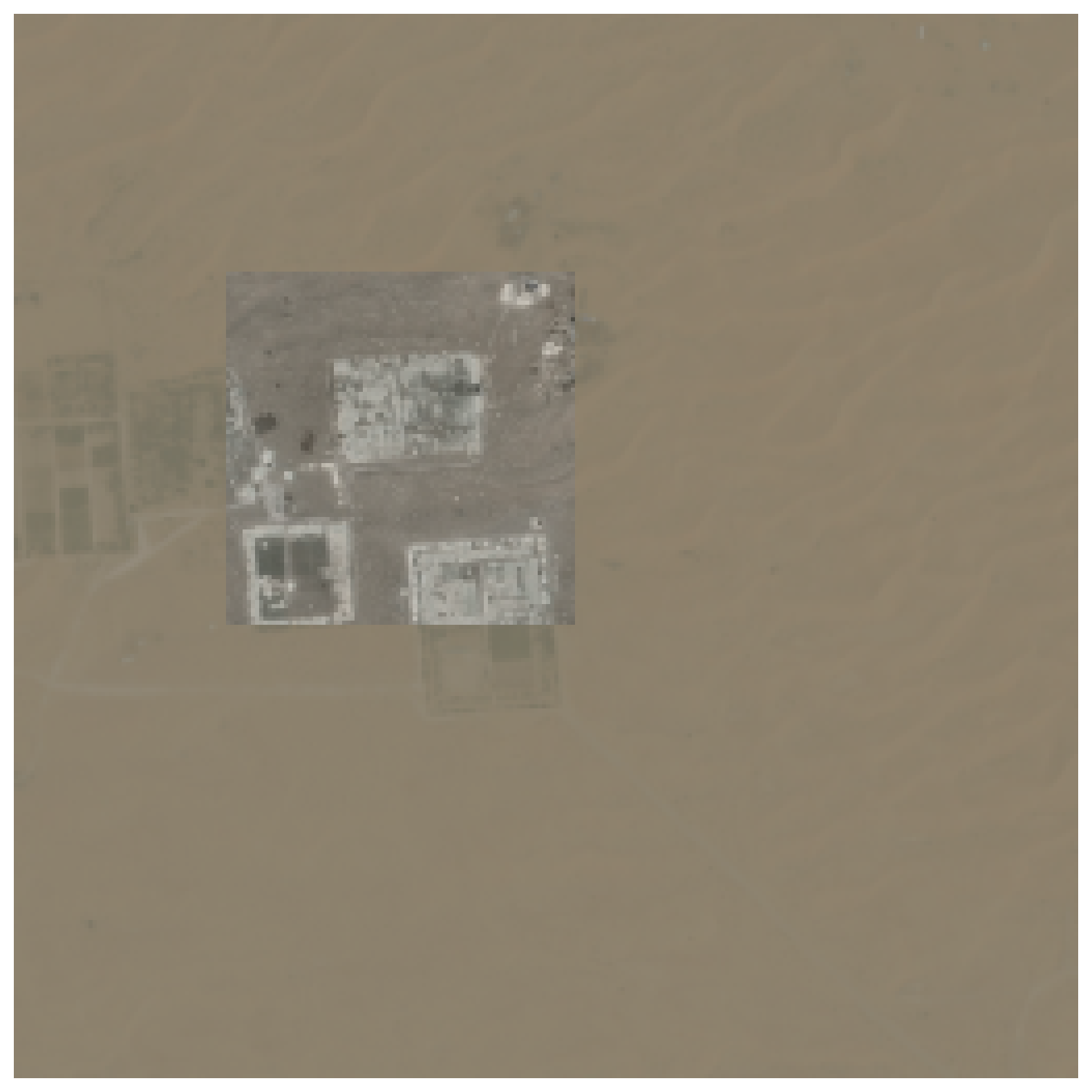}
    \vspace{-0.8\baselineskip}
\end{subfigure}
\begin{subfigure}[b]{0.11\textwidth}
    \includegraphics[width=\textwidth]{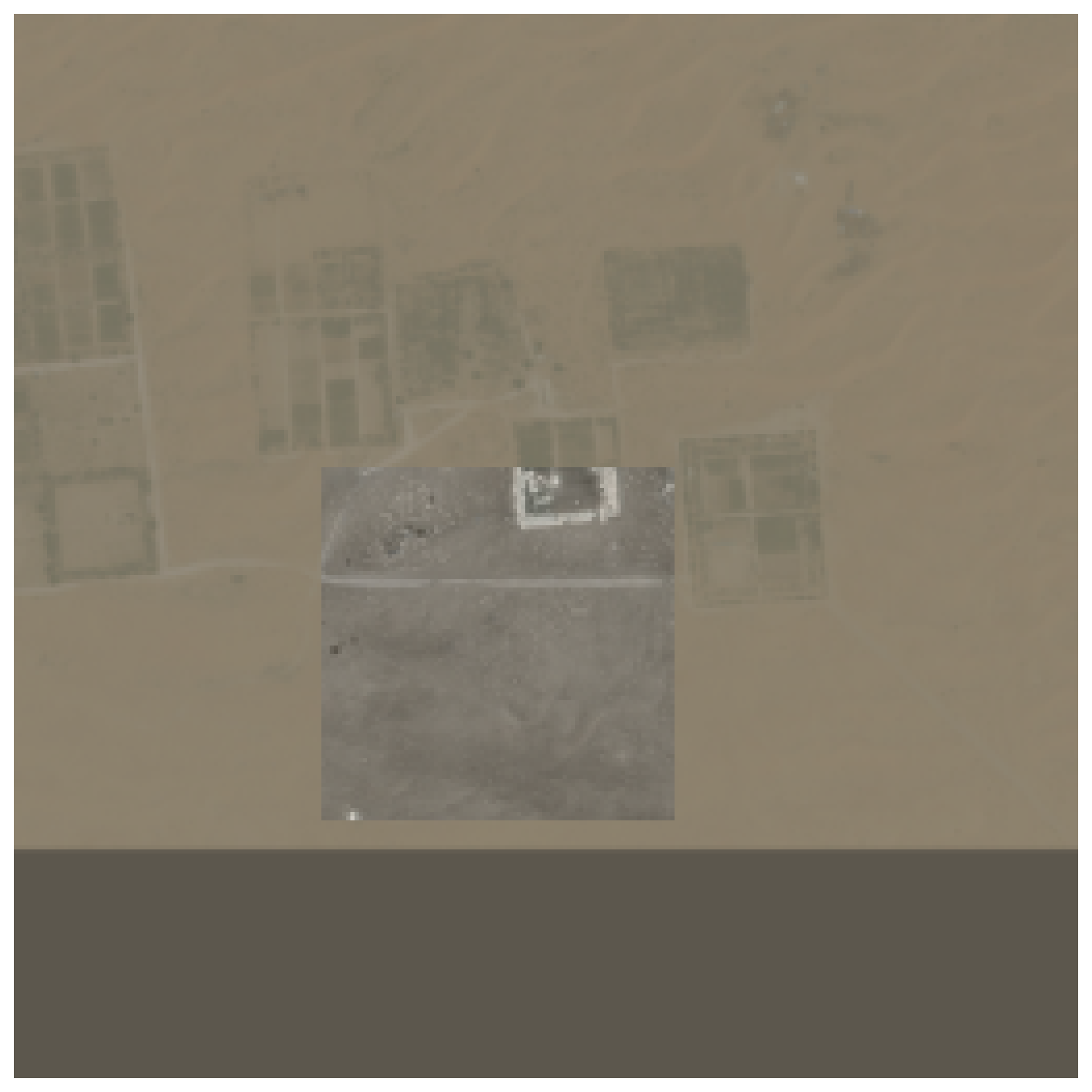}
    \vspace{-0.8\baselineskip}
\end{subfigure}
\begin{subfigure}[b]{0.11\textwidth}
    \includegraphics[width=\textwidth]{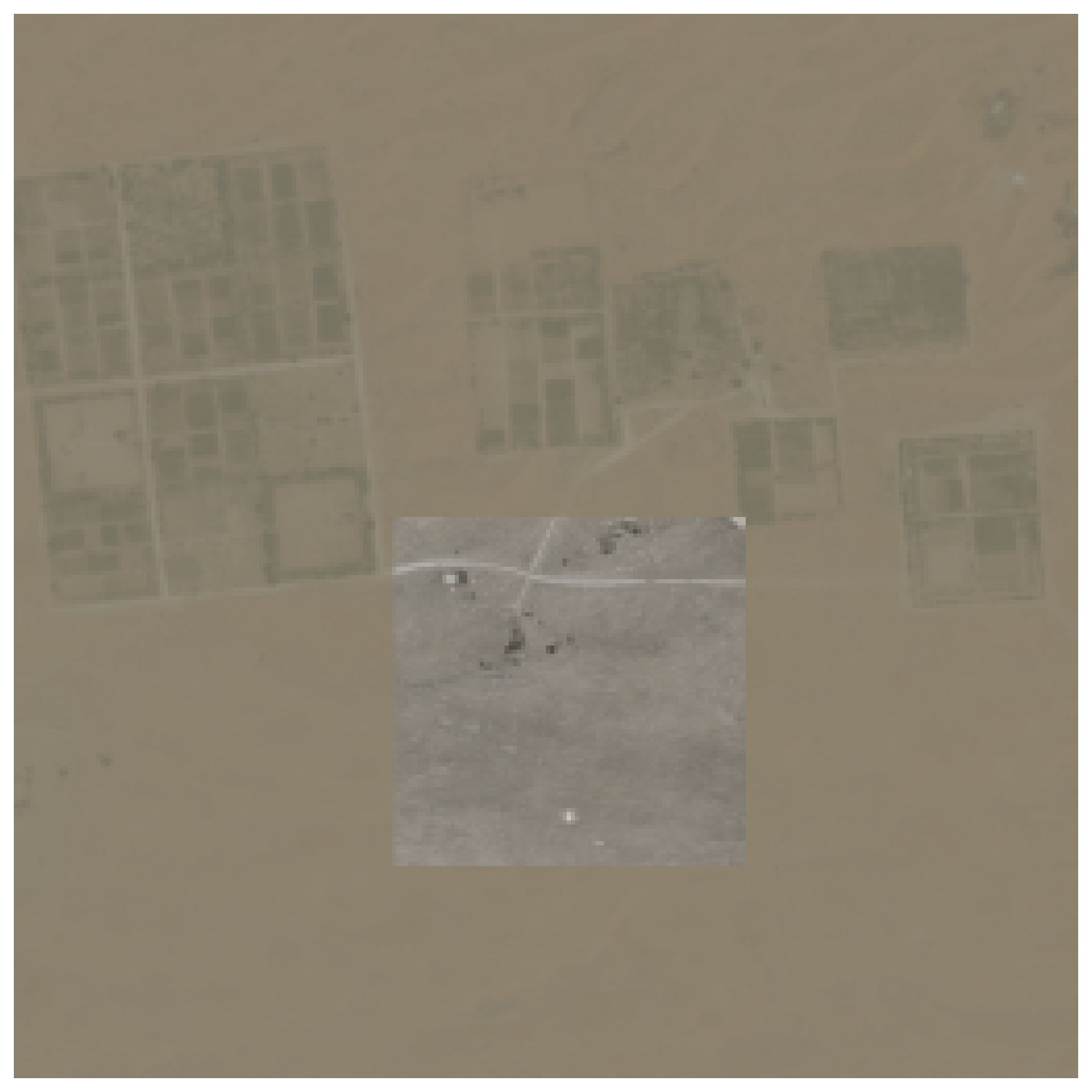}
    \vspace{-0.8\baselineskip}
\end{subfigure}
\begin{subfigure}[b]{0.11\textwidth}
    \includegraphics[width=\textwidth]{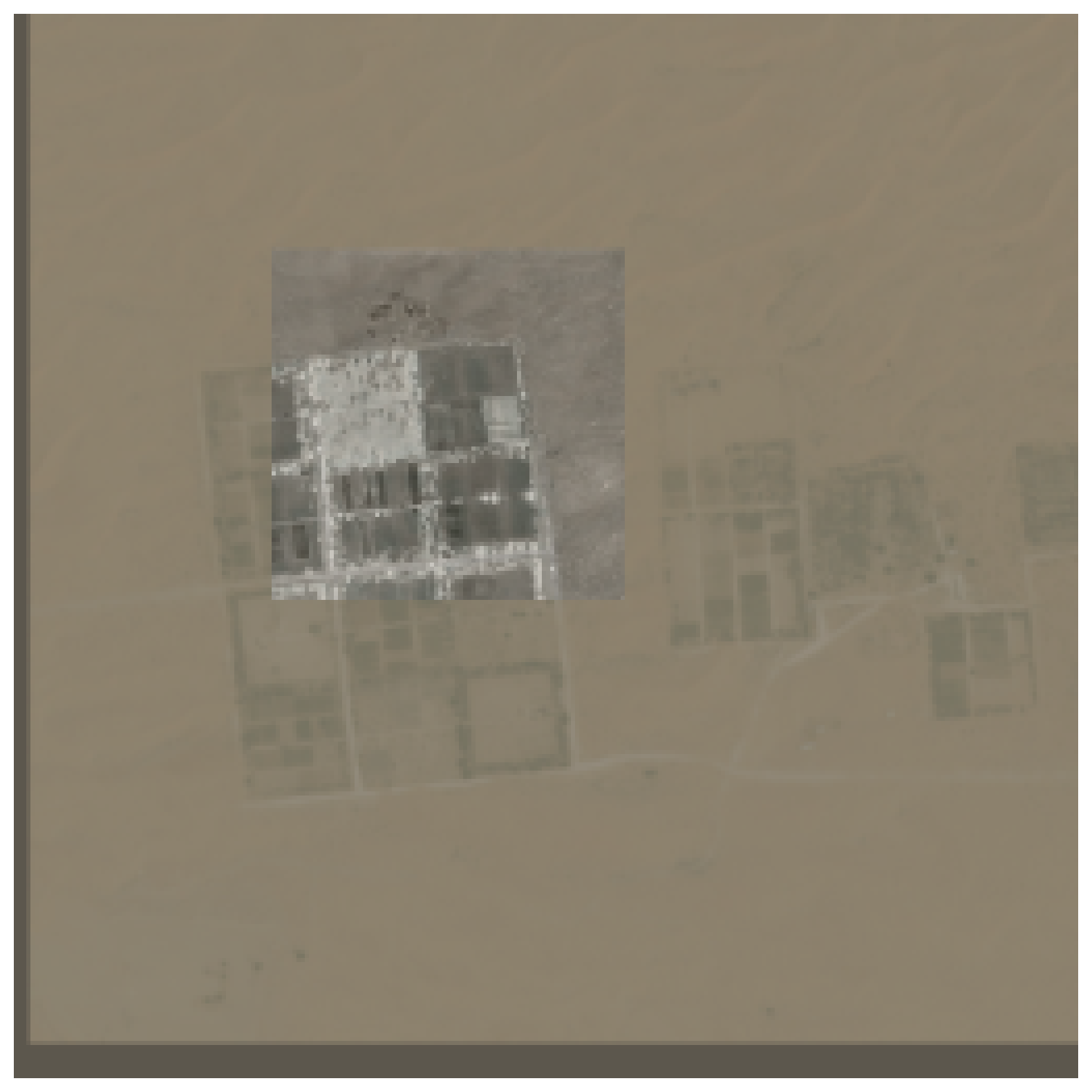}
    \vspace{-0.8\baselineskip}
\end{subfigure}
    \caption{The example images of Boson-nighttime dataset. The $1^\textrm{st}$ row and $2^\text{nd}$ row are input satellite and thermal images. The $3^\text{rd}$-$5^\text{th}$ rows are the ground truth overlap between satellite and thermal images with different $W_S$.}
    \label{example}
    \vspace{-15pt}
\end{figure}

For training and evaluation, our study utilizes the Boson-nighttime~\cite{stl} real-world dataset which contains $10,256$ train pairs, $13,011$ validation pairs, and $26,568$ test pairs of coupled satellite RGB and nadir-view 8-bit thermal imagery. We have expanded the dataset by augmenting the collection of satellite images without corresponding thermal images from $79,950$ to $163,344$ images, covering an area of $215.78~\si{km}^2$. This enhancement focuses on the desert and farm areas near the original dataset's sampling region, thereby incorporating a broader spectrum of geographical patterns. Additionally, the test region is then excluded from the generated data to ensure a robust evaluation of generalization performance. The thermal images in the dataset are captured between 9:00 PM and 4:00 AM, and they are aligned with an approx. spatial resolution of $1~\si{m/px}$. The thermal images are cropped to $W_T\times W_T$ pixels ($\si{px}$), where $W_T=512$. The satellite images \footnote{Bing RGB satellite imagery is sourced from Maxar: \url{https://www.bing.com/maps/aerial}} are cropped to $W_S\times W_S$. Fig.~\ref{example} shows the ground truth overlap between thermal images and satellite images with different $W_S$. For $W_S = 512/1024/1536$, the size ratios between thermal images and satellite images are $100\%$, $25\%$, and $11\%$.

\vspace{-10pt}
\subsection{Metrics} \label{metrics}
We deploy two accuracy metrics in our evaluation: Mean Average Corner Error (MACE) and Center Error (CE). MACE, extensively adopted in~\cite{detone2016deep, cao2022iterative, Cao_2023_CVPR}, measures the mean value of the average distances between the four corners of estimated and ground truth image alignments. Conversely, CE measures the mean value of the distances between the center points of predicted thermal image displacements and ground truth ones, thereby measuring geo-localization accuracy. 

In our experimental analysis, the maximum spatial distance between the center points of input thermal and satellite images $D_C$ emerges as a critical factor influencing estimation performance. Intuitively, a larger $D_C$ implies a greater translation from the center required for the four-corner displacement, which in turn becomes more challenging to predict accurately. To validate the robustness of our method, we cautiously ablate results across a spectrum of $D_C$, demonstrating our approach's capability under varying degrees of challenging translations.

\vspace{-5pt}
\subsection{Implementation Details}\label{imp}
For pre-processing, the resize side length $W_R$ is $256~\si{px}$. The training iteration numbers of the coarse alignment and refinement modules are $200000$ with a batch size of $16$. The AdamW optimizer~\cite{loshchilov2017decoupled} is employed for model training, utilizing a linear learning rate decay scheduler with warmup with the peak learning rate at $1\mathrm{e}{-4}$. The numbers of iterative updates $K_1$ and $K_2$ are both set to $6$. Depending on the setting, the correlation module's level is $2$ (for $W_S=512$) or $4$ (for $W_S=1024,1536$) with a search radius of $4$. For bounding box augmentation, $\Delta p_1, \Delta p_2$ is set to vary between $(-64, 64)$, $\Delta p_3$ is set within $[0, 64)$, and $\Delta p_4$ is $64$ by parameter tuning. For geometric noises, we extend the coverage of thermal images, apply corresponding data augmentations, and center crop the thermal images to avoid black padding on their boundary. Our models are developed using PyTorch. The inference speed is measured with one NVIDIA RTX-2080-Ti GPU.
\section{Results}~\label{sec:resutls}
\begin{table*}[]
    \centering
        \caption{Comparison of test MACE (m) between different homography estimation methods across different $D_C$. "Identity" indicates the error if no homography estimation is applied. If not specified, the methods are evaluated with $W_S=512$.}
    \begin{tabular}{lcccccc}
    \toprule
        Methods & $D_C=50~\si{m}$ & $D_C=64~\si{m}$& $D_C=128~\si{m}$& $D_C=256~\si{m}$& $D_C=512~\si{m}$ & Failure Rate\\
         \midrule
         \multicolumn{3}{l}{\textit{Traditional Keypoint Matching Methods}} \\
         Identity & 35.63 & 39.08 & 85.63 & 170.94 & 334.68 & - \\
         SIFT~\cite{SIFT} + RANSAC~\cite{ransac} &442.20  &654.77 & 547.29 &529.63 & 1650.46  & 99.6\%\\
        SIFT~\cite{SIFT} + MAGSAC++~\cite{magsac}& 512.60 &438.54 &529.46 &561.64 &693.03 & 99.7\%\\
        ORB~\cite{ORB} + RANSAC~\cite{ransac}& 720.80 & 733.69 & 733.94 &4614.84 & 975.83 & 82.6\%\\
         ORB~\cite{ORB} + MAGSAC++~\cite{magsac}&784.12 &558.51 &564.63 &524.99 &573.72 &82.9\%\\
        BRISK~\cite{6126542} + RANSAC~\cite{ransac}& 503.25 &771.21 &665.94 &974.80 &591.17 &95.5\%\\
         BRISK~\cite{6126542} + MAGSAC++~\cite{magsac}&536.52 &487.95 &722.76 &1948.99 &568.24 &95.6\%\\
         \midrule
         \multicolumn{3}{l}{\textit{Learned Keypoint Matching Methods}} \\
         R2D2~\cite{r2d2} + RANSAC~\cite{ransac}& 994.31&1160.73& 1160.21& 2400.31 & 2902.81 & 88.9\%\\
         LoFTR~\cite{sun2021loftr} + RANSAC~\cite{ransac}& 1123.74 &1697.33 &1317.69 &1269.71  & 2564.65 & 0\%\\
         \midrule
         \multicolumn{3}{l}{\textit{Deep Homography Estimation Methods}} \\
         DHN~\cite{detone2016deep} &16.78 &20.43 &77.68 &197.27 &457.23 & 0\%\\
         LocalTrans~\cite{shao2021localtrans} &33.31 &37.29 &86.04 &166.52 &338.21 & 0\%\\
         IHN~\cite{cao2022iterative} & 5.91 & 7.81 & 51.74 &190.93  &367.24 & 0\%\\
         Ours ($W_S=512$) & \textbf{4.24} & \textbf{4.93}  & 14.97  & 142.71  &347.50 & 0\% \\
         Ours ($W_S=1024$) & 4.92 & 5.31 & \textbf{6.03} & \textbf{9.22} & 86.74 & 0\%\\
         Ours ($W_S=1536$) & 6.50 & 7.04 & 7.27 & 16.78 & 16.42 & 0\%\\
         Ours ($W_S=1536$ + two stages) &7.51  &7.20 &7.51 &14.99 & \textbf{12.70} & 0\%\\
    \bottomrule
    \end{tabular}

    \label{baseline}
    \vspace{-10pt}
\end{table*}


\begin{table}[]
    \centering
        \caption{Comparison between different image-based matching methods and our estimation method when $D_C=512~\si{m}$.}
    \begin{tabular}{lcc}
    \toprule
        Methods & Test CE (\si{m}) & Latency (\si{ms}) \\\midrule
        \multicolumn{2}{l}{\textit{Image-based Matching Methods}} \\
         AnyLoc-VLAD-DINOv2~\cite{keetha2023anyloc}& 258.21  & 352404.03  \\
         STGL-NetVLAD-ResNet50~\cite{stl, netvlad} & 89.31 & 7180.0\\
         STGL-GeM-ResNet50~\cite{stl, 8382272} & 13.52 & 4918.9\\\midrule
         \multicolumn{2}{l}{\textit{Deep Homography Estimation Methods}} \\
         Ours ($W_S=1536$) & 15.90 & \textbf{35.2}\\
         Ours ($W_S=1536$ + two stages) & \textbf{12.12} & 63.9\\
    \bottomrule
    \end{tabular}
    \vspace{-15pt}
    \label{baseline2}
\end{table}

\vspace{-10pt}
In Sections~\ref{sec:baseline} and \ref{sec:ablation}, we assume that thermal images are aligned to the north, facilitated by an onboard compass and a gimbaled thermal camera. Subsequently, in Section~\ref{sec:robustness}, we broaden our analysis for geometric noises. 
\vspace{-10pt}
\subsection{Comparison with Baselines}\label{sec:baseline}
In the results detailed in Table~\ref{baseline}, we initiate the analysis by evaluating the efficacy of traditional keypoint matching methods, such as SIFT~\cite{SIFT}, ORB~\cite{ORB}, and BRISK~\cite{6126542}, integrated with outlier rejection methods like RANSAC~\cite{ransac} and MAGSAC++~\cite{magsac}. We also evaluate learned keypoint methods including R2D2~\cite{r2d2} trained on our dataset and LoFTR~\cite{sun2021loftr} with pretrained weights. These methods demonstrate a significantly high MACE alongside substantial failure rates (calculated by instances where the number of matching keypoints $\leq 10$). This underlines the challenges of keypoint matching inherent in complex satellite-thermal alignment.

Subsequently, our analysis compares our methods with various deep homography estimation frameworks, including DHN~\cite{detone2016deep}, LocalTrans~\cite{shao2021localtrans}, and IHN~\cite{cao2022iterative} (state-of-the-art method in real-time applications). These baselines with one-stage models are trained on the Boson-nighttime dataset. We report the baseline results considering $W_S=512$ as representative results since other $W_S$ show similar trends in our analysis. The results show the superior performance of our approach for satellite-thermal alignment and geo-localization. A notable observation from the data is the different performance preferences across varying $D_C$ distances: for $D_C=50~\si{m}$ and $D_C=64~\si{m}$, the optimal $W_S$ is $512$, while for mid-range distances of $D_C=128~\si{m}$ and $D_C=256~\si{m}$, using $W_S=1024$ leads to the best results. Additionally, for the longest distance of $D_C=512~\si{m}$, our novel two-stage method with $W_S=1536$ emerges as the most effective strategy. The findings indicate that for cases where $D_C\leq 256~\si{m}$, employing our one-stage method combined with a carefully chosen $W_S$ emerges as the most effective strategy. Further explanation of the correlation between $W_S$ and $D_C$ is in Section~\ref{sec:ablation}.

We find that our two-stage method fails to enhance performance for distances $D_C=50~\si{m}$, $64~\si{m}$, and $128~\si{m}$, instead leading to a decline in accuracy. Upon examining the visualized outcomes, we observe that for smaller distances ($D_C\leq 128~\si{m}$), the initial coarse alignment is sufficiently accurate, making the refinement module's excessive iterative updates introduce noise into the final predictions, thereby degrading performances. Nevertheless, our two-stage approach maintains an overall MACE of less than $15~\si{m}$ across all considered $D_C$, establishing robust baselines for this task. Notably, for achieving precise geo-localization at $D_C=512~\si{m}$, this two-stage strategy demonstrates the best performance, underscoring its effectiveness for large-scale search regions.

We also compare with image-based solutions~ (AnyLoc~\cite{keetha2023anyloc} and STGL~\cite{stl}) on accuracy and latency aspects in Table~\ref{baseline2}. The latency of image-based matching methods is calculated by $t_e\times (N_S + 1) + t_m$, where $t_e$ is feature extraction time per image, and $N_S=841$ is the number of database images centered within a $1024\times 1024$ area (while the complete images cover a $1536 \times 1536$ area) with a sampling stride of $35$~px following~\cite{stl}, and $t_m$ is the matching time per query. For AnyLoc, we directly apply the original DINOv2~\cite{oquab2023dinov2} weights and fit the  VLAD~\cite{VLAD} parameters using our training data. We observe a significant performance decline in AnyLoc, likely due to the domain gap between satellite and thermal imagery. STGL with GeM yields high accuracy but still suffers from high latency. Our method exhibits significant enhancements in both accuracy and latency compared to these existing image-based matching techniques. Notably, our one-stage and two-stage methods achieve latency reductions to just $7.2\%$ and $13.0\%$ of the latency of STGL-GeM-ResNet50.

\subsection{Ablation Study}\label{sec:ablation}
\begin{figure}
\includegraphics[width=0.9\linewidth]{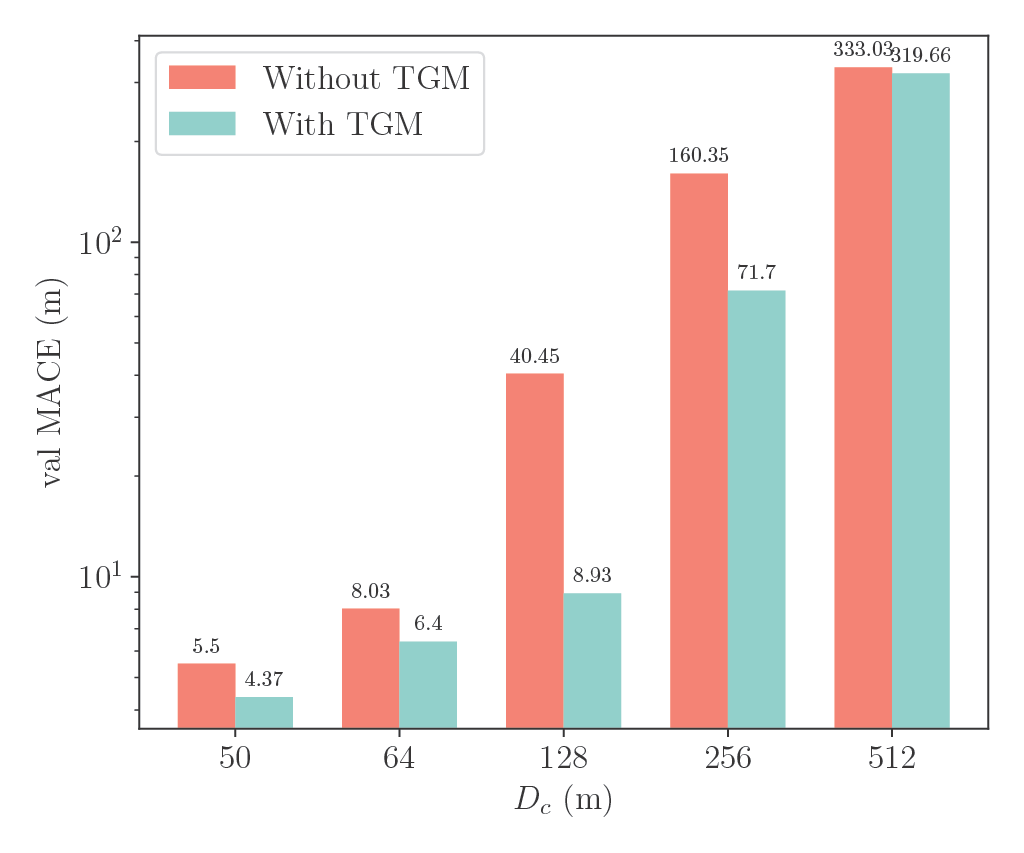}
\vspace{-10pt}
    \caption{Effectiveness of TGM in deep homography estimation across different $D_C$ when $W_S = 512$. Validation MACE (Val MACE) is plotted on a log scale.}\label{tgm}
    \vspace{-15pt}
\end{figure}

In this study (Figs.~\ref{tgm}-\ref{finetune}), we focus on the following questions
\begin{itemize}
    \item How does the incorporation of TGM affect the accuracy of homography estimation across varying $D_C$?
    \item Is the coarse alignment effective in achieving satisfactory localization accuracy for large $D_C$?
    \item Is the bounding box augmentation effective for fine-tuning the refinement module?
\end{itemize}

\subsubsection{Effectiveness of TGM}
Fig.~\ref{tgm} demonstrates the effectiveness of TGM in improving deep homography estimation over different spatial distances between centers ($D_C$) on the validation set. It showcases TGM's ability to enhance estimation accuracy by generating synthetic thermal images for satellite imagery that lacks paired thermal data. This consistent enhancement in image-based matching~\cite{stl} and deep homography estimation for satellite-thermal matching suggests TGM's potential applicability in additional computer vision tasks that do not have direct thermal imaging counterparts.

\subsubsection{Coarse alignment} 
\begin{figure}
\includegraphics[width=0.9\linewidth]{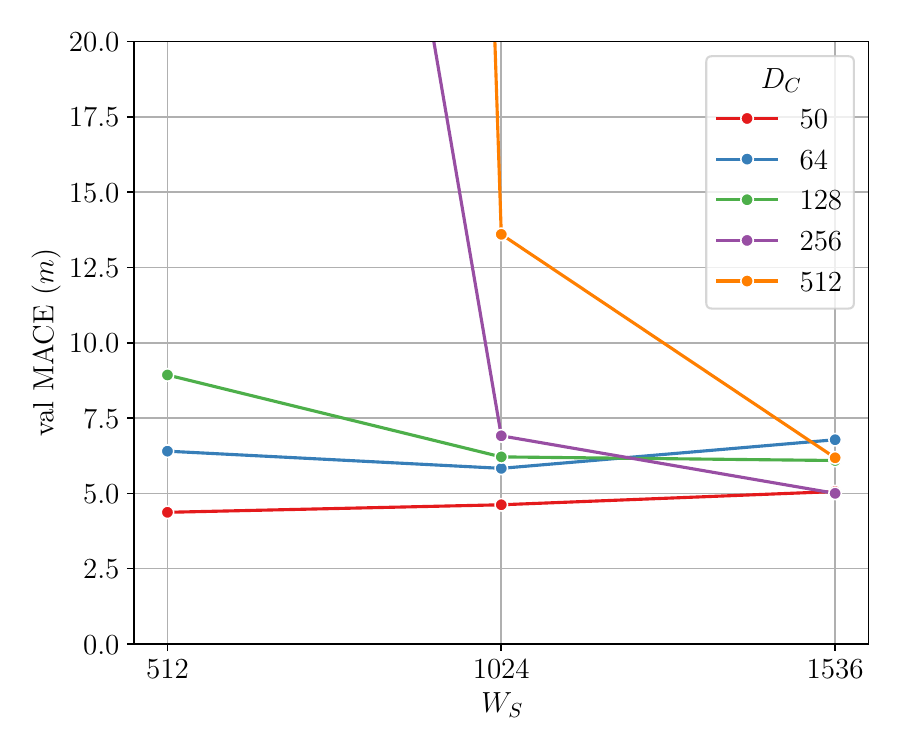}\vspace{-10pt}
    \caption{Coarse alignment under large-scale ($W_S=1536$), median-scale ($W_S=1024$), small-scale ($W_S=512$) satellite images with TGM.}\label{coarse}
    \vspace{-15pt}
\end{figure}

Fig.~\ref{coarse}
illustrates the correlation between validation MACE and $W_S$ across various $D_C$. The figure shows that as $W_S$ increases, the validation MACE for smaller translation distances ($D_C=50~\si{m}$ and $64~\si{m}$) slightly increases, suggesting a deterioration in alignment accuracy. In contrast, for larger translation distances ($D_C=128~\si{m}$, $256~\si{m}$, $512~\si{m}$), the validation MACE decreases, indicating improved alignment accuracy. The intuition is that an increase in $W_S$, without a corresponding adjustment in $W_R$, leads to a higher pixel-per-meter (ppm) ratio after image resizing. This increment in ppm ratio can negatively affect the alignment accuracy. Conversely, a larger $W_S$ enhances alignment accuracy for greater translation ($D_C$), especially for $W_S=1536$ and $D_C=512~\si{m}$. In these cases, a larger $W_S$ ensures the full coverage of the thermal image, which is crucial for accurately calculating correlation volumes $\mathbf{C}$.

\subsubsection{Effectiveness of Bounding Box Augmentation}\label{abaug} 
\begin{figure}[]
\smallskip
\smallskip
    \centering
\rotatebox{90}{\scriptsize\hspace{1.75em}w/o bbox aug}
\begin{subfigure}[b]{0.11\textwidth}
    \includegraphics[width=\textwidth]{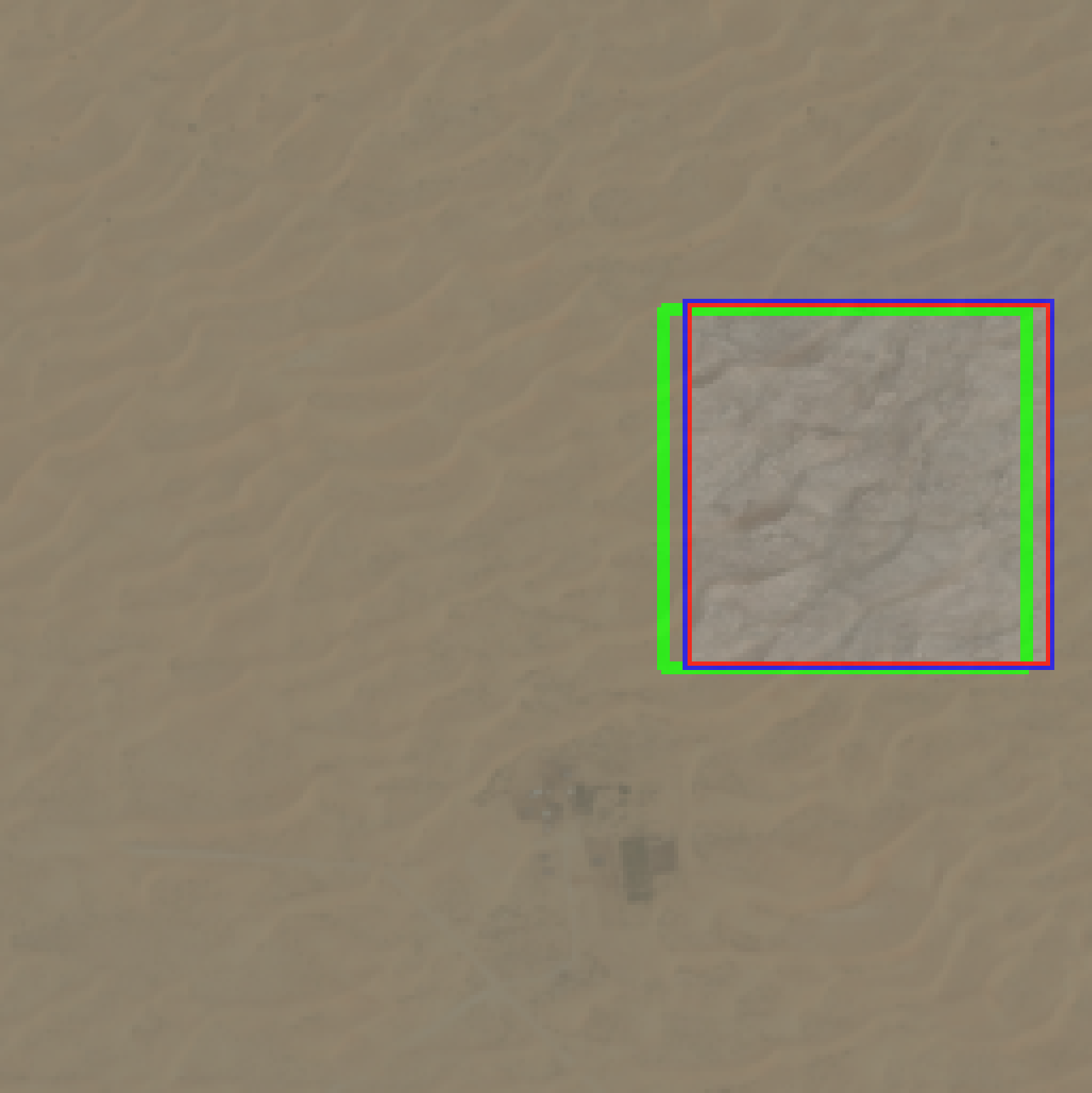}
    \vspace{-0.8\baselineskip}
\end{subfigure}
\begin{subfigure}[b]{0.11\textwidth}
    \includegraphics[width=\textwidth]{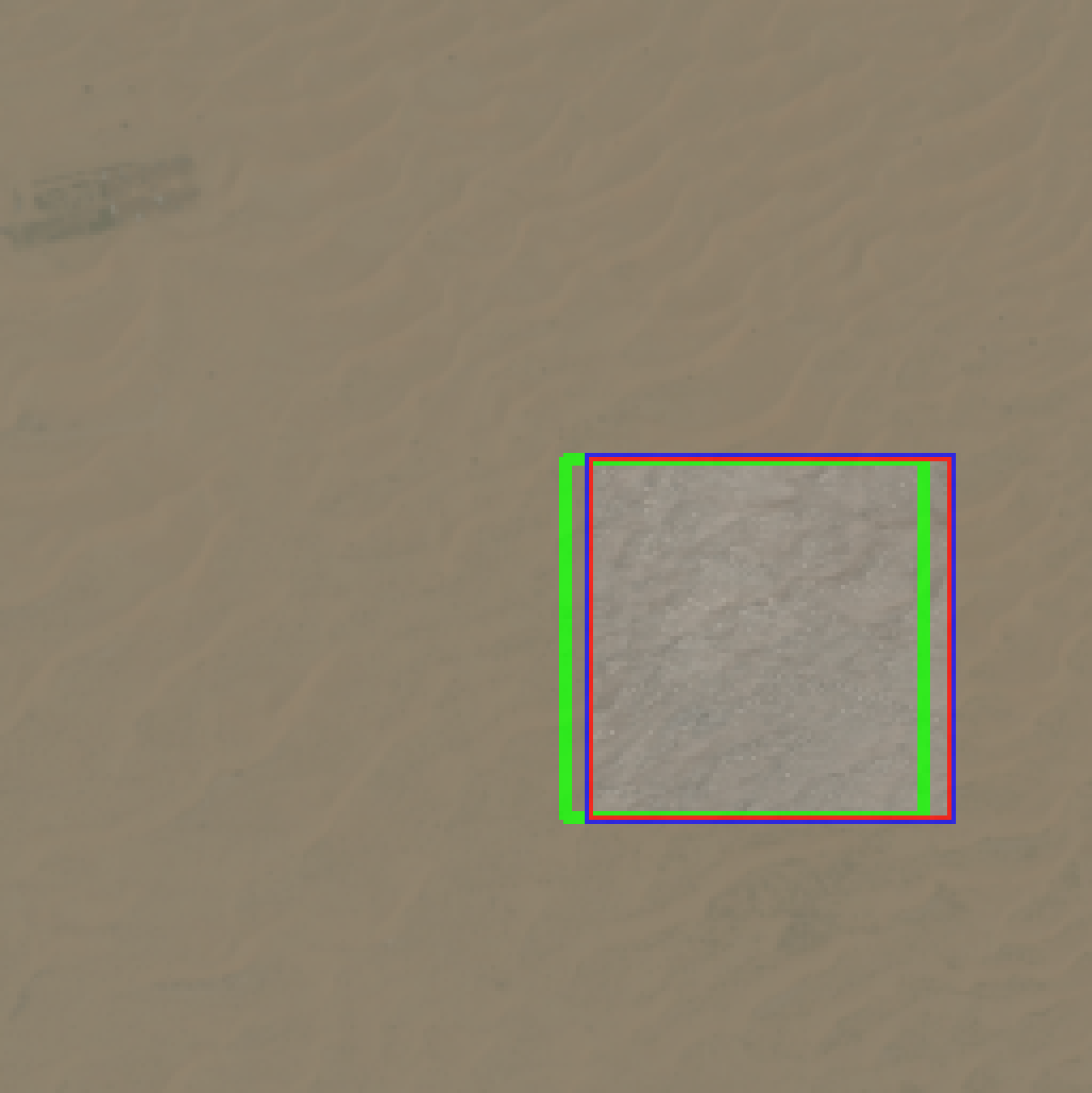}
    \vspace{-0.8\baselineskip}
\end{subfigure}
\begin{subfigure}[b]{0.11\textwidth}
    \includegraphics[width=\textwidth]{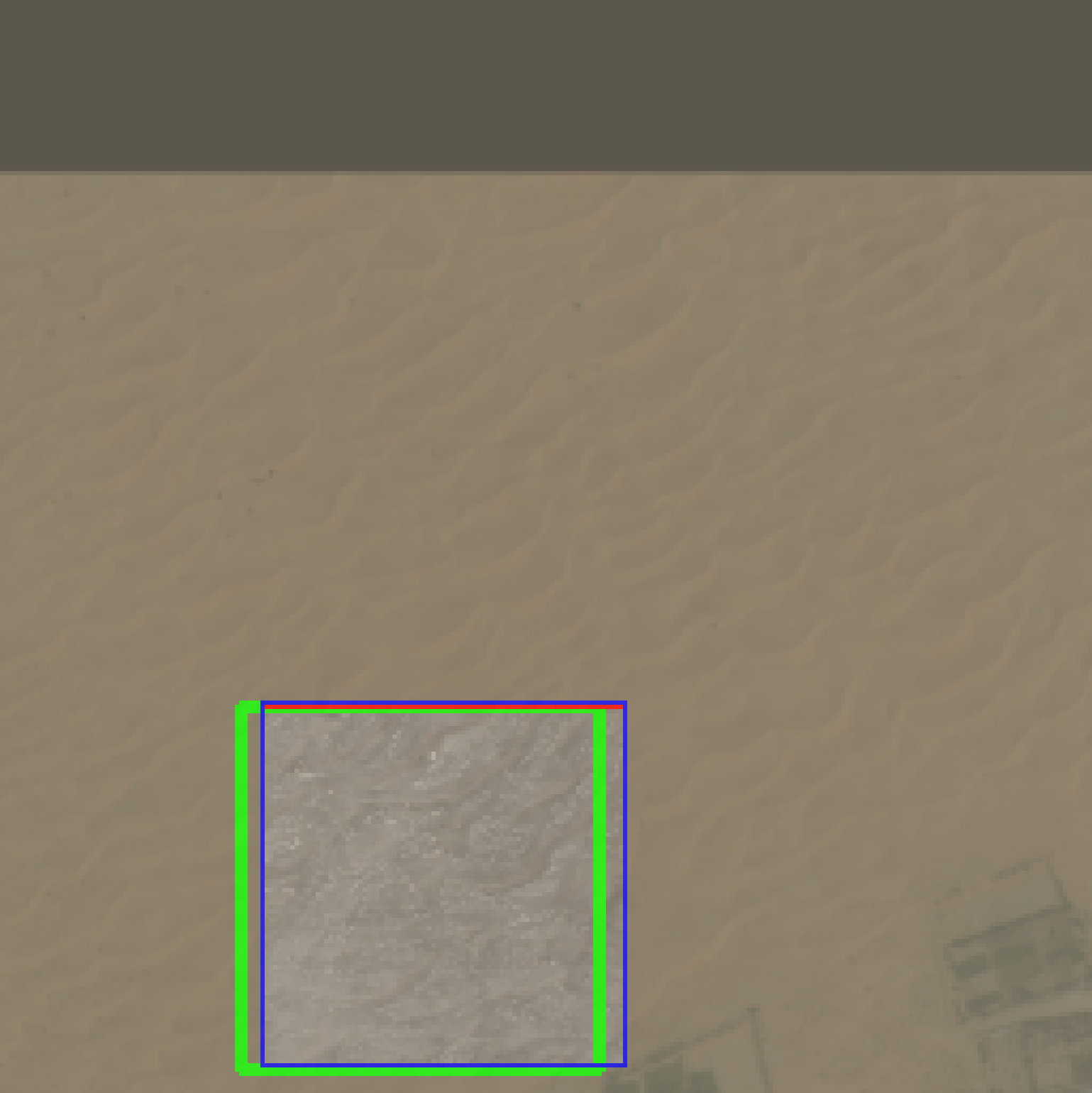}
    \vspace{-0.8\baselineskip}
\end{subfigure}
\begin{subfigure}[b]{0.11\textwidth}
    \includegraphics[width=\textwidth]{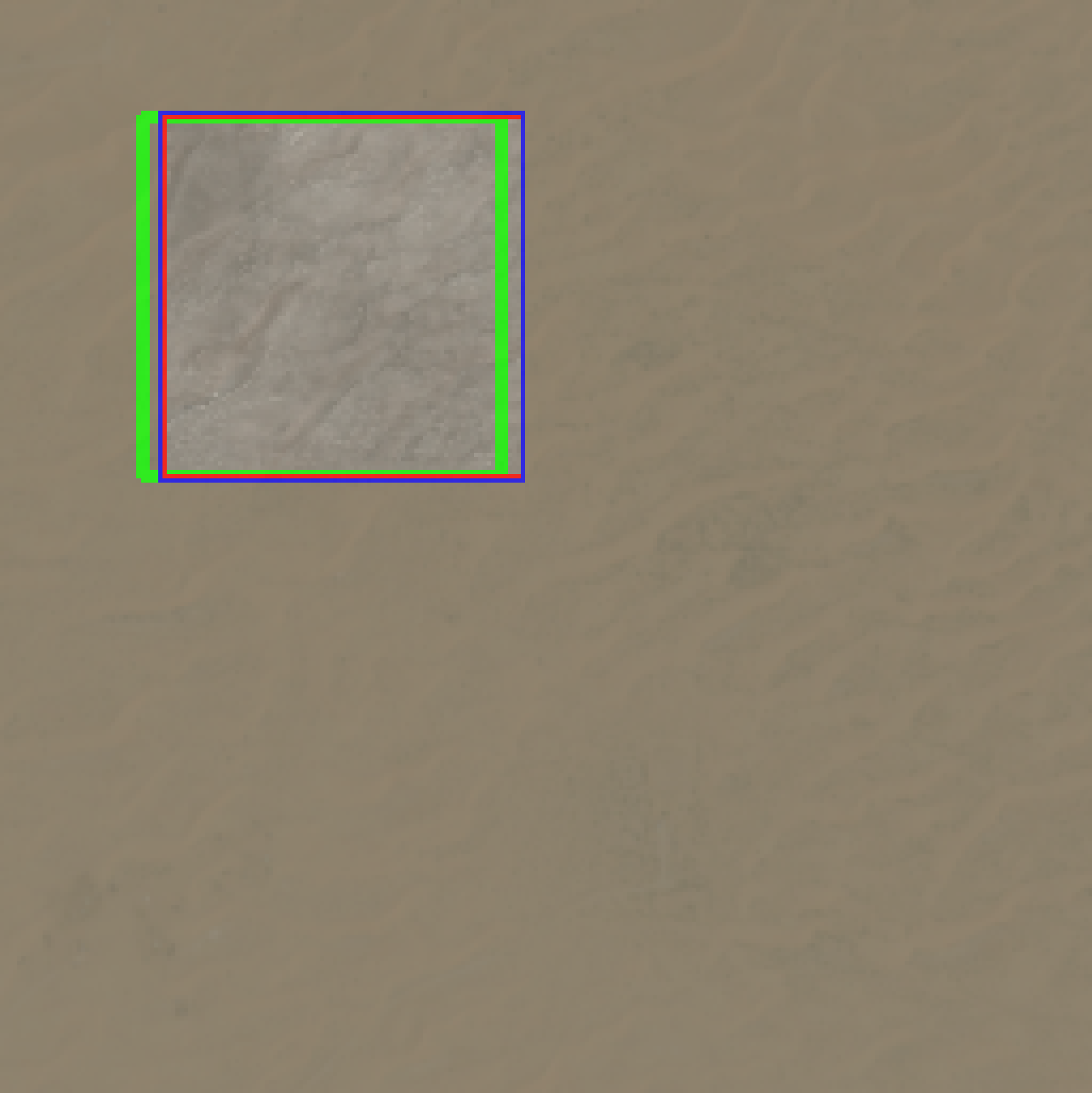}
    \vspace{-0.8\baselineskip}
\end{subfigure}

\rotatebox{90}{\scriptsize\hspace{2em}w/ bbox exp}
\begin{subfigure}[b]{0.11\textwidth}
    \includegraphics[width=\textwidth]{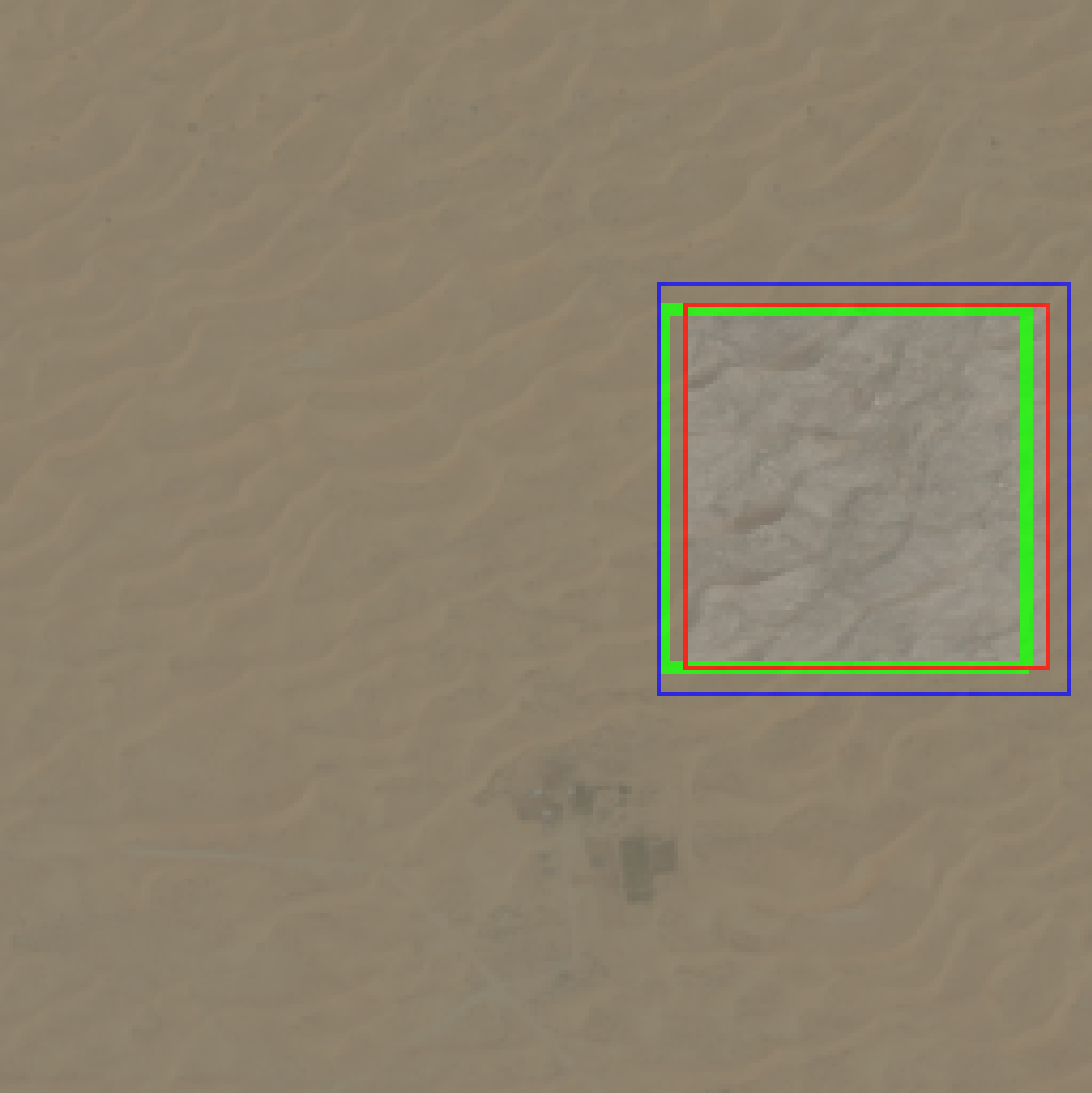}
    \vspace{-0.8\baselineskip}
\end{subfigure}
\begin{subfigure}[b]{0.11\textwidth}
    \includegraphics[width=\textwidth]{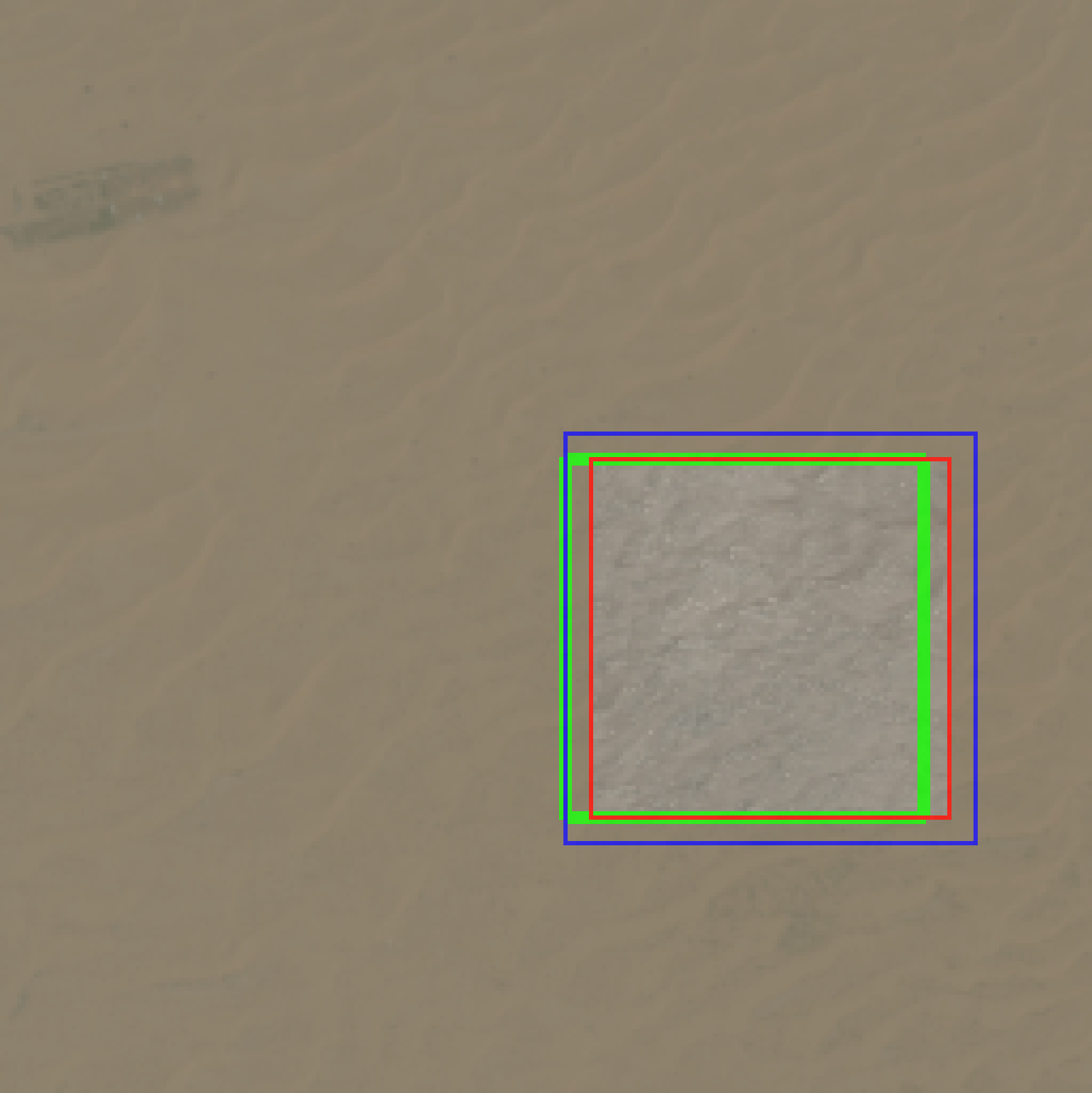}
    \vspace{-0.8\baselineskip}
\end{subfigure}
\begin{subfigure}[b]{0.11\textwidth}
    \includegraphics[width=\textwidth]{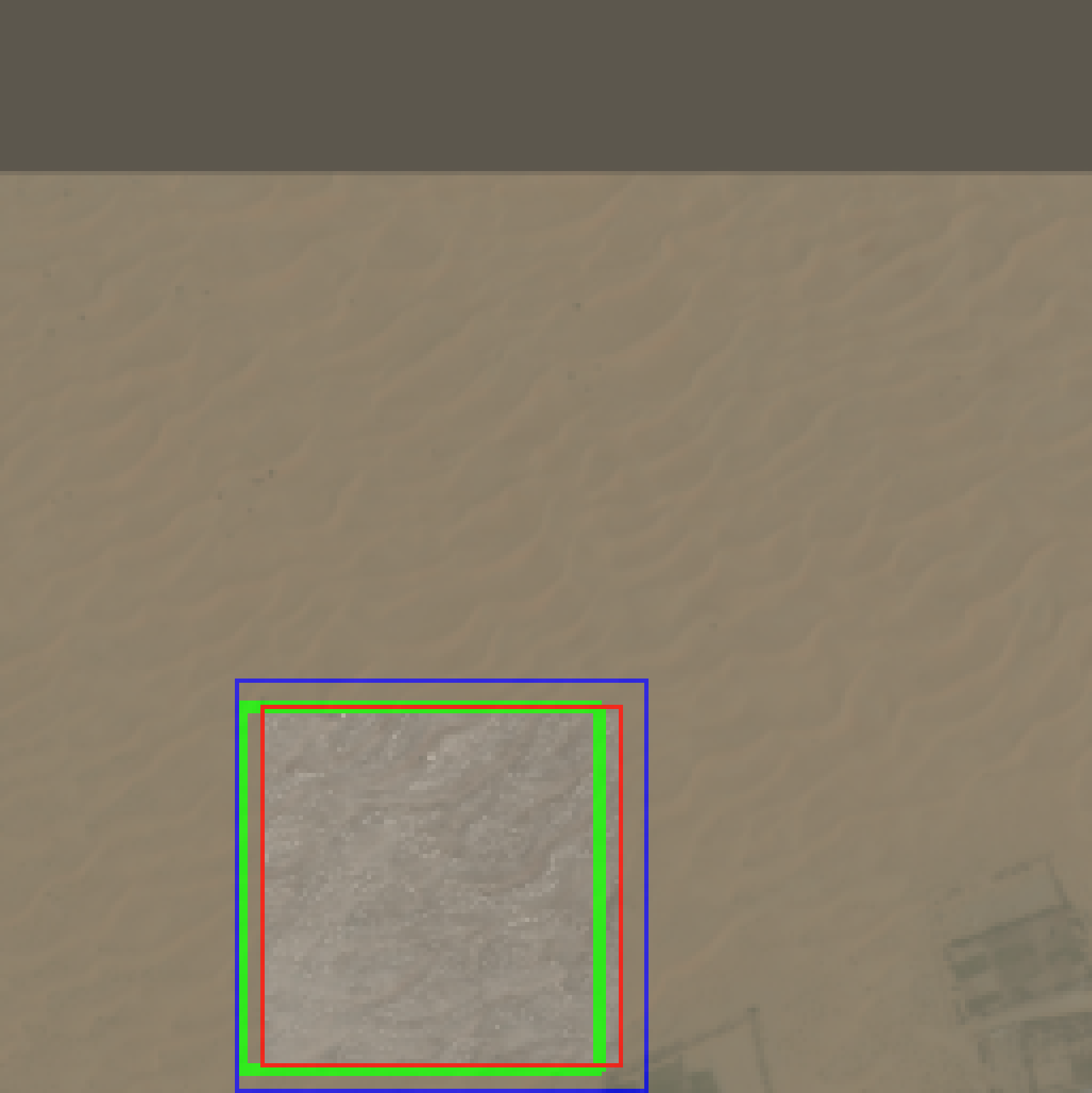}
    \vspace{-0.8\baselineskip}
\end{subfigure}
\begin{subfigure}[b]{0.11\textwidth}
    \includegraphics[width=\textwidth]{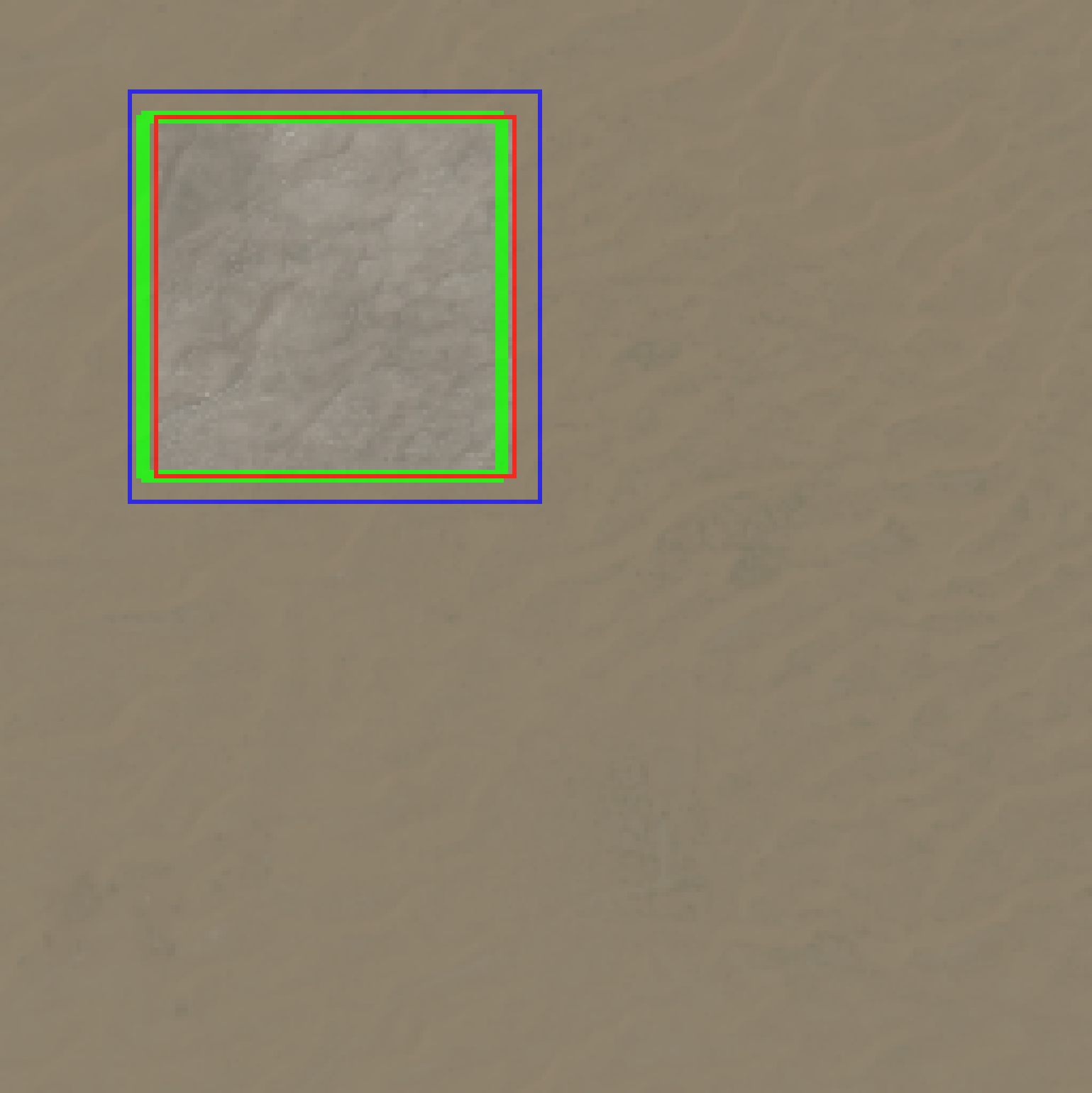}
    \vspace{-0.8\baselineskip}
\end{subfigure}

\rotatebox{90}{\scriptsize\hspace{1.75em}w/ bbox aug}
\begin{subfigure}[b]{0.11\textwidth}
    \includegraphics[width=\textwidth]{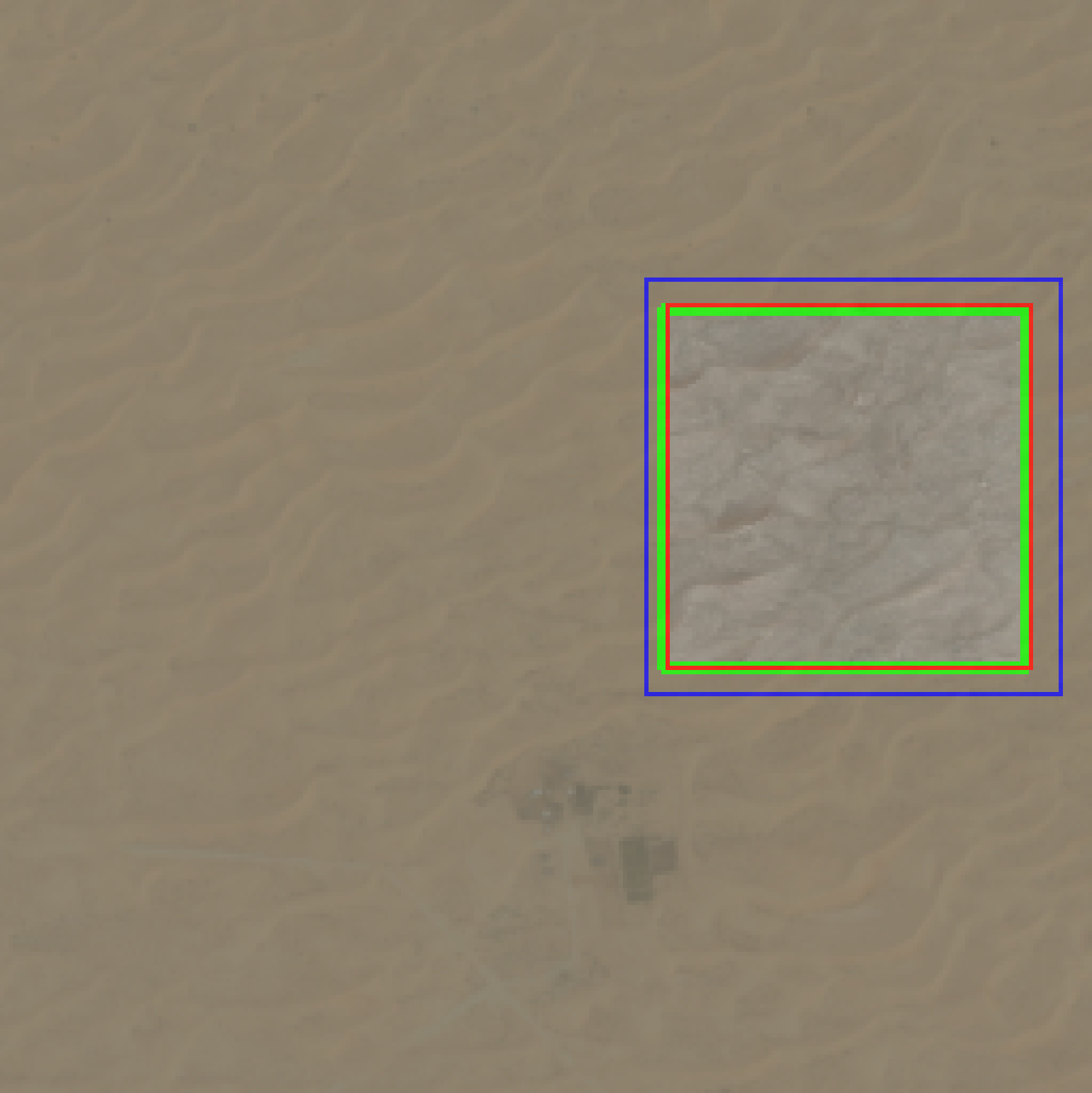}
    \vspace{-0.8\baselineskip}
\end{subfigure}
\begin{subfigure}[b]{0.11\textwidth}
    \includegraphics[width=\textwidth]{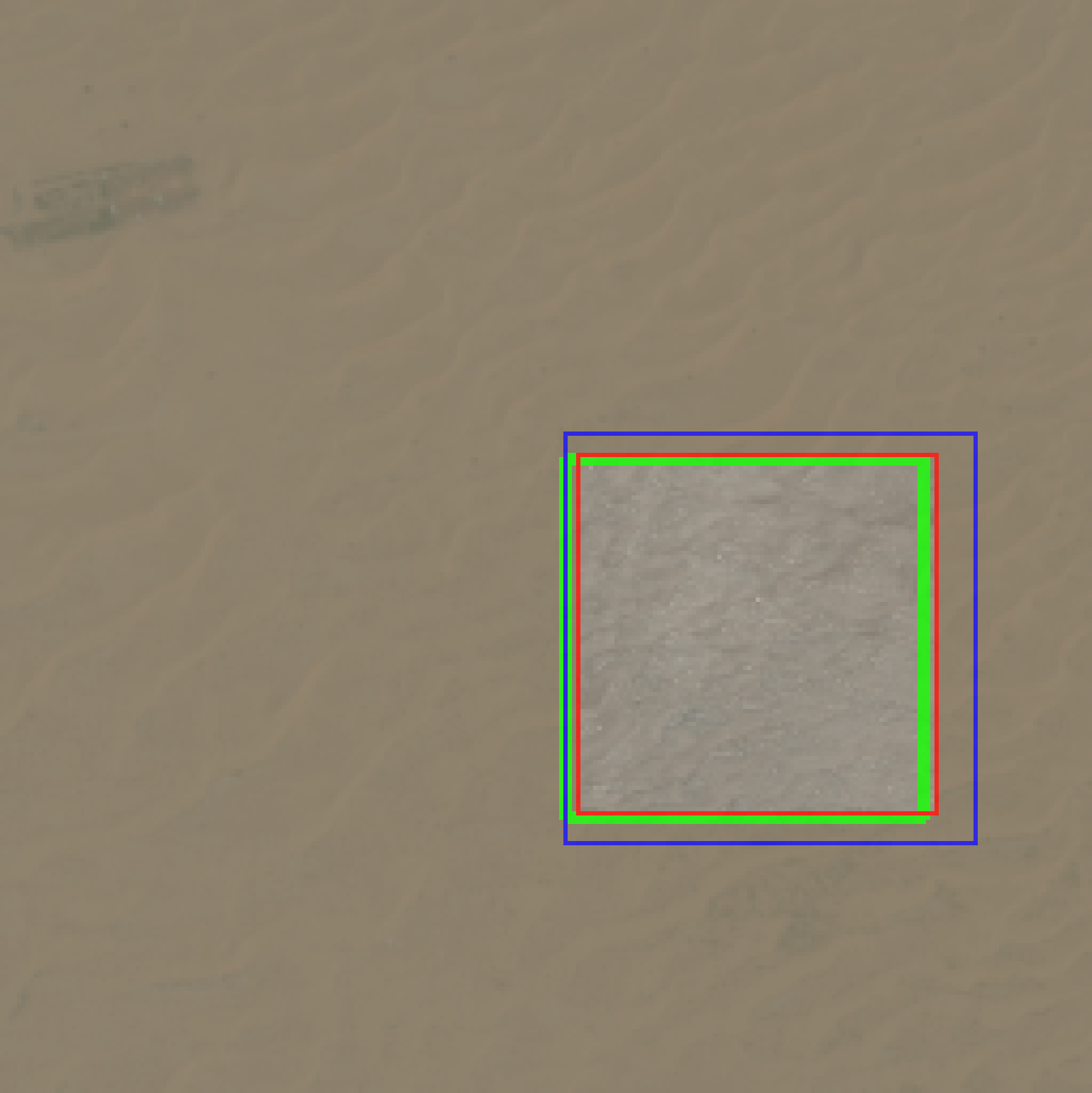}
    \vspace{-0.8\baselineskip}
\end{subfigure}
\begin{subfigure}[b]{0.11\textwidth}
    \includegraphics[width=\textwidth]{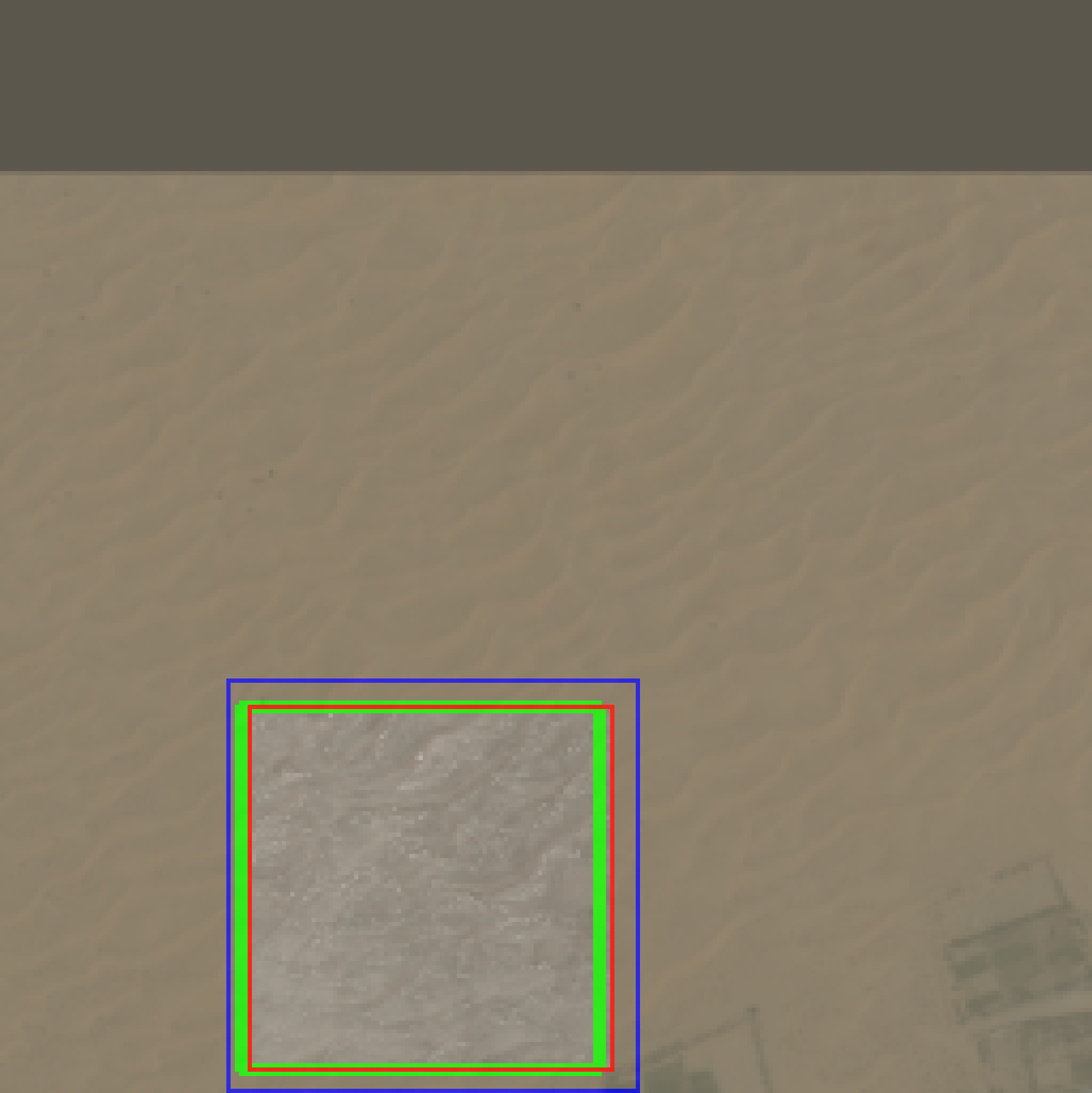}
    \vspace{-0.8\baselineskip}
\end{subfigure}
\begin{subfigure}[b]{0.11\textwidth}
    \includegraphics[width=\textwidth]{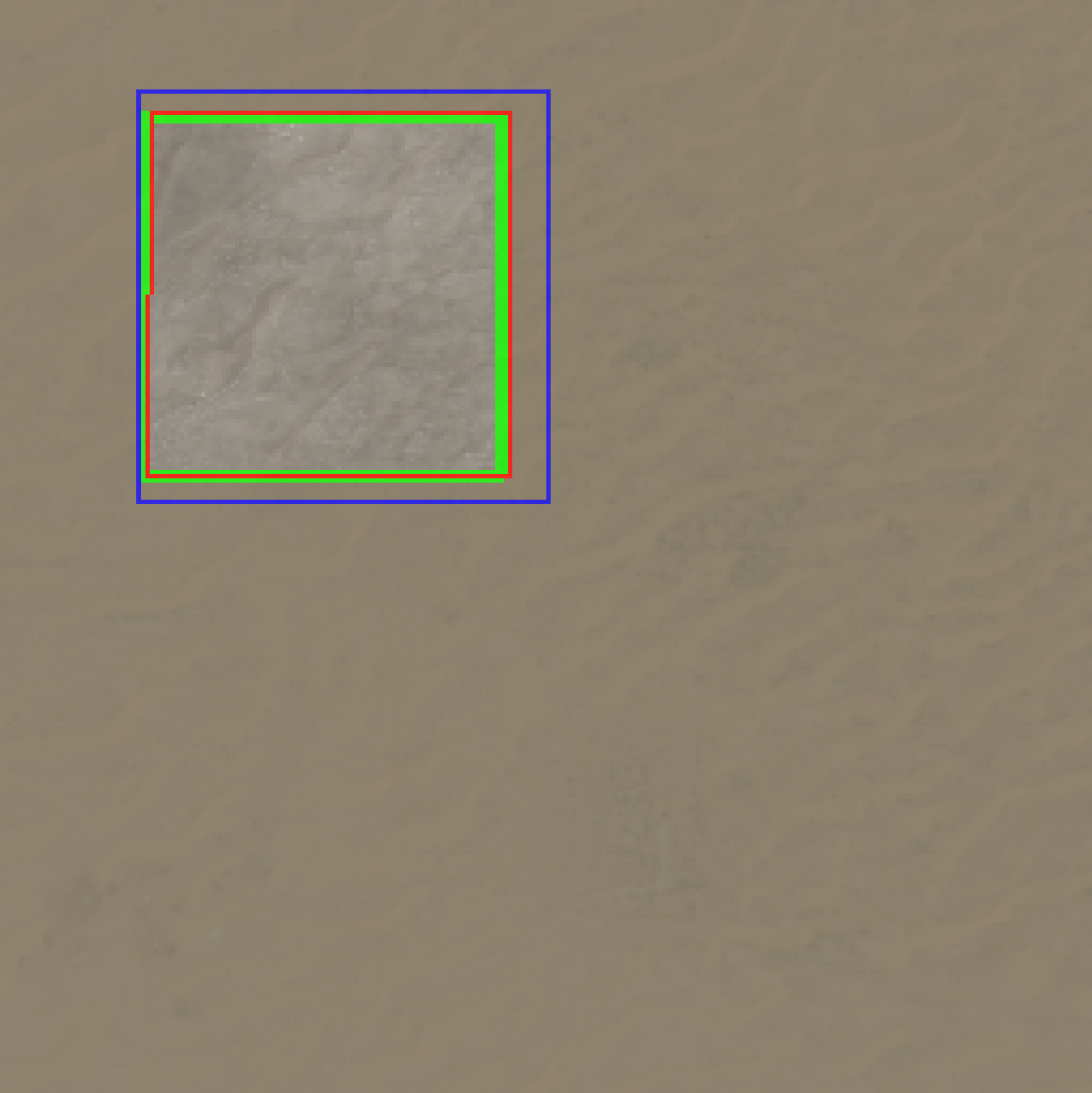}
    \vspace{-0.8\baselineskip}
\end{subfigure}
    \caption{The qualitative comparison between finetuning the refinement module without bbox aug, with only bbox exp, and with bbox aug with $W_S=1536$. {\color{green} Green} boxes are the ground truth, {\color{blue} blue} boxes are the bounding boxes from coarse alignment, and {\color{red} red} boxes are final predictions after refinement.}
    \label{finetune}
    \vspace{-15pt}
\end{figure}

We present a qualitative comparison in Fig.~\ref{finetune} to demonstrate the impact of fine-tuning with and without bounding box augmentation (bbox aug). Given that bounding box augmentation requires an expansion of the bounding box (bbox exp) during the evaluation phase, we also include results featuring solely bbox exp without bbox aug to ablate the effects. The findings illustrate that in the absence of augmentation, the refinement module tends to make only minimal adjustments when not trained with bbox exp. On the other hand, if we train the refinement module with only bbox exp, it always tends to reduce the size of the predicted box towards the center, rather than correctly repositioning it. However, the incorporation of augmentation addresses these limitations by augmenting the width and the center coordinates of the region.

\subsection{Robustness Evaluation and Visualization}\label{sec:robustness}
Ideally, the UAV onboard compass and gimbal camera would supply precise data, enabling the accurate alignment of images to the north. However, it is crucial for our algorithm to demonstrate tolerance towards certain rotation and perspective transformation inaccuracies during active flights. Additionally, understanding how our algorithm performs when there is a change in flight altitude—which results in a change of the thermal image's coverage area, denoted as resizing noise—is essential. To assess the algorithm's robustness under these conditions, we perform experiments that introduce specific rotation, resizing, and perspective transformation noises. For rotation disturbances, the thermal images undergo random rotations up to $5^\circ$, $10^\circ$, or $30^\circ$. For resizing disturbances, the images are randomly scaled by a factor of $1+\Delta r$, with $\Delta r$ varying within either $\pm 0.1$, $\pm 0.2$, or $\pm 0.3$. For perspective transformation, we randomly adjust the four corners of $512\times512$ thermal images up to $8~\si{px}$, $16~\si{px}$, or $32~\si{px}$.

\begin{figure*}
\smallskip
    \centering

\rotatebox{90}{\scriptsize\hspace{2em}One stage}
\begin{subfigure}[b]{0.11\textwidth}
    \includegraphics[width=\textwidth]{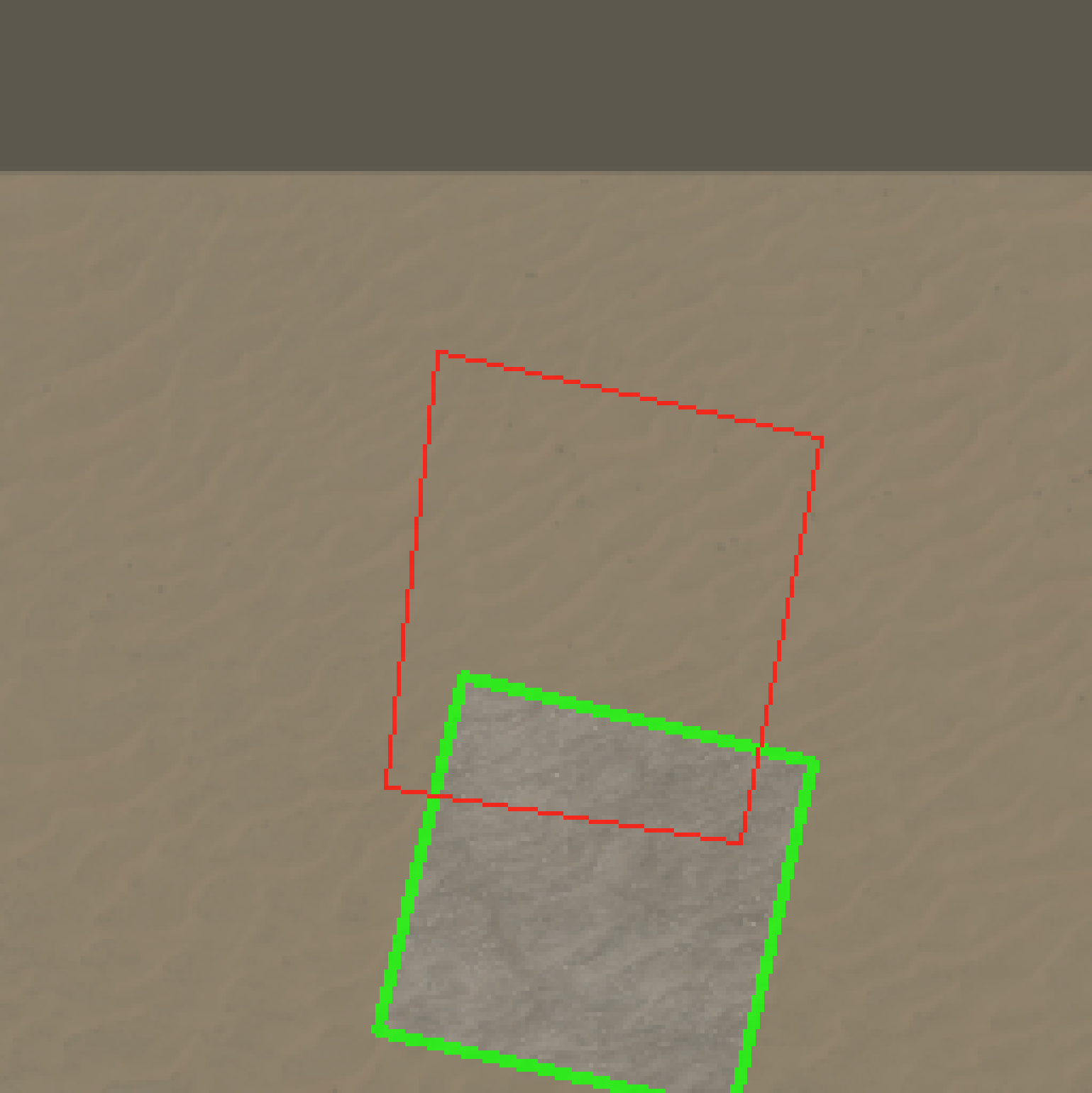}
    \vspace{-0.8\baselineskip}
\end{subfigure}
\begin{subfigure}[b]{0.11\textwidth}
    \includegraphics[width=\textwidth]{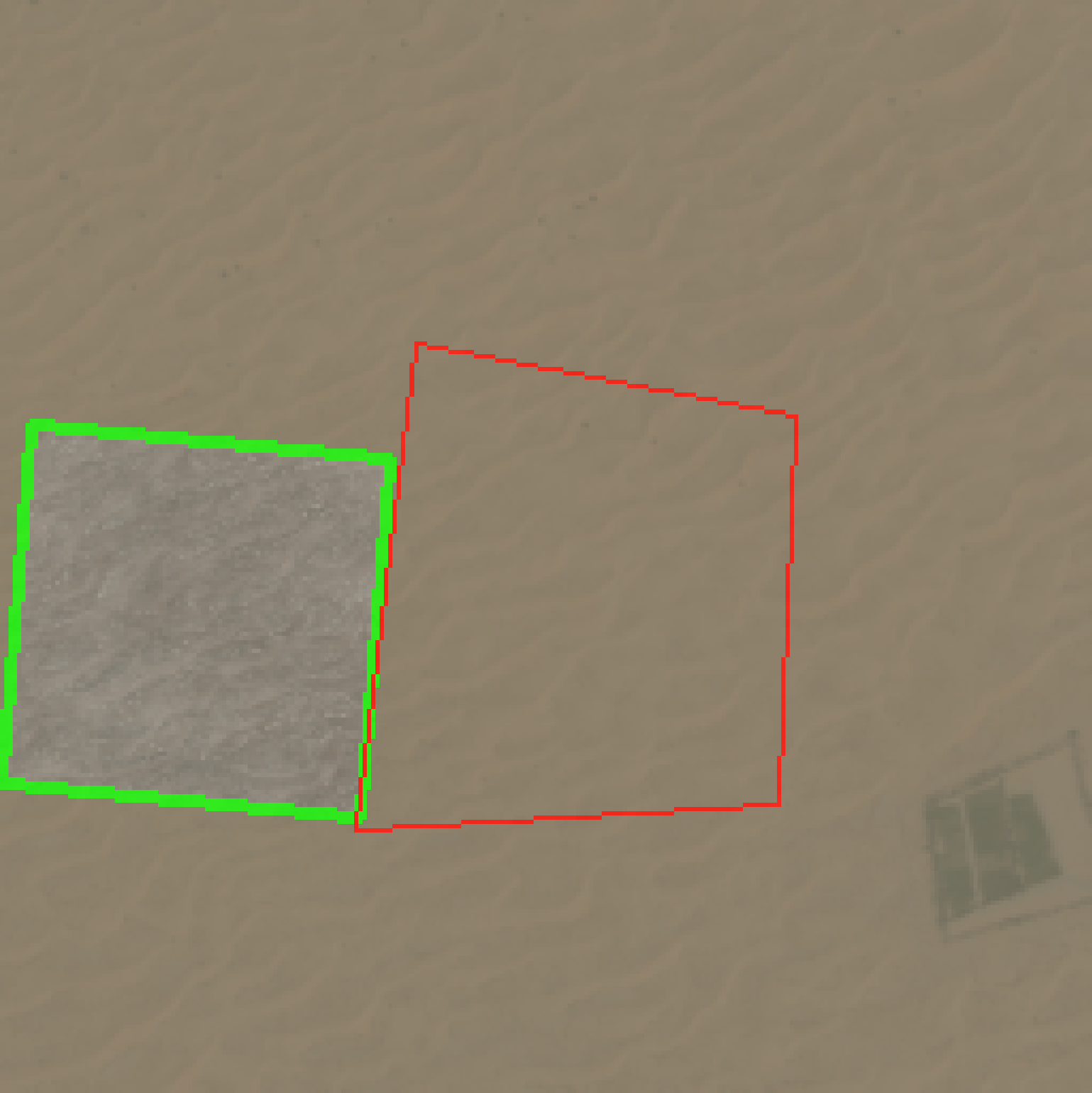}
    \vspace{-0.8\baselineskip}
\end{subfigure}
\begin{subfigure}[b]{0.11\textwidth}
    \includegraphics[width=\textwidth]{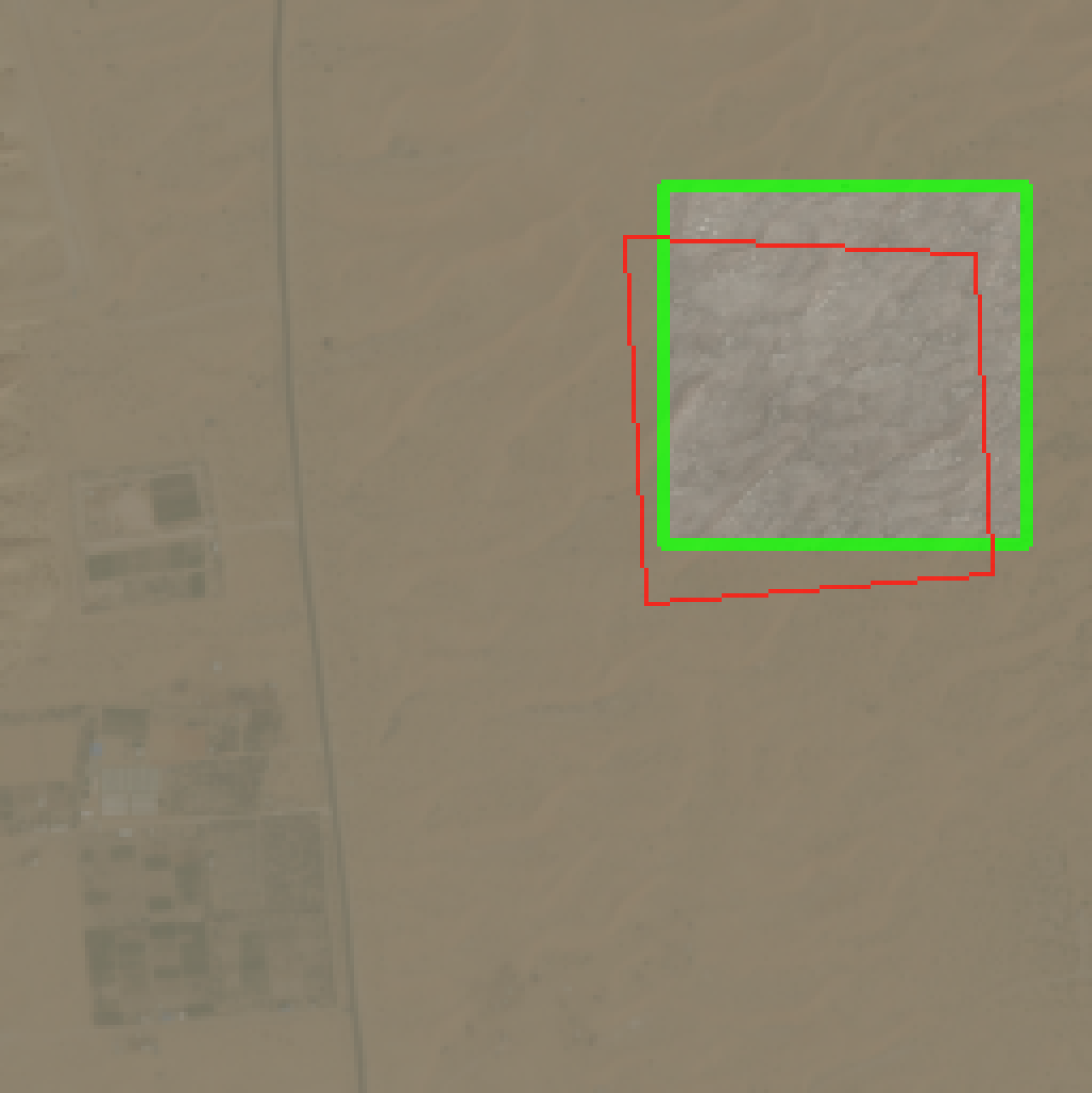}
    \vspace{-0.8\baselineskip}
\end{subfigure}
\begin{subfigure}[b]{0.11\textwidth}
    \includegraphics[width=\textwidth]{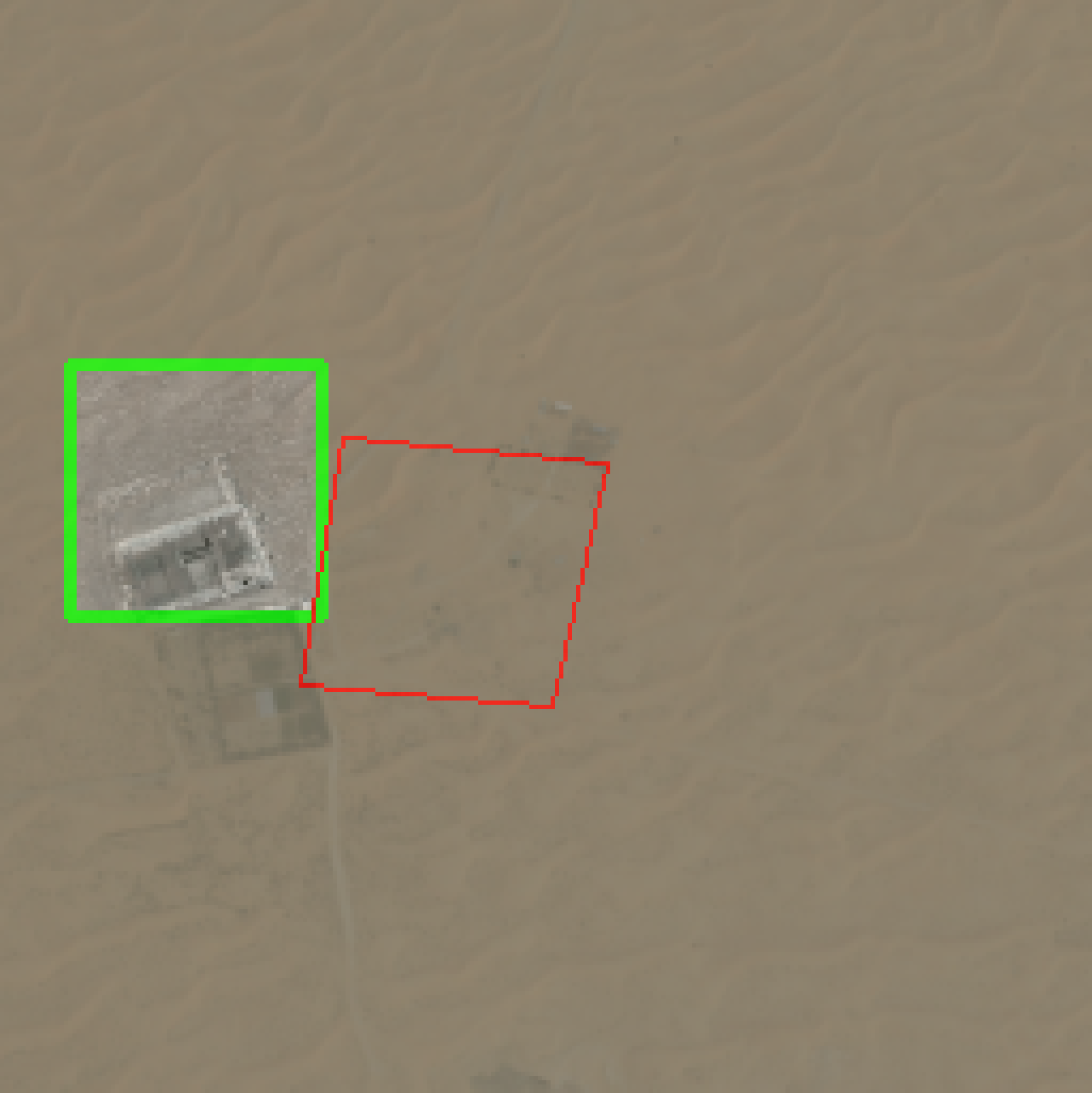}
    \vspace{-0.8\baselineskip}
\end{subfigure}
\begin{subfigure}[b]{0.11\textwidth}
    \includegraphics[width=\textwidth]{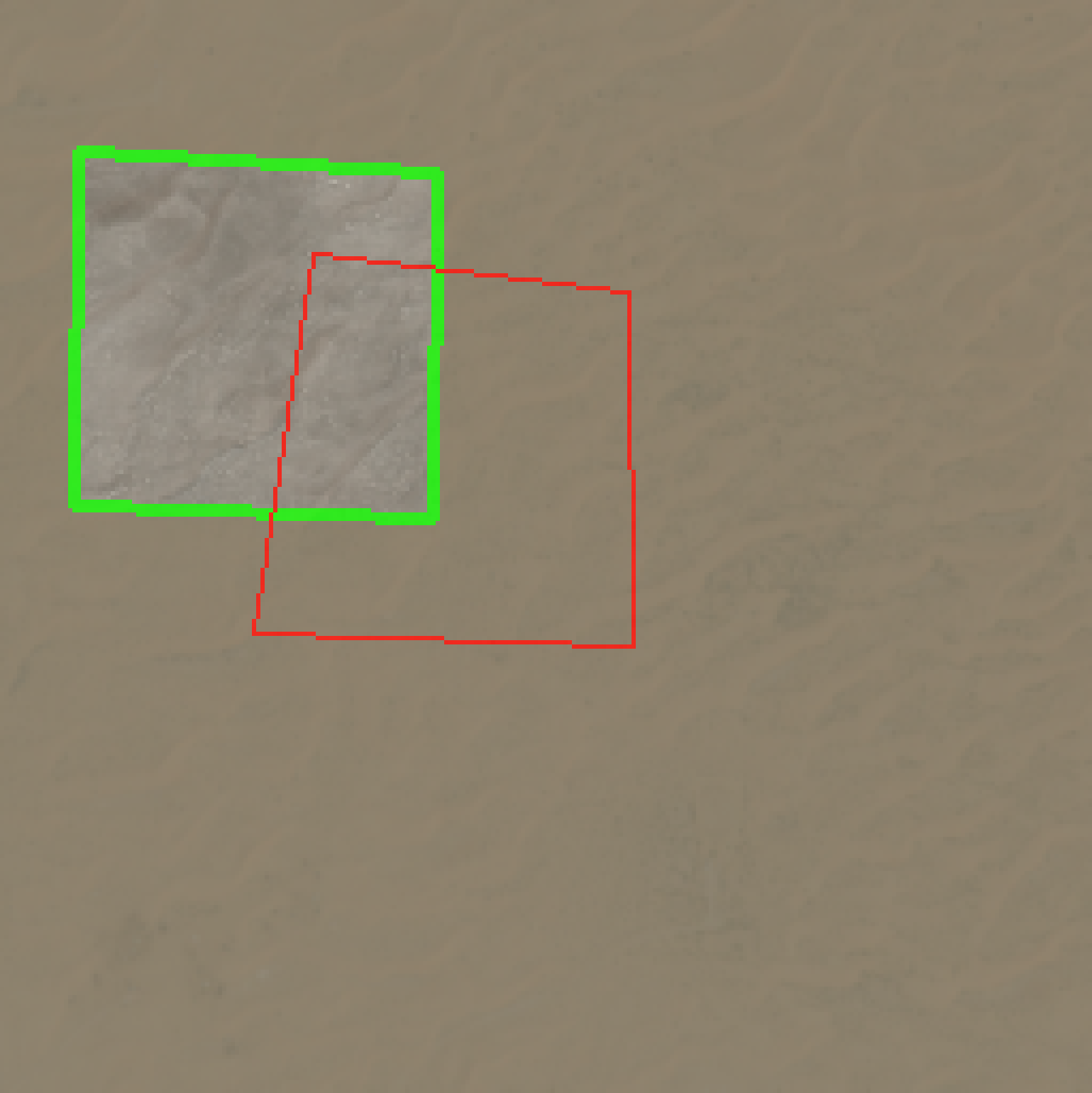}
    \vspace{-0.8\baselineskip}
\end{subfigure}
\begin{subfigure}[b]{0.11\textwidth}
    \includegraphics[width=\textwidth]{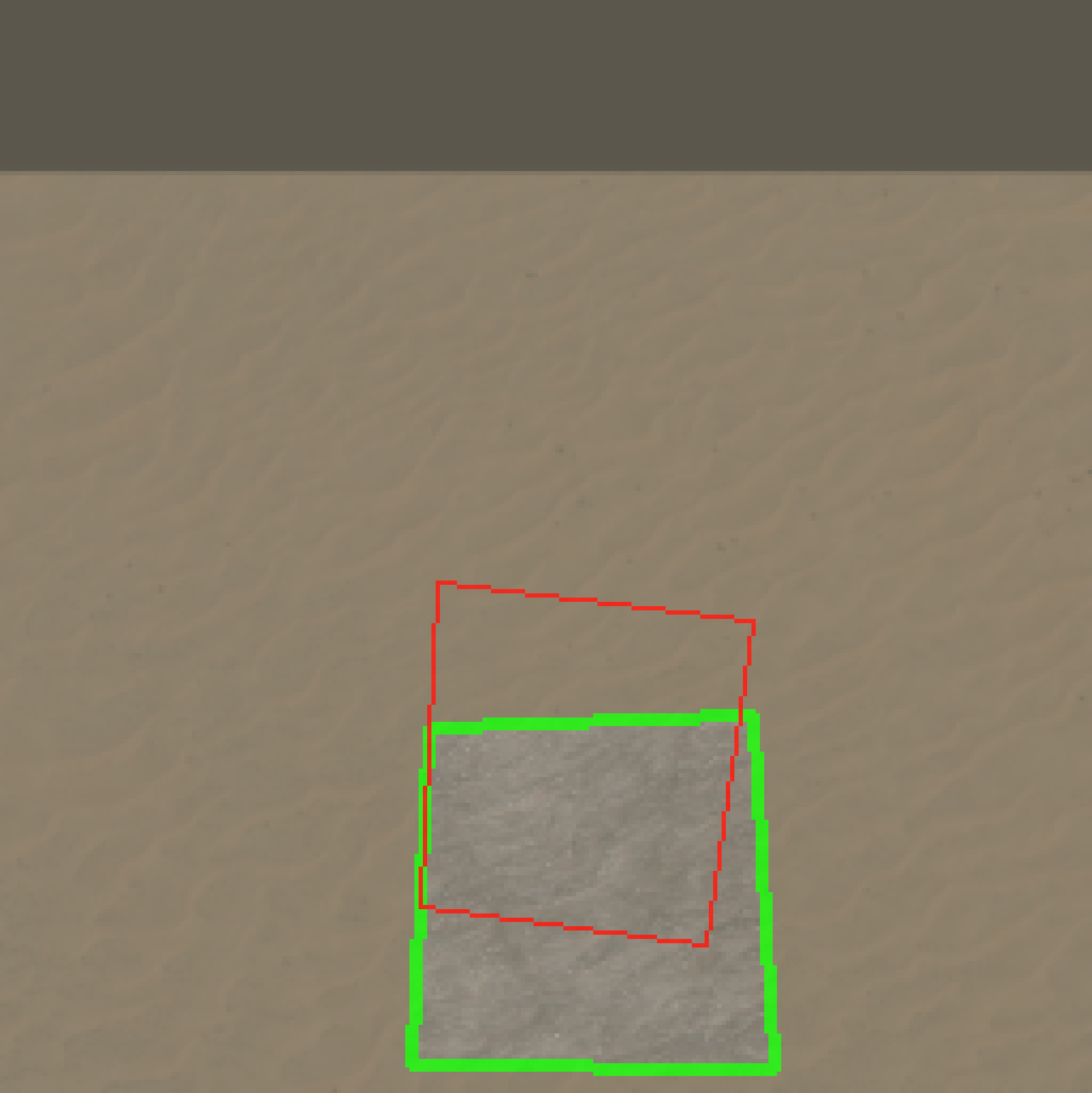}
    \vspace{-0.8\baselineskip}
\end{subfigure}
\begin{subfigure}[b]{0.11\textwidth}
    \includegraphics[width=\textwidth]{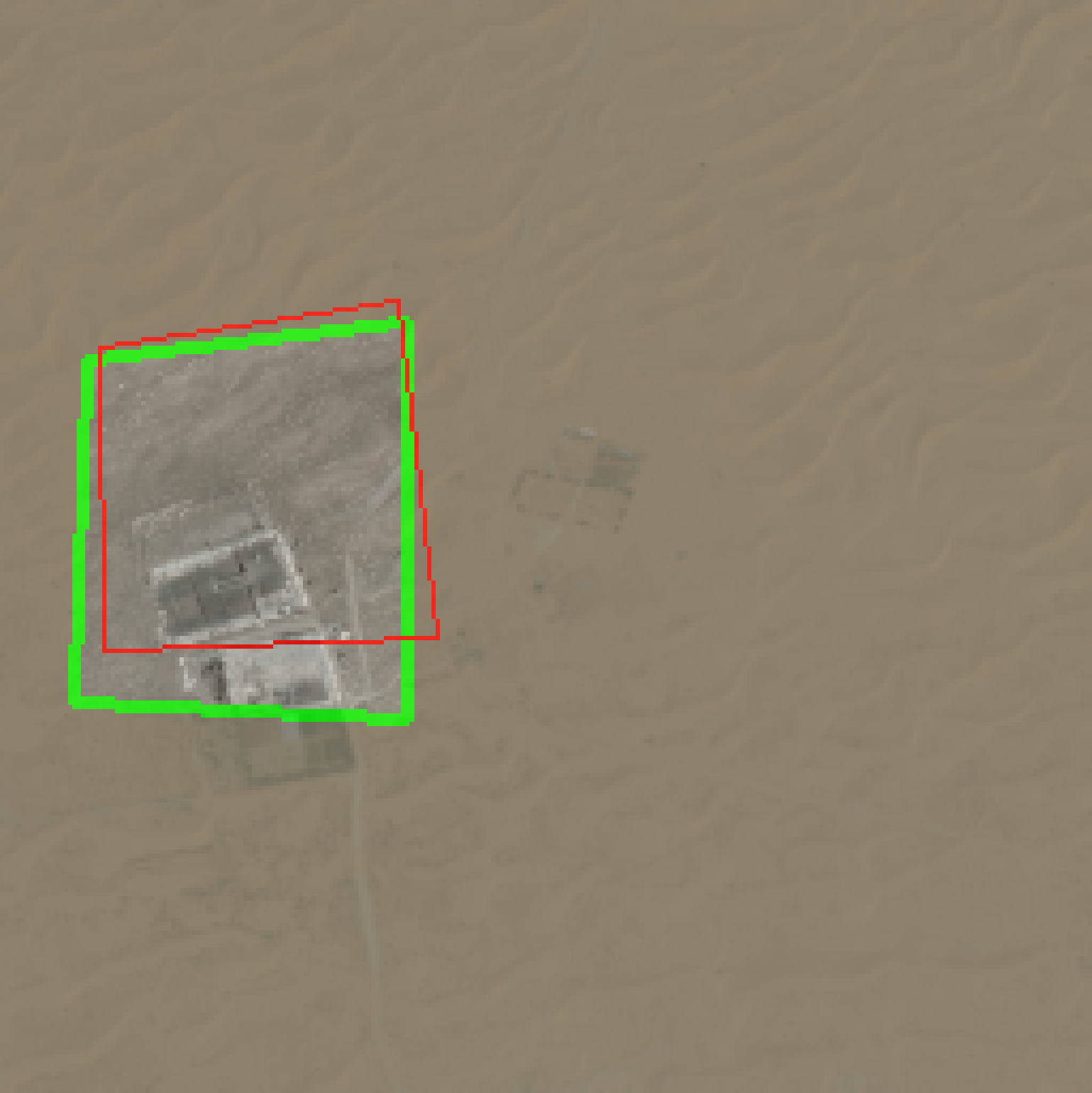}
    \vspace{-0.8\baselineskip}
\end{subfigure}
\begin{subfigure}[b]{0.11\textwidth}
    \includegraphics[width=\textwidth]{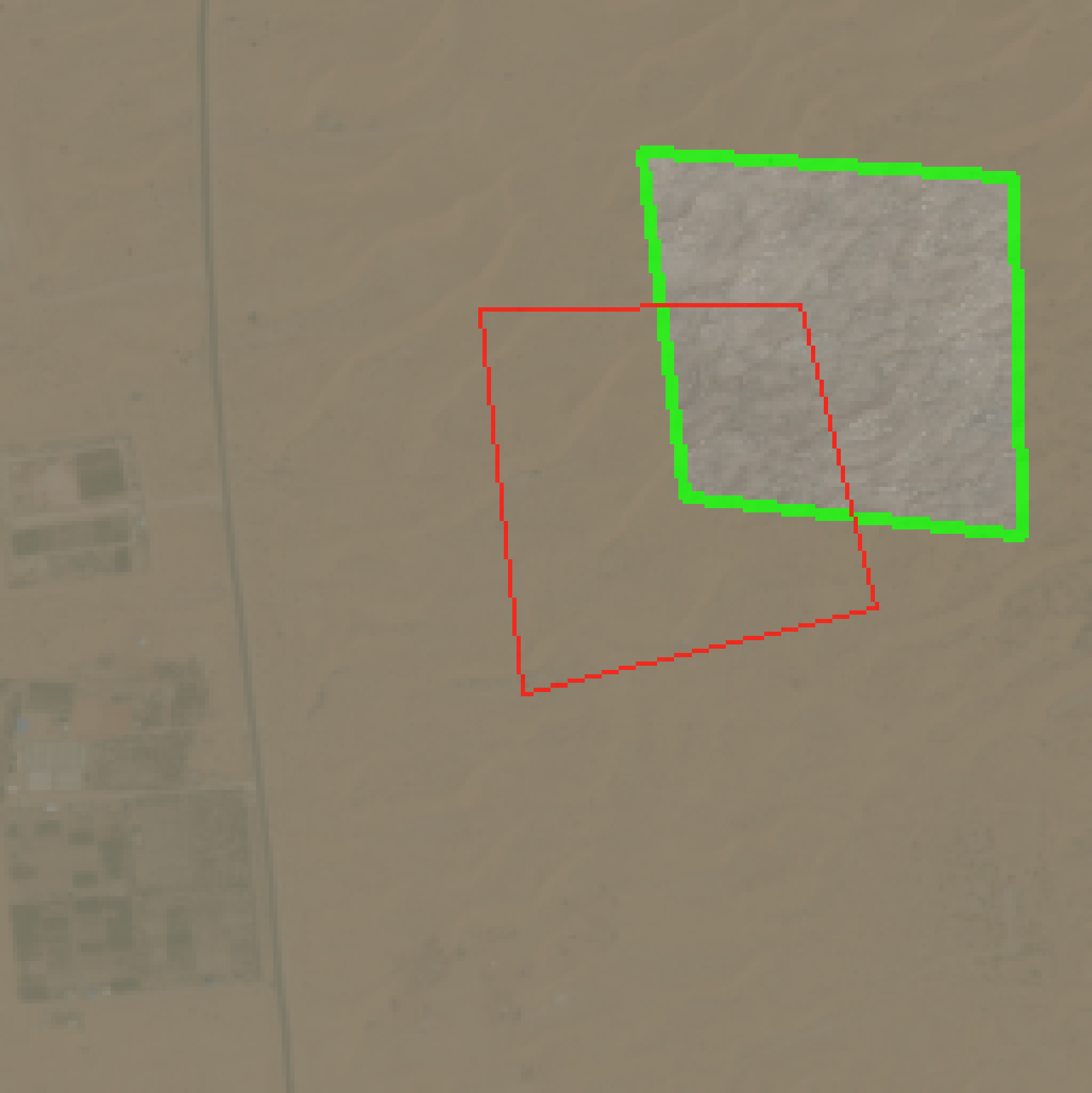}
    \vspace{-0.8\baselineskip}
\end{subfigure}

\rotatebox{90}{\scriptsize\hspace{2em}Two stages}
\begin{subfigure}[b]{0.11\textwidth}
    \includegraphics[width=\textwidth]{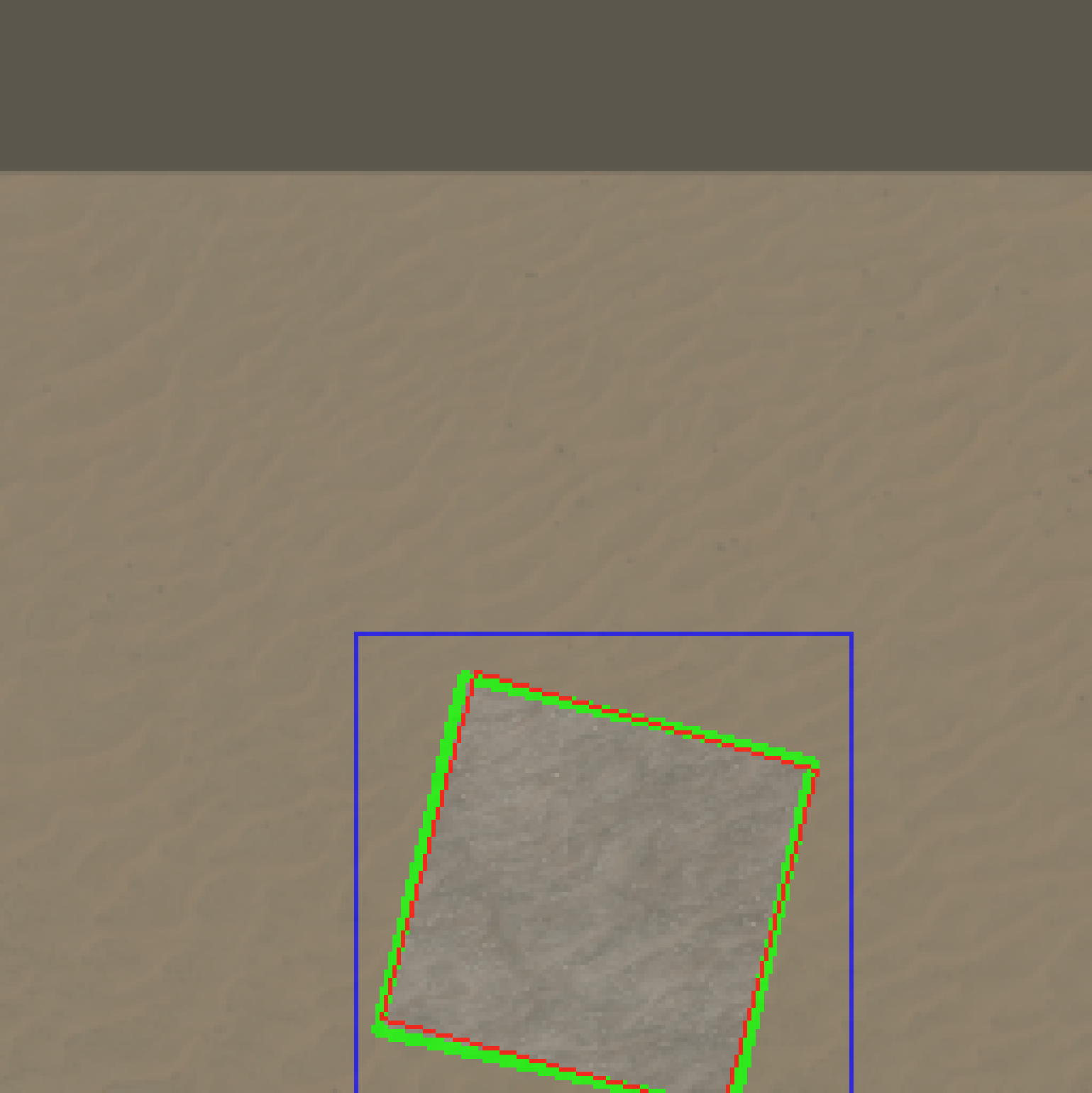}
    \vspace{-0.8\baselineskip}
\end{subfigure}
\begin{subfigure}[b]{0.11\textwidth}
    \includegraphics[width=\textwidth]{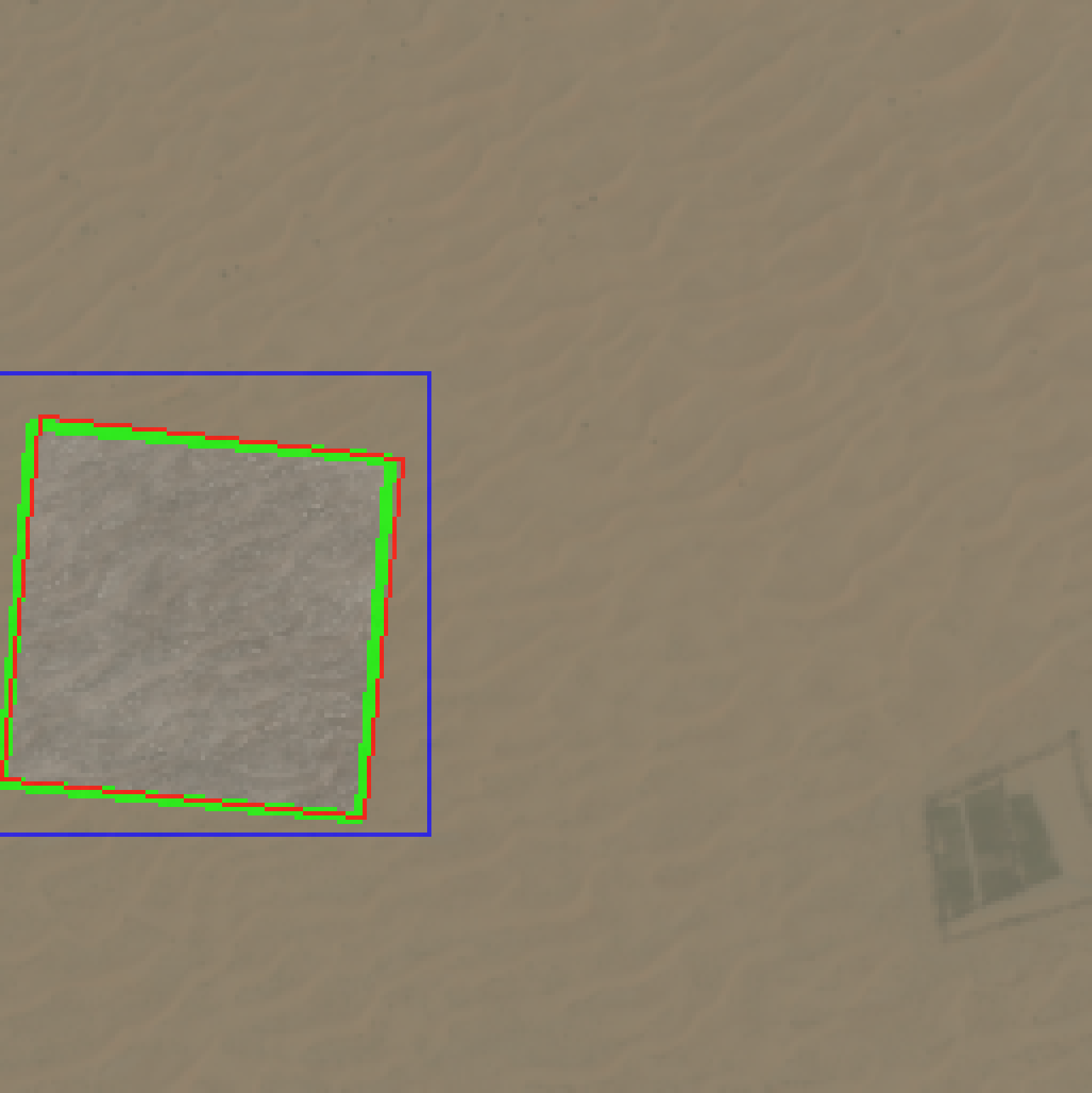}
    \vspace{-0.8\baselineskip}
\end{subfigure}
\begin{subfigure}[b]{0.11\textwidth}
    \includegraphics[width=\textwidth]{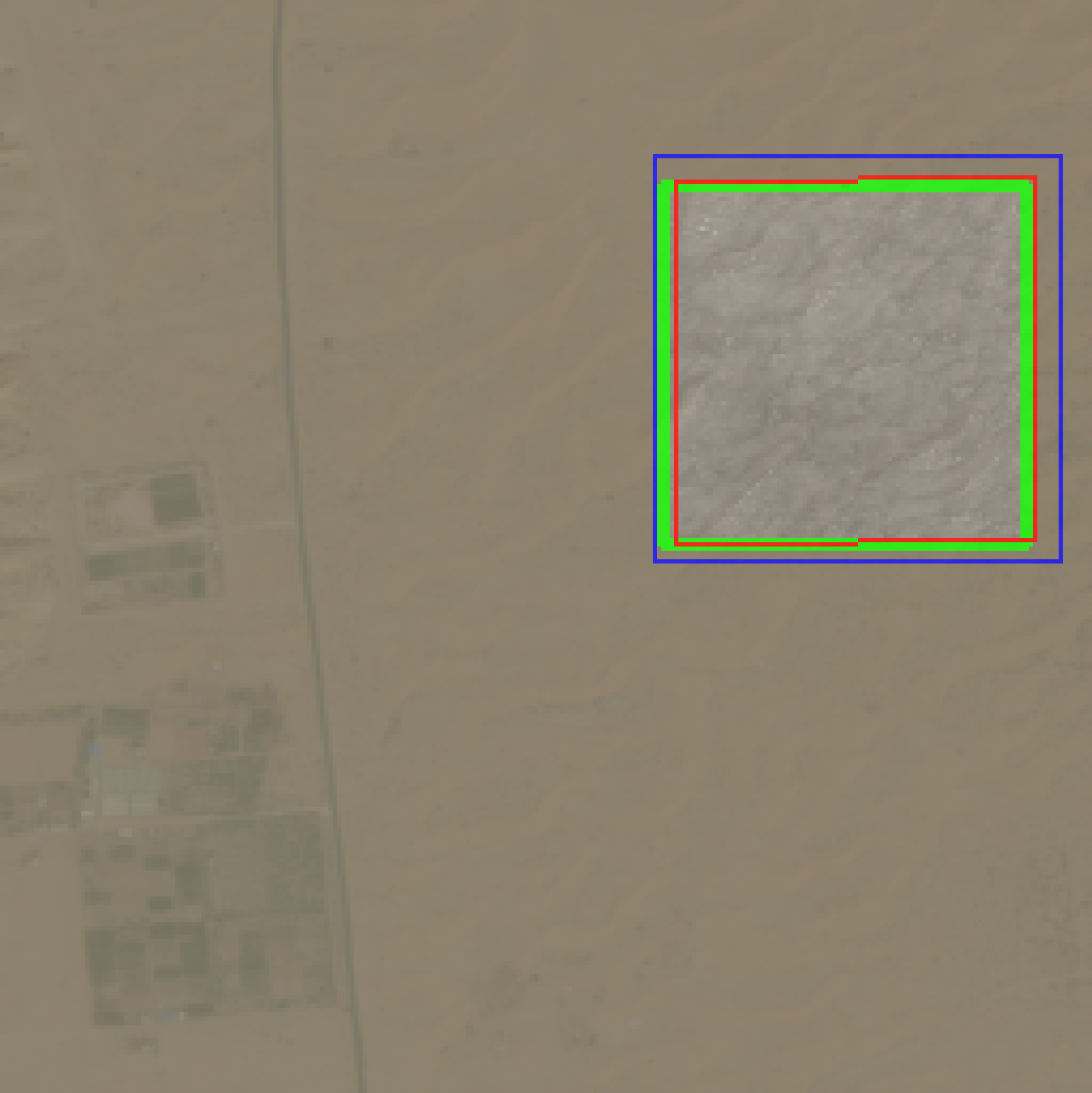}
    \vspace{-0.8\baselineskip}
\end{subfigure}
\begin{subfigure}[b]{0.11\textwidth}
    \includegraphics[width=\textwidth]{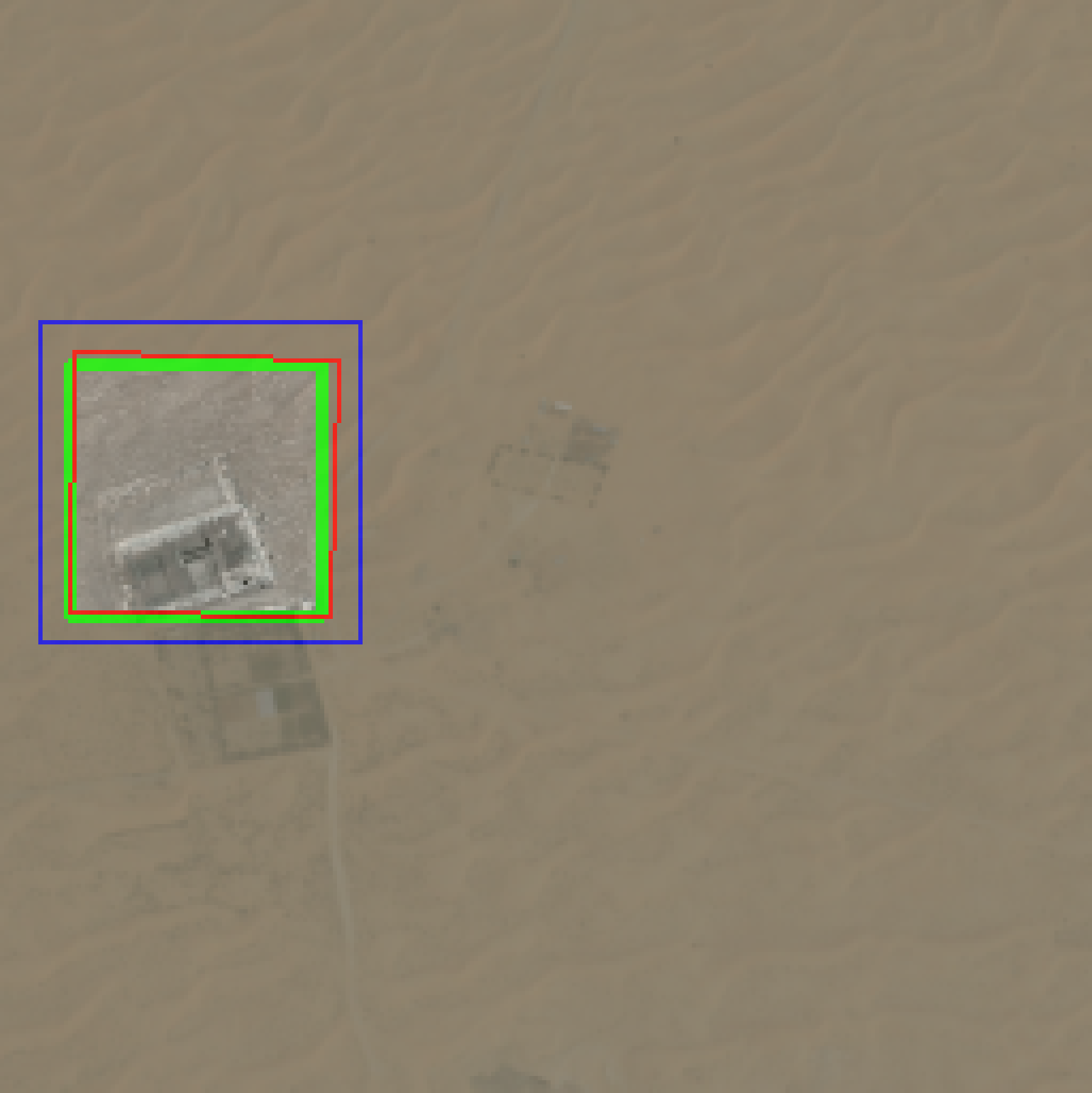}
    \vspace{-0.8\baselineskip}
\end{subfigure}
\begin{subfigure}[b]{0.11\textwidth}
    \includegraphics[width=\textwidth]{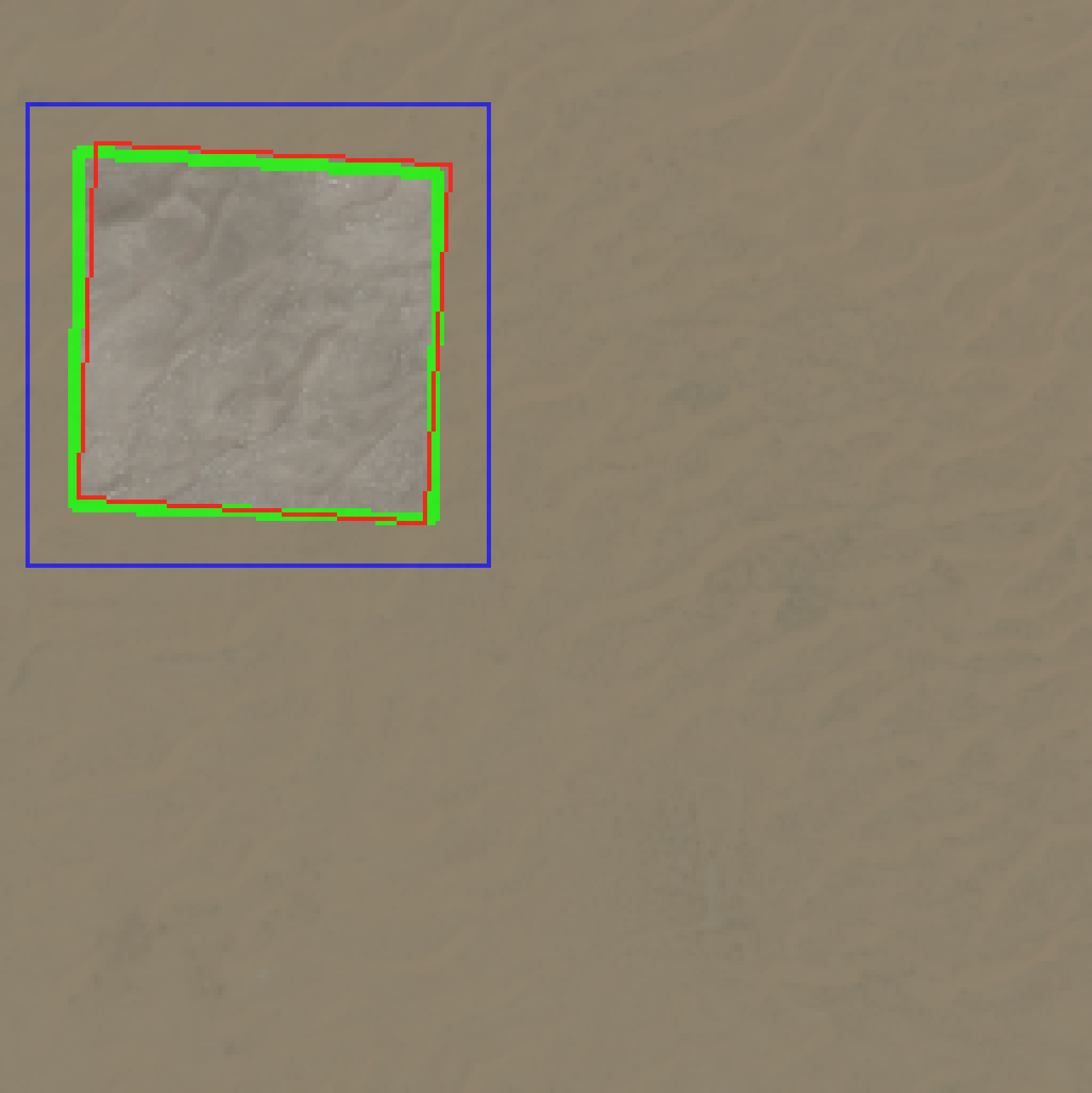}
    \vspace{-0.8\baselineskip}
\end{subfigure}
\begin{subfigure}[b]{0.11\textwidth}
    \includegraphics[width=\textwidth]{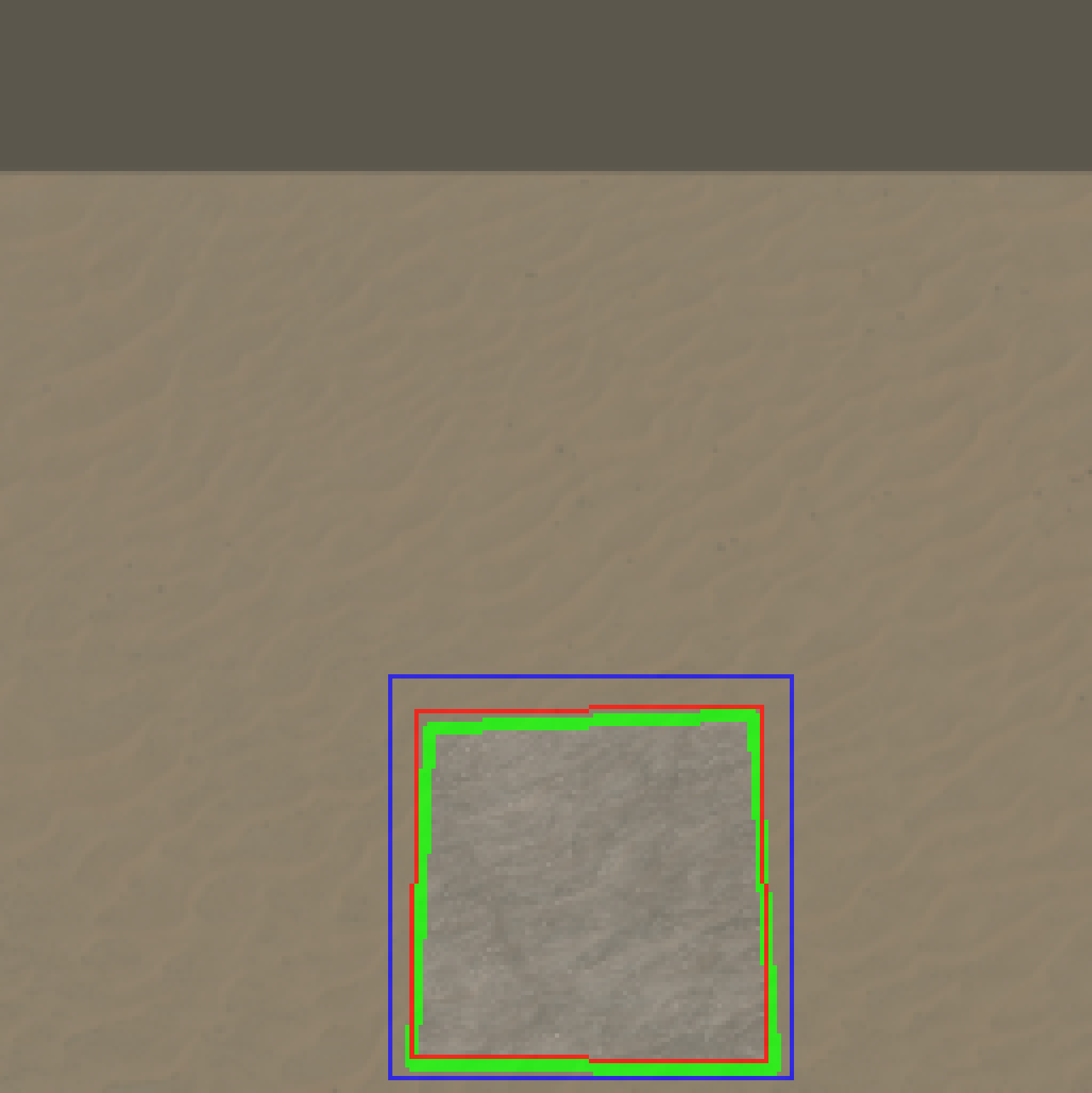}
    \vspace{-0.8\baselineskip}
\end{subfigure}
\begin{subfigure}[b]{0.11\textwidth}
    \includegraphics[width=\textwidth]{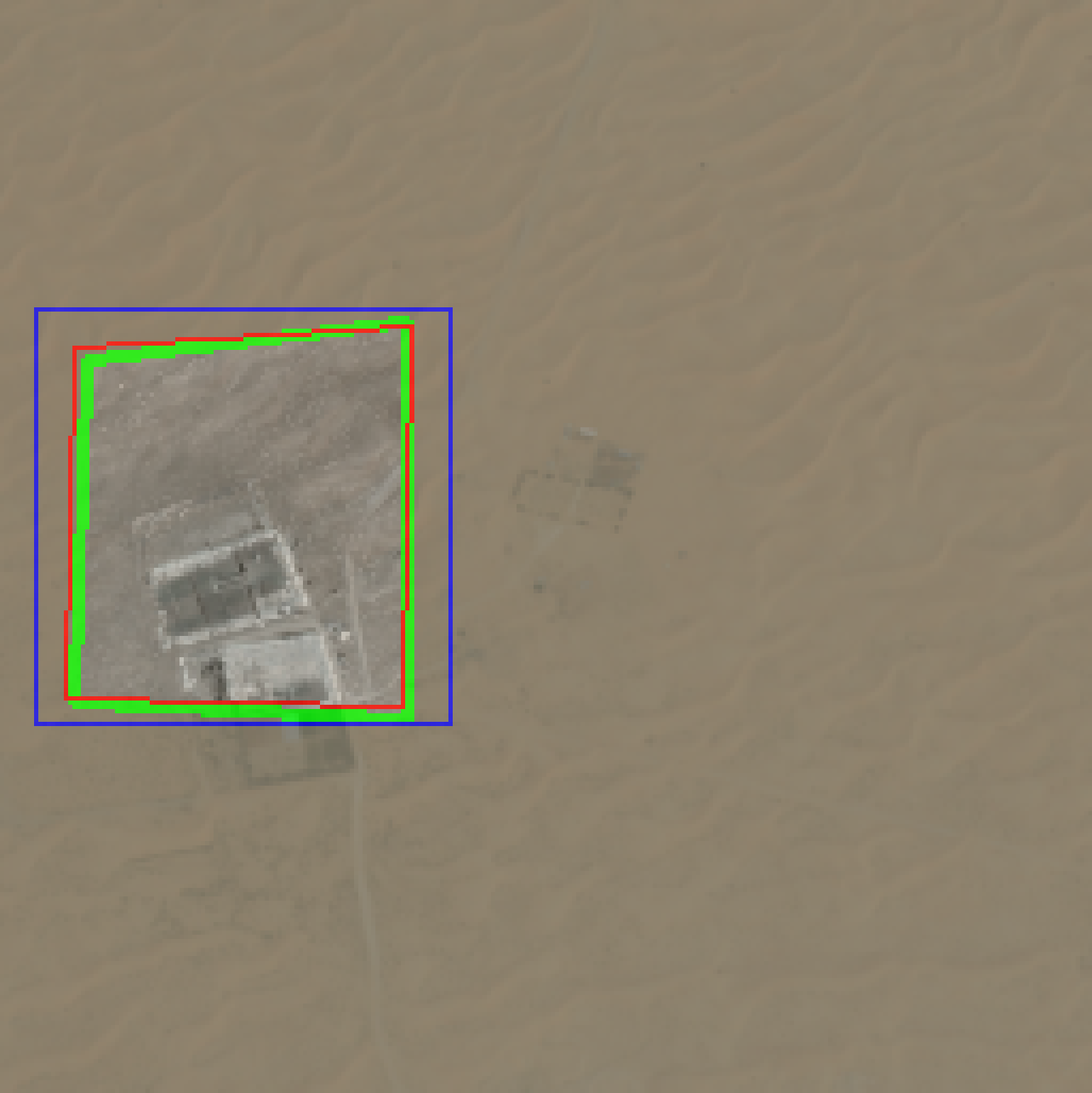}
    \vspace{-0.8\baselineskip}
\end{subfigure}
\begin{subfigure}[b]{0.11\textwidth}
    \includegraphics[width=\textwidth]{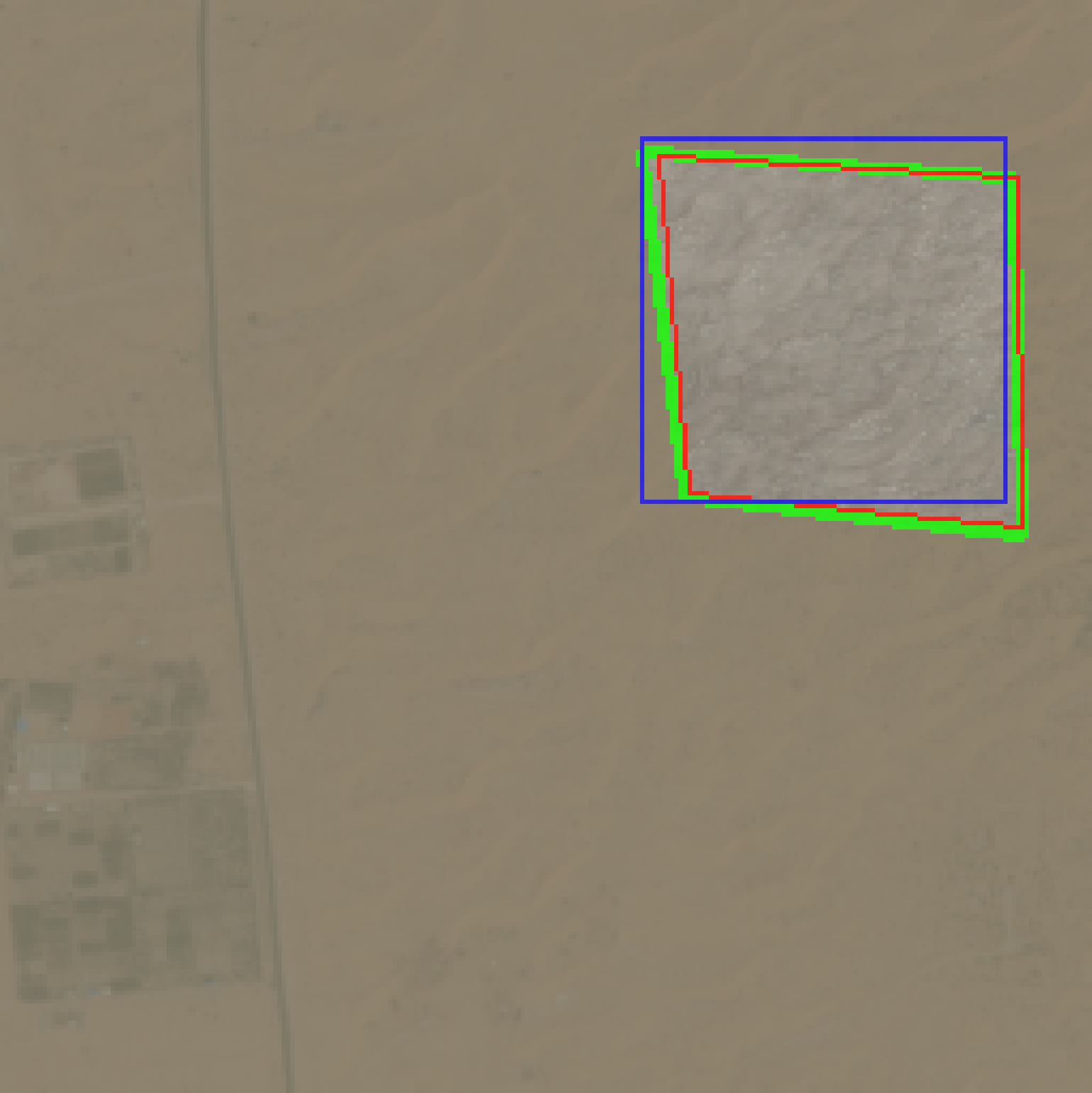}
    \vspace{-0.8\baselineskip}
\end{subfigure}
    \caption{Visualization results with geometric noises for our one-stage and two-stage methods with $W_S=1536$ and $D_C=512$m. {\color{green} Green} boxes are the ground truth, {\color{blue} blue} boxes are the bounding boxes from coarse alignment for our two-stage method, and {\color{red} red} boxes are the final predictions of one-stage and two-stage methods. The $1^\textrm{st}$-$2^\textrm{nd}$ columns show rotation noises, the $3^\textrm{rd}$-$4^\textrm{th}$ columns show resizing noises, and the $5^\textrm{th}$-$8^\textrm{th}$ columns show perspective transformation noises.}
    \label{vis}
    \vspace{-18pt}
\end{figure*}

In Table~\ref{robust}, we evaluate the robustness of our one-stage and two-stage strategies against a variety of geometric noise conditions with $D_C=512~\si{m}$ and $W_S=1536$. The analysis indicates a significant decrease in performance for the one-stage method under these conditions, in contrast to the two-stage strategy, which demonstrates a notable robustness against geometric perturbations. Specifically, the two-stage strategy effectively maintains test MACE below $22~\si{m}$ and test CE below $20~\si{m}$ in most scenarios, with notable exceptions being in instances of $30^\circ$ rotation noise. While incremental perspective transformations and resizing have minimal impact on accuracy, large rotation noise can significantly degrade performance. This suggests the tolerance of our strategies to different types of geometric noise. Overall, the results validate our method's robustness and its ability to estimate these disturbances, underscoring the two-stage strategy's superior effectiveness and reliability in mitigating the negative effects of these disturbances. Fig.~\ref{vis} further illustrates this point by showcasing visual comparisons between the failure instances of the one-stage method and the success cases of the two-stage method, demonstrating the latter's improved robustness.
\vspace{-5pt}

\section{Conclusions}~\label{sec:conclusions}
This paper presents a novel deep homography estimation approach for UAV thermal geo-localization tasks. We validate the capability of STHN to precisely align thermal images, captured by UAV onboard sensors, with large-scale satellite maps, achieving successful alignment even with a size ratio of $11\%$. Additionally, we showcase STHN's superior performances in terms of speed and accuracy with respect to several state-of-the-art approaches as well as its resilience to geometric distortions, which significantly enhances the reliability of geo-localization outcomes.

Our future endeavors will aim to develop a hierarchical geo-localization framework. This framework will integrate deep homography estimation for local matching with image-based matching techniques for broad-scale global matching, thereby building up universal geo-localization solutions.
\vspace{-10pt}

\begin{table}
    \centering
        \caption{Robustness evaluation with geometric noises, including rotation, resizing, and perspective transformation noises when $D_C=512$~m and $W_S=1536$. "Baseline" is our method with only translation.}
        \resizebox{\linewidth}{!}{
    \begin{tabular}{lccccc}
    \toprule
         & \multicolumn{2}{c}{Ours (one stage)} & &\multicolumn{2}{c}{Ours (two stages)}
       \\
       \cline{2-3}  \cline{5-6}\vspace{-5pt}\\
        & Test MACE (\si{m}) & Test CE (\si{m}) & &Test MACE (\si{m}) & Test CE (\si{m}) \\\midrule
        Baseline & 16.42 & 15.90 & & 12.70 & 12.12\\\midrule
        \multicolumn{2}{l}{\textit{Rotation Noises}} \\
         $5^{\circ}$ &31.82 &27.00 & &14.48 &11.42\\
         $10^{\circ}$ &33.78 &28.40 & & 14.55 & 11.90\\
         $30^{\circ}$ &64.63 &56.98  & & 33.48 & 28.93\\\midrule
        \multicolumn{2}{l}{\textit{Resizing Noises}} \\
         $0.1$ & 32.00& 27.38& &13.03 &11.20 \\
         $0.2$ & 36.79& 31.88& &15.16 & 13.23\\
         $0.3$ & 46.22& 40.50& &21.68 & 19.36\\\midrule
         \multicolumn{3}{l}{\textit{Perspective Transformation Noises}} \\
         $8~\si{px}$ &30.53 &26.41 & & 13.70 & 10.94\\
         $16~\si{px}$ &34.07 &27.22 & & 16.24 &11.73 \\
         $32~\si{px}$ &44.03 &30.35 & & 20.90 & 14.11\\
         
    \bottomrule
    \end{tabular}}
    \label{robust}
    \vspace{-15pt}
\end{table}

\bibliographystyle{IEEEtran} 
\bibliography{mybib}

\begin{thebibliography}{10}
\providecommand{\url}[1]{#1}
\csname url@samestyle\endcsname
\providecommand{\newblock}{\relax}
\providecommand{\bibinfo}[2]{#2}
\providecommand{\BIBentrySTDinterwordspacing}{\spaceskip=0pt\relax}
\providecommand{\BIBentryALTinterwordstretchfactor}{4}
\providecommand{\BIBentryALTinterwordspacing}{\spaceskip=\fontdimen2\font plus
\BIBentryALTinterwordstretchfactor\fontdimen3\font minus \fontdimen4\font\relax}
\providecommand{\BIBforeignlanguage}[2]{{%
\expandafter\ifx\csname l@#1\endcsname\relax
\typeout{** WARNING: IEEEtran.bst: No hyphenation pattern has been}%
\typeout{** loaded for the language `#1'. Using the pattern for}%
\typeout{** the default language instead.}%
\else
\language=\csname l@#1\endcsname
\fi
#2}}
\providecommand{\BIBdecl}{\relax}
\BIBdecl

\bibitem{agriculture}
D.~C. Tsouros, S.~Bibi, and P.~G. Sarigiannidis, ``A review on uav-based applications for precision agriculture,'' \emph{Information}, vol.~10, no.~11, p. 349, 2019.

\bibitem{searchrescue}
M.~Półka, S.~Ptak, and Łukasz Kuziora, ``The use of uav's for search and rescue operations,'' \emph{Procedia Engineering}, vol. 192, pp. 748--752, 2017, 12th international scientific conference of young scientists on sustainable, modern and safe transport.

\bibitem{saviolo2023unifying}
A.~Saviolo, P.~Rao, V.~Radhakrishnan, J.~Xiao, and G.~Loianno, ``Unifying foundation models with quadrotor control for visual tracking beyond object categories,'' in \emph{IEEE International Conference on Robotics and Automation (ICRA)}, 2024.

\bibitem{powerline}
J.~Xing, G.~Cioffi, J.~Hidalgo-Carrió, and D.~Scaramuzza, ``Autonomous power line inspection with drones via perception-aware mpc,'' in \emph{IEEE/RSJ International Conference on Intelligent Robots and Systems (IROS)}, 2023, pp. 1086--1093.

\bibitem{drones6110347}
L.~Morando, C.~T. Recchiuto, J.~Calla, P.~Scuteri, and A.~Sgorbissa, ``Thermal and visual tracking of photovoltaic plants for autonomous uav inspection,'' \emph{Drones}, vol.~6, no.~11, 2022.

\bibitem{review_avl}
A.~Couturier and M.~A. Akhloufi, ``A review on absolute visual localization for uav,'' \emph{Robotics and Autonomous Systems}, vol. 135, p. 103666, 2021.

\bibitem{foundloc}
Y.~He, I.~Cisneros, N.~Keetha, J.~Patrikar, Z.~Ye, I.~Higgins, Y.~Hu, P.~Kapoor, and S.~Scherer, ``Foundloc: Vision-based onboard aerial localization in the wild,'' \emph{arXiv preprint arXiv:2310.16299}, 2023.

\bibitem{vgscience}
A.~T. Fragoso, C.~T. Lee, A.~S. McCoy, and S.-J. Chung, ``A seasonally invariant deep transform for visual terrain-relative navigation,'' \emph{Science Robotics}, vol.~6, no.~55, p. eabf3320, 2021.

\bibitem{directalign3}
B.~Patel, T.~D. Barfoot, and A.~P. Schoellig, ``Visual localization with google earth images for robust global pose estimation of uavs,'' in \emph{IEEE International Conference on Robotics and Automation (ICRA)}, 2020, pp. 6491--6497.

\bibitem{imgregistration}
M.~Shan, F.~Wang, F.~Lin, Z.~Gao, Y.~Z. Tang, and B.~M. Chen, ``Google map aided visual navigation for uavs in gps-denied environment,'' in \emph{IEEE International Conference on Robotics and Biomimetics (ROBIO)}, 2015, pp. 114--119.

\bibitem{stl}
J.~Xiao, D.~Tortei, E.~Roura, and G.~Loianno, ``Long-range uav thermal geo-localization with satellite imagery,'' in \emph{IEEE/RSJ International Conference on Intelligent Robots and Systems (IROS)}, 2023, pp. 5820--5827.

\bibitem{cao2022iterative}
S.-Y. Cao, J.~Hu, Z.~Sheng, and H.-L. Shen, ``Iterative deep homography estimation,'' in \emph{Proceedings of the IEEE/CVF Conference on Computer Vision and Pattern Recognition (CVPR)}, 2022, pp. 1879--1888.

\bibitem{detone2016deep}
D.~DeTone, T.~Malisiewicz, and A.~Rabinovich, ``Deep image homography estimation,'' \emph{arXiv preprint arXiv:1606.03798}, 2016.

\bibitem{10.1007/978-3-030-58452-8_38}
J.~Zhang, C.~Wang, S.~Liu, L.~Jia, N.~Ye, J.~Wang, J.~Zhou, and J.~Sun, ``Content-aware unsupervised deep homography estimation,'' in \emph{Computer Vision -- ECCV 2020}, A.~Vedaldi, H.~Bischof, T.~Brox, and J.-M. Frahm, Eds.\hskip 1em plus 0.5em minus 0.4em\relax Cham: Springer International Publishing, 2020, pp. 653--669.

\bibitem{zhao2021deep}
Y.~Zhao, X.~Huang, and Z.~Zhang, ``Deep lucas-kanade homography for multimodal image alignment,'' in \emph{Proceedings of the IEEE/CVF conference on computer vision and pattern recognition (CVPR)}, 2021, pp. 15\,950--15\,959.

\bibitem{shao2021localtrans}
R.~Shao, G.~Wu, Y.~Zhou, Y.~Fu, L.~Fang, and Y.~Liu, ``Localtrans: A multiscale local transformer network for cross-resolution homography estimation,'' in \emph{Proceedings of the IEEE/CVF international conference on computer vision (CVPR)}, 2021, pp. 14\,890--14\,899.

\bibitem{Cao_2023_CVPR}
S.-Y. Cao, R.~Zhang, L.~Luo, B.~Yu, Z.~Sheng, J.~Li, and H.-L. Shen, ``Recurrent homography estimation using homography-guided image warping and focus transformer,'' in \emph{Proceedings of the IEEE/CVF Conference on Computer Vision and Pattern Recognition (CVPR)}, June 2023, pp. 9833--9842.

\bibitem{le2020deep}
H.~Le, F.~Liu, S.~Zhang, and A.~Agarwala, ``Deep homography estimation for dynamic scenes,'' in \emph{Proceedings of the IEEE/CVF conference on computer vision and pattern recognition (CVPR)}, 2020, pp. 7652--7661.

\bibitem{directalign}
G.~J.~V. Dalen, D.~P. Magree, and E.~N. Johnson, ``Absolute localization using image alignment and particle filtering,'' in \emph{AIAA Guidance, Navigation, and Control Conference}, 2016.

\bibitem{directalign2}
A.~Yol, B.~Delabarre, A.~Dame, J.-E. Dartois, and E.~Marchand, ``Vision-based absolute localization for unmanned aerial vehicles,'' in \emph{IEEE/RSJ International Conference on Intelligent Robots and Systems (IROS)}, 2014, pp. 3429--3434.

\bibitem{imgregistration2}
M.~Mantelli, D.~Pittol, R.~Neuland, A.~Ribacki, R.~Maffei, V.~Jorge, E.~Prestes, and M.~Kolberg, ``A novel measurement model based on abbrief for global localization of a uav over satellite images,'' \emph{Robotics and Autonomous Systems}, vol. 112, pp. 304--319, 2019.

\bibitem{voplusvg}
A.~Shetty and G.~X. Gao, ``Uav pose estimation using cross-view geolocalization with satellite imagery,'' in \emph{IEEE International Conference on Robotics and Automation (ICRA)}, 2019, pp. 1827--1833.

\bibitem{VGUAV5}
H.~Goforth and S.~Lucey, ``Gps-denied uav localization using pre-existing satellite imagery,'' in \emph{IEEE International Conference on Robotics and Automation (ICRA)}, 2019, pp. 2974--2980.

\bibitem{VGUAV4}
M.~Bianchi and T.~D. Barfoot, ``Uav localization using autoencoded satellite images,'' \emph{IEEE Robotics and Automation Letters}, vol.~6, no.~2, pp. 1761--1768, 2021.

\bibitem{lecun2015deep}
Y.~LeCun, Y.~Bengio, and G.~Hinton, ``Deep learning,'' \emph{nature}, vol. 521, no. 7553, pp. 436--444, 2015.

\bibitem{thermaluav3}
J.~Delaune, R.~Hewitt, L.~Lytle, C.~Sorice, R.~Thakker, and L.~Matthies, ``Thermal-inertial odometry for autonomous flight throughout the night,'' in \emph{IEEE/RSJ International Conference on Intelligent Robots and Systems (IROS)}, 2019, pp. 1122--1128.

\bibitem{thermaluav4}
V.~Polizzi, R.~Hewitt, J.~Hidalgo-Carrió, J.~Delaune, and D.~Scaramuzza, ``Data-efficient collaborative decentralized thermal-inertial odometry,'' \emph{IEEE Robotics and Automation Letters}, vol.~7, no.~4, pp. 10\,681--10\,688, 2022.

\bibitem{pmlr-v155-achermann21a}
F.~Achermann, A.~Kolobov, D.~Dey, T.~Hinzmann, J.~J. Chung, R.~Siegwart, and N.~Lawrance, ``Multipoint: Cross-spectral registration of thermal and optical aerial imagery,'' in \emph{Proceedings of the 2020 Conference on Robot Learning}, ser. Proceedings of Machine Learning Research, J.~Kober, F.~Ramos, and C.~Tomlin, Eds., vol. 155.\hskip 1em plus 0.5em minus 0.4em\relax PMLR, 16--18 Nov 2021, pp. 1746--1760.

\bibitem{keetha2023anyloc}
N.~Keetha, A.~Mishra, J.~Karhade, K.~M. Jatavallabhula, S.~Scherer, M.~Krishna, and S.~Garg, ``Anyloc: Towards universal visual place recognition,'' \emph{IEEE Robotics and Automation Letters}, vol.~9, no.~2, pp. 1286--1293, 2023.

\bibitem{nguyen2018unsupervised}
T.~Nguyen, S.~W. Chen, S.~S. Shivakumar, C.~J. Taylor, and V.~Kumar, ``Unsupervised deep homography: A fast and robust homography estimation model,'' \emph{IEEE Robotics and Automation Letters}, vol.~3, no.~3, pp. 2346--2353, 2018.

\bibitem{electronics12040788}
Y.~Luo, X.~Wang, Y.~Wu, and C.~Shu, ``Infrared and visible image homography estimation using multiscale generative adversarial network,'' \emph{Electronics}, vol.~12, no.~4, 2023.

\bibitem{cgan}
M.~Mirza and S.~Osindero, ``Conditional generative adversarial nets,'' \emph{arXiv preprint arXiv:1411.1784}, 2014.

\bibitem{electronics12214441}
X.~Wang, Y.~Luo, Q.~Fu, Y.~He, C.~Shu, Y.~Wu, and Y.~Liao, ``Coarse-to-fine homography estimation for infrared and visible images,'' \emph{Electronics}, vol.~12, no.~21, 2023.

\bibitem{pix2pix}
P.~Isola, J.-Y. Zhu, T.~Zhou, and A.~A. Efros, ``Image-to-image translation with conditional adversarial networks,'' in \emph{IEEE/CVF Conference on Computer Vision and Pattern Recognition (CVPR)}, 2017, pp. 1125--1134.

\bibitem{raft}
Z.~Teed and J.~Deng, ``Raft: Recurrent all-pairs field transforms for optical flow,'' in \emph{Computer Vision -- ECCV 2020}, A.~Vedaldi, H.~Bischof, T.~Brox, and J.-M. Frahm, Eds.\hskip 1em plus 0.5em minus 0.4em\relax Cham: Springer International Publishing, 2020, pp. 402--419.

\bibitem{abdel2015direct}
Y.~I. Abdel-Aziz, H.~M. Karara, and M.~Hauck, ``Direct linear transformation from comparator coordinates into object space coordinates in close-range photogrammetry,'' \emph{Photogrammetric engineering \& remote sensing}, vol.~81, no.~2, pp. 103--107, 2015.

\bibitem{loshchilov2017decoupled}
I.~Loshchilov and F.~Hutter, ``Decoupled weight decay regularization,'' in \emph{International Conference on Learning Representations}, 2019.

\bibitem{SIFT}
D.~G. Lowe, ``Distinctive image features from scale-invariant keypoints,'' \emph{International journal of computer vision}, vol.~60, no.~2, pp. 91--110, 2004.

\bibitem{ransac}
M.~A. Fischler and R.~C. Bolles, ``Random sample consensus: a paradigm for model fitting with applications to image analysis and automated cartography,'' \emph{Commun. ACM}, vol.~24, no.~6, p. 381–395, jun 1981.

\bibitem{magsac}
D.~Baráth, J.~Noskova, M.~Ivashechkin, and J.~Matas, ``Magsac++, a fast, reliable and accurate robust estimator,'' in \emph{IEEE/CVF Conference on Computer Vision and Pattern Recognition (CVPR)}, 2020, pp. 1301--1309.

\bibitem{ORB}
E.~Rublee, V.~Rabaud, K.~Konolige, and G.~Bradski, ``Orb: An efficient alternative to sift or surf,'' in \emph{International Conference on Computer Vision (ICCV)}, 2011, pp. 2564--2571.

\bibitem{6126542}
S.~Leutenegger, M.~Chli, and R.~Y. Siegwart, ``Brisk: Binary robust invariant scalable keypoints,'' in \emph{International Conference on Computer Vision (ICCV)}, 2011, pp. 2548--2555.

\bibitem{r2d2}
J.~Revaud, P.~Weinzaepfel, C.~R. de~Souza, and M.~Humenberger, ``{R2D2:} repeatable and reliable detector and descriptor,'' in \emph{NeurIPS}, 2019.

\bibitem{sun2021loftr}
J.~Sun, Z.~Shen, Y.~Wang, H.~Bao, and X.~Zhou, ``{LoFTR}: Detector-free local feature matching with transformers,'' \emph{CVPR}, 2021.

\bibitem{netvlad}
R.~Arandjelović, P.~Gronat, A.~Torii, T.~Pajdla, and J.~Sivic, ``Netvlad: Cnn architecture for weakly supervised place recognition,'' \emph{IEEE Transactions on Pattern Analysis and Machine Intelligence}, vol.~40, no.~6, pp. 1437--1451, 2018.

\bibitem{8382272}
F.~Radenović, G.~Tolias, and O.~Chum, ``Fine-tuning cnn image retrieval with no human annotation,'' \emph{IEEE Transactions on Pattern Analysis and Machine Intelligence}, vol.~41, no.~7, pp. 1655--1668, 2019.

\bibitem{oquab2023dinov2}
M.~Oquab, T.~Darcet, T.~Moutakanni, H.~Vo, M.~Szafraniec, V.~Khalidov, P.~Fernandez, D.~Haziza, F.~Massa, A.~El-Nouby \emph{et~al.}, ``Dinov2: Learning robust visual features without supervision,'' \emph{arXiv preprint arXiv:2304.07193}, 2023.

\bibitem{VLAD}
H.~Jégou, M.~Douze, C.~Schmid, and P.~Pérez, ``Aggregating local descriptors into a compact image representation,'' in \emph{IEEE/CVF Conference on Computer Vision and Pattern Recognition (CVPR)}, 2010, pp. 3304--3311.

\end{thebibliography}

\end{document}